\pdfoutput=1
\documentclass[11pt]{article}

\usepackage{fullpage}
\usepackage{amsmath,amssymb,amsthm,graphicx} 
\usepackage{epsfig}
\usepackage[numbers]{natbib} 
\usepackage{psfrag}
\usepackage{enumerate} 
\usepackage{enumitem} 
\usepackage{setspace}
\usepackage{float} 
\usepackage{color}
\usepackage{array}
\usepackage{pgf,tikz}
\usepackage{mathrsfs}
\usepackage{hyperref}
\usepackage{subcaption}
\usepackage{wrapfig}
\usepackage[ruled,vlined]{algorithm2e}
\usepackage{tabularx}
\usepackage{makecell}

\usepackage[T1]{fontenc}
\usepackage{lmodern}
\usepackage{caption}
\usepackage{adjustbox}
\usepackage{geometry}

\usepackage[utf8]{inputenc} 
\usepackage[T1]{fontenc}    
\usepackage{hyperref}       
\usepackage{url}            
\usepackage{booktabs}       
\usepackage{amsfonts}       
\usepackage{nicefrac}       
\usepackage{microtype}      

\usetikzlibrary{arrows}
\definecolor{ffqqqq}{rgb}{1.,0.,0.}
\definecolor{xfqqff}{rgb}{0.4980392156862745,0.,1.}

\definecolor{mypink}{rgb}{0.858, 0.188, 0.478}
\definecolor{myred}{rgb}{1, 0, 0}

\makeatletter
\long\def\@makecaption#1#2{
        \vskip 0.8ex
        \setbox\@tempboxa\hbox{\small {\bf #1:} #2}
        \dimen0=\hsize
        \advance\dimen0 by 0cm
        \ifdim \wd\@tempboxa >\dimen0
                \hbox to \hsize{
                        \parindent 0em
                        \hfil 
                        \parbox{\dimen0}{\def\baselinestretch{0.96}\small
                                {\bf #1.} #2
                                } 
                        \hfil}
        \else \hbox to \hsize{\hfil \box\@tempboxa \hfil}
        \fi
        \vspace{0.4cm}
        }
\makeatother

\usepackage{macros}

\usepackage{amsmath,amsfonts,bm}

\def\eqref#1{equation~\ref{#1}}

\def\1{\bm{1}}

\def\eps{{\epsilon}}

\DeclareMathAlphabet{\mathsfit}{\encodingdefault}{\sfdefault}{m}{sl}
\SetMathAlphabet{\mathsfit}{bold}{\encodingdefault}{\sfdefault}{bx}{n}

\usepackage{etoolbox}
\makeatletter
\patchcmd{\@algocf@start}
  {-1.5em}
  {0pt}
  {}{}
\makeatother

\makeatletter
\renewcommand{\paragraph}{%
  \@startsection{paragraph}{4}%
  {\z@}{0.25ex \@plus 1ex \@minus .2ex}{-1em}%
  {\normalfont\normalsize\bfseries}%
}
\makeatother

\newif\ifarxiv
\arxivtrue

\renewcommand\citet{\citep}

\makeatletter
\appto\@floatboxreset{%
  \ifx\@captype\andy@table
    \sffamily
  \fi
}
\def\andy@table{table}
\makeatother

\usepackage{caption} 
\title{Semi-supervised novelty detection using\\ ensembles with regularized disagreement}

\author{Alexandru Țifrea, Eric Stavarache, Fanny Yang \\
Department of Computer Science\\
ETH Zurich, Switzerland \\
\texttt{\{tifreaa,ericst,fan.yang\}@ethz.ch} \\
}
 
\begin{document}
\hypersetup{pageanchor=false}
\maketitle

\begin{abstract}

Deep neural networks often predict samples with high confidence even when they
come from unseen classes and should instead be flagged for expert evaluation.
Current novelty detection algorithms cannot reliably identify such near OOD
points unless they have access to labeled data that is similar to these novel
samples. In this paper, we develop a new ensemble-based procedure for
\emph{semi-supervised novelty detection} (SSND) that successfully leverages a
mixture of unlabeled ID and novel-class samples to achieve good detection
performance.  In particular, we show how to achieve disagreement only on OOD
data using early stopping regularization. While we prove this fact for a simple
data distribution, our extensive experiments suggest that it holds true for more
complex scenarios: our approach significantly outperforms state-of-the-art SSND
methods on standard image data sets (SVHN/CIFAR-10/CIFAR-100) and medical image
data sets with only a negligible increase in computation cost.

\end{abstract}

\section{Introduction}

Despite achieving great in-distribution (ID) prediction performance, deep neural
networks (DNN) often have trouble dealing with test samples that are
out-of-distribution (OOD), i.e.\ test inputs that are unlike the data seen
during training. In particular, DNNs often make incorrect predictions with high
confidence when new unseen classes emerge over time (e.g.\ undiscovered bacteria
\citep{jieren}, new diseases \citep{Katsamenis20}).
Instead, 
we would like to automatically \emph{detect} such novel samples and bring them
to the attention of human experts.

\begin{figure}[t]
  \begin{center}
    \includegraphics[width=0.7\columnwidth]{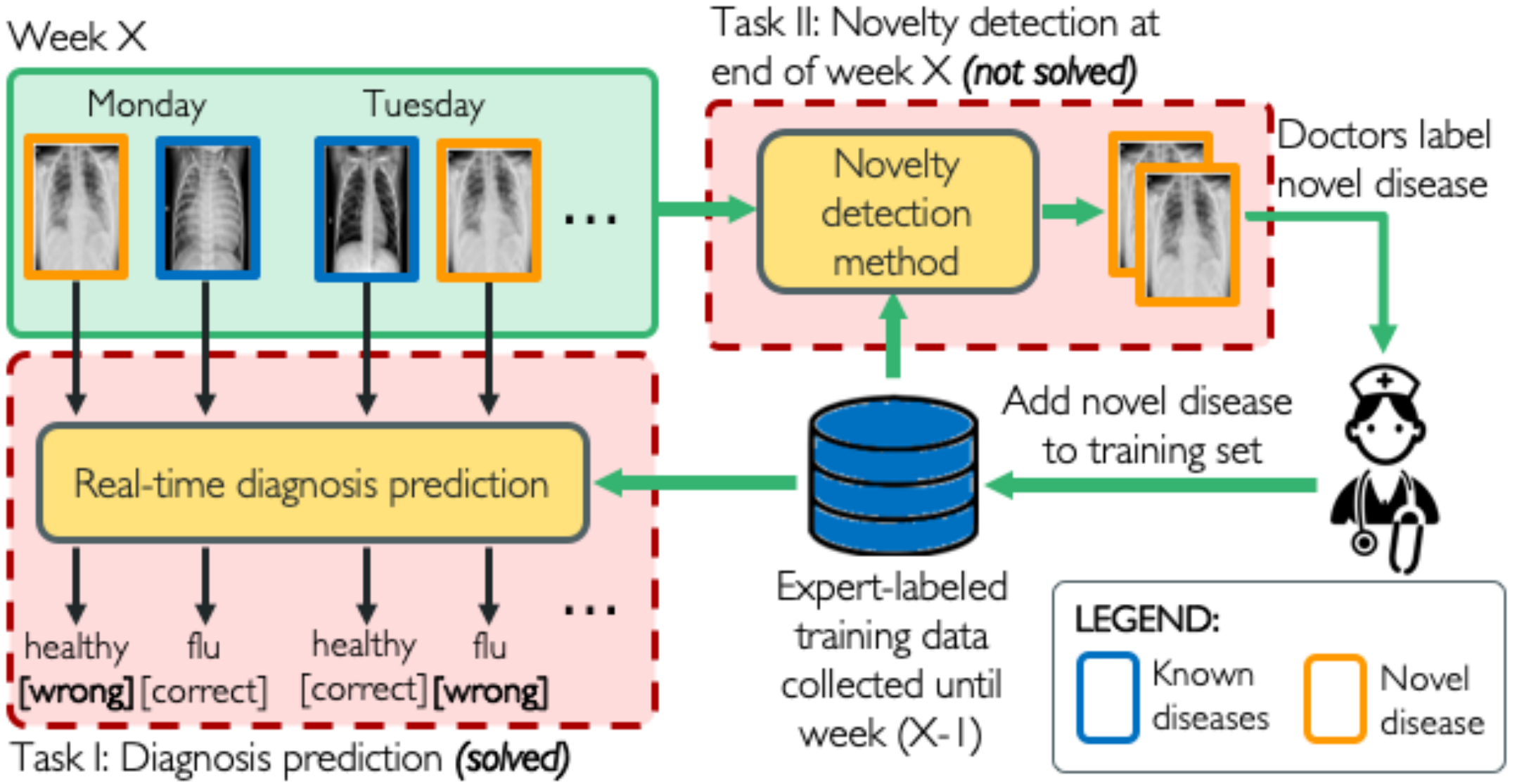}
  \end{center}

  \vspace{-0.3cm} \caption{ \small{Novelty detection is challenging since
  X-rays of novel diseases are remarkably similar to known conditions. The unlabeled batch
  of inference-time data can be used to adapt a semi-supervised novelty
  detection approach to emerging novel
  diseases.}}

  \label{fig:practical_sketch}
\end{figure}

Consider, for instance, a hospital with a severe shortage of qualified
personnel. To make up for the lack of doctors, the hospital would like to use an
automated system for real-time diagnosis from X-ray images (Task I) and a
novelty detection system, which can run at the end of each week, to detect
outbreaks of novel disease variants (Task II) (see
Figure~\ref{fig:practical_sketch}).  In particular, the detection algorithm can
be fine-tuned weekly with the unlabeled batch of data collected during the
respective week.
 
While the experts are examining the peculiar X-rays over the course of the next
week, the novelty detection model helps to collect more instances of the same
new condition and can request human review for these patients.
The human experts can
then label these images and include them in the labeled training set to update
both the diagnostic prediction and the novelty detection systems. This process
repeats each week and enables both diagnostic and novelty detection models to
adjust to new emerging diseases.



Note that, in this example, the novelties are a particular kind of
out-of-distribution samples with two properties.
First, several novel-class samples may appear in the unlabeled batch at the end
of a week, e.g.\ a contagious disease will lead to several people in a small
area to be infected. This situation is different from cases when outliers are
assumed to be singular, e.g.\ anomaly detection problems.  Second, the
novel-class samples share many features in common with the ID data, and only
differ from known classes in certain minute details. For instance, both ID and
OOD samples are frontal chest X-rays, with the OOD samples showing distinctive
signs of a pneumonia caused by a new virus. In what follows, we use the terms
\emph{novelty detection} and \emph{OOD samples} to refer to data with these
characteristics.

Automated diagnostic prediction systems (Task I) can already often have
satisfactory performance \citep{Calli2021}. In contrast, novelty
detection (Task II) still poses a challenging problem in these scenarios. Many
prior approaches can be used for semi-supervised novelty detection (SSND), when
a batch of unlabeled data that may contain OOD samples is available, like in
Figure~\ref{fig:practical_sketch}.\footnote{We use the same definition of SSND
  as the survey by \citet{Bulusu2020}, whereas some works use the term to refer
  to supervised \citep{Gornitz2013, Daniel2019, Ruff2020} or unsupervised ND
\citep{Song2017, ganomaly2018} according to our taxonomy in
Section~\ref{sec:setting}.} However, all of these methods fail to detect
novel-class data when used with complex models, like neural networks.

\begin{figure*}[t]
  \centering

    \includegraphics[width=0.8\textwidth]{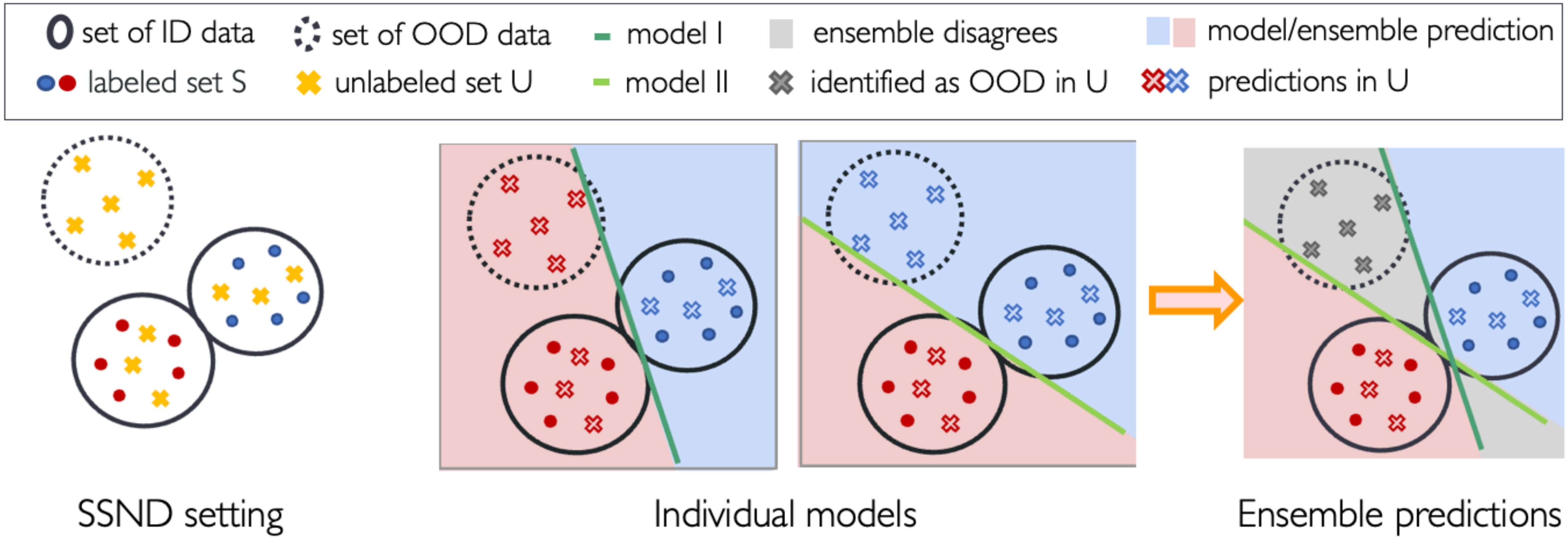}

    \caption{\small{\textbf{Left:} Sketch of the SSND setting.
    \textbf{Middle and Right:} Novelty detection with a diverse ensemble.}}

   \label{fig:setting_ensemble}
\end{figure*}

Despite showing great success on simple benchmarks like SVHN vs CIFAR10, SOTA
unsupervised OOD detection methods perform poorly on near OOD data
\citep{winkens2020} where OOD inputs are similar to the training samples.
Furthermore, even though unlabeled data can benefit novelty detection
\citet{scott09}, existing SSND methods for deep neural networks \citep{Kiryo17,
guo20, yujie2020, mcd_ood} cannot improve upon unsupervised methods on near OOD
data sets. Even methods that violate fundamental OOD detection assumptions by
using known test OOD data for hyperparameter tuning \citep{odin, mahalanobis,
mcd_ood} fail to work on challenging novelty detection tasks. Finally, large
pretrained models seem to solve near OOD detection \citep{fort2021}, but they
only work for extremely specific OOD data sets (see Section~\ref{sec:setting}
for details).

This situation naturally raises the following question:

\vspace{-0.4cm}
\begin{center}
  \emph{Can we improve semi-supervised novelty detection for neural
  networks}?
\end{center}
\vspace{-0.4cm}

In this paper, we introduce a new method that successfully leverages unlabeled
data to obtain diverse ensembles for novelty detection. Our
contributions are as follows:

\begin{itemize}[leftmargin=*]

  \item We propose to find Ensembles with Regularized Disagreement (ERD), that
    is, disagreement only on OOD data. Our algorithm produces ensembles just
    diverse enough to be used for novelty detection with a disagreement test
    statistic (Section~\ref{sec:method}).

  \item We prove that training with early stopping leads to regularized
    disagreement, for data that satisfies certain simplifying assumptions
    (Section~\ref{sec:earlystopping}).

  \item We show experimentally that $\method$ significantly outperforms existing
    methods on novelty detection tasks derived from standard image data sets, as
    well as on medical image benchmarks (Section~\ref{sec:experiments}).

\end{itemize}

\section{Proposed method}
\label{sec:method}

\vspace{-0.1cm}
\begin{figure*}[t] 
  \begin{minipage}[t]{0.50\linewidth}
    \removelatexerror
        \setlength{\algomargin}{-0.0cm}
        \begin{algorithm}[H]
        \DontPrintSemicolon
        \small
 
     \SetKwInOut{KwInput}{Input}

     \KwInput{Train set $\sourceset$, Validation set $\validset$, Unlabeled set
     $\targetset$, Model $\tilde{f}$ pretrained on $\sourceset$, Ensemble size $K$}

     \KwResult{$\method$ ensemble $\{\fhat_{y_i}\}_{i=1}^K$}

     Sample $K$ different labels $\{y_1, ..., y_K\}$ from $\YY$

     \For(\tcp*[h]{fine-tune $K$ models}){$c \gets \{ y_1, ..., y_K\}$} {
       $\fhat_c \gets \textit{Initialize}(\tilde{f})$\;
       $\labeledtarget \gets \{ (x, c) : x \in \targetset \}$\;
       $\fhat_c \gets \textit{EarlyStoppedFineTuning}\left(\fhat_c, \sourceset \cup \labeledtarget;
       \validset \right)$\;
     }

     \KwRet $\{\fhat_{y_i}\}_{i=1}^K$ \;

     \caption{Obtaining $\method$ ensemble via early stopping}
     \label{algo:reto_training}

    \end{algorithm}
  \end{minipage}
  \hspace{0.2cm}
  \begin{minipage}[t]{0.48\textwidth}
    \removelatexerror
        \setlength{\algomargin}{0.2cm}
        \begin{algorithm}[H]
        \DontPrintSemicolon
        \small

   \vspace{0.1cm}
     \SetKwInOut{KwInput}{Input}

     \KwInput{Ensemble $\{\fhat_{y_i}\}_{i=1}^K$, Test set $\testset$, $\outputs
       = \emptyset$, Threshold~$\thresh$,
   Disagreement metric $\dis$}

   \vspace{0.1cm}
     \KwResult{$\outputs$, i.e.\ the novel-class samples from $\testset$}
     \vspace{0.075in}

     \For(\tcp*[h]{run hypothesis test}){$x \in \testset$} {
     \vspace{0.075in}
         \If{$\Tdis(\fhat_{y_1}, ..., \fhat_{y_{K}})(x) > t_{0}$} {
         \vspace{0.05in}
          $\outputs \gets \outputs \cup \{x\}$\;
        }
     }
     \vspace{0.075in}
     \KwRet $\outputs$ \;

     \caption{\vspace{0.1cm}Novelty detection using $\method$\vspace{0.07cm}}
   \vspace{0.1cm}
     \label{algo:reto_detection}

    \end{algorithm}
  \end{minipage}

\end{figure*}

%
%
In this section we first introduce our proposed method to obtain Ensembles with
Regularized Disagreement ($\method$) and describe how they can be used for
novelty detection.

\subsection{Training ensembles with regularized disagreement ($\method$)}
\label{sec:RETOproc}

Recall from Figure~\ref{fig:practical_sketch} that we have access to both a
labeled training set $\sourceset = \{(x_i, y_i)\}_{i=1}^{n} \sim \idjoint$, with
covariates $x_i \in \idsupp$ and discrete labels $y_i \in \YY$, and an unlabeled
set $\targetset$, which contains both ID and unknown OOD samples. Moreover, we
initialize the models of the ensemble using the weights of a predictor with good
in-distribution performance, pretrained on $\sourceset$. In the scenarios we
consider, such a well-performing pretrained classifier is readily available, as
it solves Task~I in Figure~\ref{fig:practical_sketch}.
  
The entire training procedure is described in
Algorithm~\ref{algo:reto_training}.  For training a single model in the
ensemble, we assign a label $c \in \YY$ to all the unlabeled points in
$\targetset$, resulting in the $\targetlabel$-labeled set that we denote as
$\labeledtarget \defn \{(x,c): x\in \targetset\}$. We then fine-tune a
classifier $\fhat_c$ on the union $\sourceset \cup \labeledtarget$ of the
correctly-labeled training set $\sourceset$, and the unlabeled set
$\labeledtarget$.  In particular, we choose an early stopping time at which
validation accuracy is high and training error on $\sourceset \cup
\labeledtarget$ is low.
We create a diverse ensemble of $K$ classifiers $\fhat_c$ by choosing a
different artificial label $c \in \YY$ for every model.


Intuitively, encouraging each model in the ensemble to fit different labels to
the unlabeled set $\targetset$ promotes disagreement, as shown in
Figure~\ref{fig:setting_ensemble}.  In the next sections, we elaborate on how to
use diverse ensembles for novelty detection.

\vspace{-0.2cm}
\subsection{Ensemble disagreement for novelty detection}
\label{sec:disagreement}

We now discuss how we can use ensembles with disagreement to detect OOD
samples and why the right amount of diversity is crucial. Note that we
can cast the novelty detection problem as a hypothesis test with the null
hypothesis $H_0: x \in \suppid$.


As usual, we test the null hypothesis by comparing a test statistic with a
threshold $\thresh$: 
The null hypothesis is \emph{rejected} and we report $x$ as OOD
(\emph{positive}) if the test statistic is larger than $\thresh$
(Section~\ref{sec:erd_eval} elaborates on the choice of $\thresh$).
In particular, we use as test statistic the following disagreement score, which computes
the average distance between the softmax outputs of the $K$ models in the
ensemble:

\vspace{-0.5cm}
\begin{align*}
  \Tdis(f_1(x), ..., f_K(x)):=\frac{2\sum_{i\neq j} \dis \left(f_i(x),
  f_j(x)\right)}{K(K-1)},
\end{align*}

\vspace{-0.2cm}
\noindent where $\dis$ is a measure of disagreement between the softmax outputs
of two predictors, for example the total variation distance
$\dis_{\text{TV}}(f_i(x), f_j(x))=\frac{1}{2} \|f_i(x) - f_j(x) \|_1$ used in
our experiments\footnote{We also expect other distance metrics to be similarly
effective.}. We provide a thorough discussion on the soundness of this
test statistic for disagreeing models and compare it with previous metrics in
Appendix~\ref{sec:appendix_statistic}.

Even though previous work like \citet{mcd_ood} used a similar disagreement
score, their detection performance is notably worse. The reason lies in the lack
of diversity in their trained ensemble (see Figure~\ref{fig:scores_mcd_es} in
Appendix~\ref{sec:appendix_statistic}). On the other hand
Algorithm~\ref{algo:reto_training} without early stopping would lead to a too
diverse ensemble, that also disagrees on ID points, and hence, has a high false
positive rate (see Appendix~\ref{sec:appendix_score_curves}). In the next
section, we explain why novelty detection with this test statistic crucially
relies on the right amount of ensemble diversity and how ensembles may achieve
this goal if they are trained to have regularized disagreement.

\begin{figure*}[t]
  \centering

    \includegraphics[width=\textwidth]{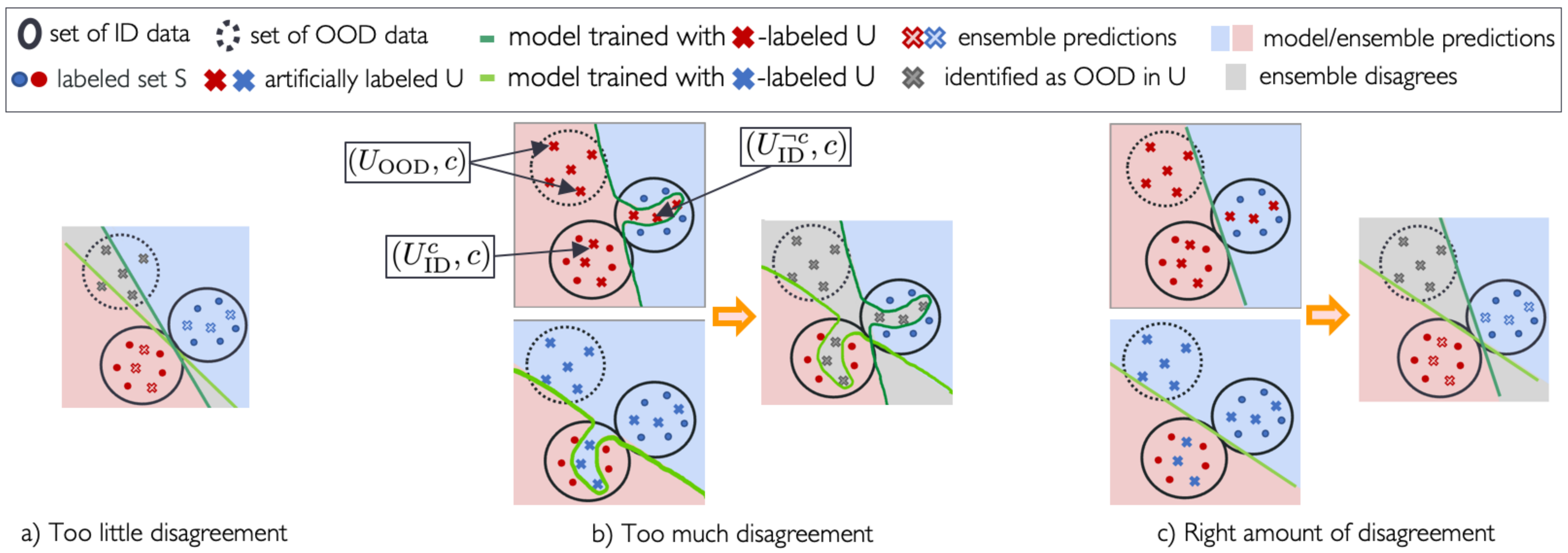}


    \caption{\small{a) Ensembles with too little disagreement fail to detect OOD
        samples. b) An ensemble of two models trained on $\sourceset \cup
        \labeledtarget$ disagrees on both ID and OOD data. b) Regularization
        prevents models from fitting $\wronglylabeledtargetid$,
    limiting disagreement to only OOD samples.}}

   \label{fig:disagreement_types}
 \end{figure*}

\vspace{-0.3cm}
\subsection{Desired ensemble diversity via regularized disagreement}
\label{sec:disagreement}
\vspace{-0.1cm}

For simplicity of illustration, let us first assume a training set with binary labels
and a semi-supervised novelty detection setting as depicted in
Figure~\ref{fig:setting_ensemble}~a).
For an ensemble with two models, like in Figure~\ref{fig:setting_ensemble}~b),
the model predictions \emph{agree} on the blue and red areas and \emph{disagree}
on the gray area depicted in Figure~\ref{fig:setting_ensemble}~c).  Note that
the two models in Figure~\ref{fig:setting_ensemble} are \emph{just diverse
enough} to obtain both high power (flag true OOD as OOD) and low false positive
rate (avoid flagging true ID as OOD) at the same time. 


Previous methods that try to leverage unlabeled data to obtain more
diverse ensembles either do not work with deep neural networks
\citep{Bennett2002,Zhang2010,Jain2020} or do not disagree enough on OOD data
\citep{mcd_ood}, leading to subpar novelty detection performance (see
Figure~\ref{fig:scores_mcd_es} in Appendix~\ref{sec:appendix_statistic}).

To obtain the right amount of diversity, it is crucial to train ensembles with
\emph{regularized disagreement} on the unlabeled set: The models
should disagree on the unlabeled OOD samples, but \emph{agree} on the unlabeled
ID points (Figure~\ref{fig:disagreement_types}c).
Thus, we avoid having too little disagreement as in
Figure~\ref{fig:disagreement_types}a), which results in low power, or too much
diversity, resulting in high false positive rate as in
Figure~\ref{fig:disagreement_types}b). In particular, if models $f_c$ predict
the correct label on ID points and the label $c$ on OOD data, we can effectively
use disagreement to detect novel-class samples.  Since classifiers with good ID
generalization need to be smooth, we expect the model predictions on holdout OOD
data from the same distributions to be in line with the predictions on the
unlabeled set.


In Section~\ref{sec:earlystopping} we argue that the training procedure in
Algorithm~\ref{algo:reto_training} successfully induces
\emph{regularized disagreement} and prove it in a synthetic setting.
Our experiments in Section~\ref{sec:experiments} further corroborate our
theoretical statements. Finally, we note that one could also use other
regularization techniques like dropout or weight decay. However, running a grid
search to select the right hyperparameters can be more computationally expensive
than simply using one run of the training process to select the optimal stopping
time.



%
%
%

\vspace{-0.5cm}
\section{Provable regularized disagreement via early stopping}
\label{sec:earlystopping}
\vspace{-0.2cm}

In this section, we show how using early stopping in
Algorithm~\ref{algo:reto_training} prevents fitting the incorrect artificial
label on the unlabeled ID samples.
Albeit for a simplified setting, this result provides a rigorous proof of
concept and intuition for why $\method$ ensembles achieve the right amount of
diversity necessary for good novelty detection.


\vspace{-0.2cm}
\subsection{Preliminary definitions}

We first introduce necessary definitions to prepare the mathematical
statement. Recall that in our approach, in addition to the correct
labels of the ID training set $\sourceset$, each member of the
ensemble tries to fit one label $\targetlabel$ to the entire unlabeled
set $\targetset$ that can be further partitioned into

\vspace{-0.7cm}
\begin{align*}
  \labeledtarget &= \labeledtargetid \cup \labeledtargetood \\
                 &= \{ (x, c) : x \in
  \targetidset \} \cup \{ (x, c) : x \in \targetoodset \},
\end{align*}
\vspace{-0.7cm}

where $\targetidset := \targetset \cap \idsupp$ and $\targetoodset := \targetset
\setminus \targetidset$. Moreover, assuming that the label of an ID input $x$ is
deterministically given by $\ystar(x)$,
we can partition the set $\labeledtargetid$ (see
Figure~\ref{fig:disagreement_types}b) into a subset of effectively ``correctly labeled'' samples $\correctlylabeledtargetid$ and ``incorrectly labeled'' samples $\wronglylabeledtargetid$:

\vspace{-0.7cm}
\begin{align*}
  \wronglylabeledtargetid := \{ (x, c) : x \in \targetidset \text{ with }
  \ystar(x) \neq \targetlabel \} \\
  \correctlylabeledtargetid := \{ (x, c) : x \in \targetidset \text{ with }
  \ystar(x) = \targetlabel \}.
\end{align*}
\vspace{-0.7cm}

Note that $\wronglylabeledtargetid$ can be viewed as the subset of noisy samples
from the entire training set $\sourceset \cup \labeledtarget$.


\vspace{-0.2cm}
\subsection{Main result}

We now prove that there exists indeed an optimal stopping time at which a
two-layer neural network trained with gradient descent does not fit the
incorrectly labeled subset $\wronglylabeledtargetid$, under mild distributional
assumptions.

\at{NEW: assumptions and statement} For the formal statement, we assume that the artificially labeled set
$\sourceset \cup \labeledtarget$ is \emph{clusterable}, i.e.\ the points
can be grouped in $K$ clusters of similar sizes. Each class may comprise several
clusters, but every cluster contains only samples from one class. Any cluster
may include at most a fraction $\rho \in [0, 1]$ of samples with label noise,
e.g.\ $\wronglylabeledtargetid$.  We denote by $c_1, ..., c_K$ the cluster
centers and define the matrix $C:=[c_1, ..., c_K]^T \in \RR^{K \times d}$.
Further, let $\separability$ be a measure of how well a two-layer neural network
can separate the cluster centers ($\separability=0$ if $c_i = c_j$ for some $i,
j \le K$). Under these assumptions we have the following:

\begin{proposition} (informal)
  \label{proposition_informal}
  It holds with high probability
  over the initialization of the weights that a two-layer neural network
  trained on $\sourceset \cup \labeledtarget$ perfectly fits $\sourceset$,
  $\correctlylabeledtargetid$ and $\labeledtargetood$, but not
  $\wronglylabeledtargetid$, after $T\simeq\frac{\|C\|^2}{\separability}$
  iterations.
\end{proposition}

\vspace{-0.2cm}
The precise assumptions for the proposition can be found in
Appendix~\ref{sec:appendix_theory}. On a high level, the reasoning follows from
two simple insights: 1. When the artificial label is not equal to the 
true label, the ID samples in the unlabeled set can be seen as
noisy samples in the set $S \cup (U,c)$.  2. It is well known that early
stopping prevents models from fitting incorrect labels since noisy samples with
incorrect labels are often fit later during training (see e.g.\ theoretical and
empirical evidence here \cite{Yilmaz2019,mahdi,song2020,liu2020}).  In
particular, our proof heavily relies on Theorem~2.2 of \citet{mahdi} which shows
that early stopped predictors are robust to label noise.


Proposition~\ref{proposition_informal} gives a flavor of the theoretical
guarantees that $\method$ enjoys.  Albeit simple, the clusterable data
model actually includes data with non-linear decision boundaries.  On the other
hand, the requirement that the clusters are balanced seems rather restrictive.
In our experiments we show that this condition is in fact more
stringent than it should.  In particular, our method still works when the
number of OOD samples $|\targetoodset|$ is considerably smaller than the number
of ID samples from any given class, as we show in Section~\ref{sec:ablations}.

\vspace{-0.3cm}
\subsection{Choosing the early stopping time}
\vspace{-0.3cm}

\begin{figure}[t]
  \begin{center}
    \includegraphics[width=0.7\columnwidth]{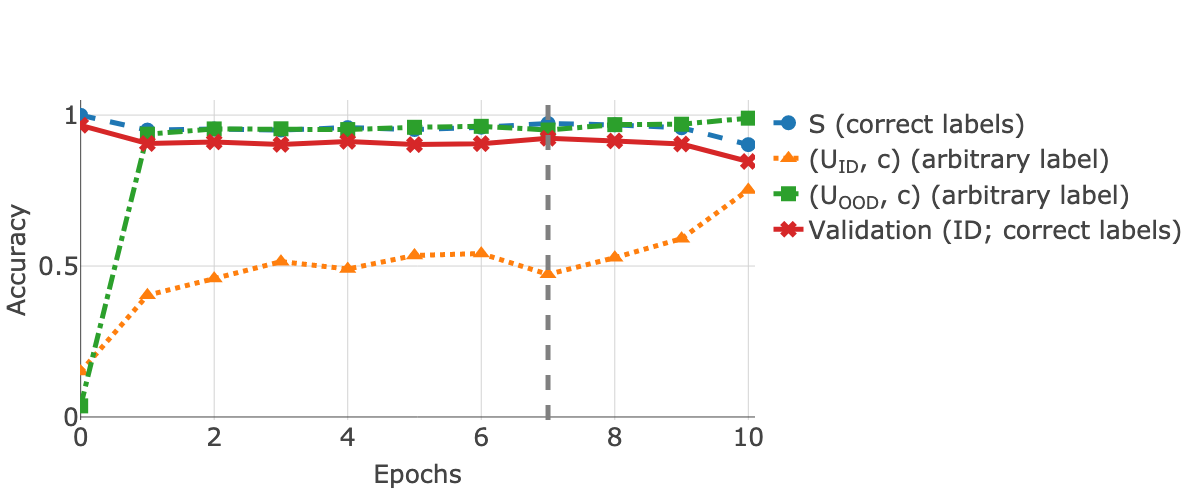}
  \end{center}

  \vspace{-0.4cm} \caption{ \small{Accuracy during fine-tuning a model
      pretrained on $\sourceset$ (epoch 0 indicates values obtained with the
      initial pretrained weights). The samples in $\labeledtargetood$ are fit
      first, while the model reaches high accuracy on $\labeledtargetid$ much
  later. We fine-tune for at least one epoch and then early stop when the
  validation accuracy starts decreasing after 7 epochs (vertical line). The
  model is trained on SVHN[0:4] as ID and SVHN[5:9] as OOD.  }}

  \label{fig:training_curves}
\end{figure}

In practice, we avoid computing the exact value of $T$ by using instead a
heuristic for picking the early stopping iteration with the highest validation
accuracy (indicated by the vertical line in Figure~\ref{fig:training_curves}).
As shown in the figure, the model fits the noisy training points, i.e.\
$\wronglylabeledtargetid$, late during fine-tuning, which causes the validation
accuracy to decrease, since the model will also predict the incorrect label
$\targetlabel$ on some validation ID samples. In
Appendix~\ref{sec:appendix_learning_curves} we show that the trend in
Figure~\ref{fig:training_curves} is consistent across data sets.

\vspace{-0.1cm}



\vspace{-0.2cm}
\section{Experimental results}
\label{sec:experiments}

In this section we evaluate the novelty detection performance of $\method$ with
deep neural networks on several image data sets. On difficult near OOD data
sets, we find that our approach outperforms all baselines, including SSND
methods, but also methods operating in other, sometimes more favorable settings.
In addition, we discuss some of the trade-offs that impact $\method$'s
performance.

\subsection{Data sets}


\at{NEW: added clarification re what's special about novelty detection} Our
experiments focus on novel-class detection scenarios where the ID and OOD data
share many similar features and only differ in a few characteristics.
We use standard image data sets (e.g.\ CIFAR10/CIFAR100) and consider half of
the classes as ID, and the other half as novel. We also assess $\method$'s
performance on a medical image benchmark \citep{Cao2020}, where near OOD data
consists of novel unseen diseases (e.g.\ X-rays of the same body part from
patients with different conditions; see Appendix~\ref{sec:appendix_datasets} for
details). Further, we also include far OOD data sets (e.g.\ CIFAR10/CIFAR100 vs
SVHN) for completeness.

For all scenarios, we used a labeled training set (e.g.\ 40K samples for
CIFAR10), a validation set with ID samples (e.g.\ 10K samples for CIFAR10) and
an unlabeled test set where half of the samples are ID and the other half are
OOD (e.g.\ 5K ID samples and 5K OOD samples for CIFAR10 vs SVHN). For
evaluation, we use a holdout set containing ID and OOD samples in the same
proportions as the unlabeled set. Moreover, in
Appendix~\ref{sec:appendix_small_test_set} we present results obtained with a
smaller unlabeled set of only 1K samples.

\begin{table*}[t]
\tiny
\centering

\caption{\small{AUROC and TNR@95 for \bestreto{$\method$} and various baselines
    (we highlight the \bestnonreto{best baseline}). Numbers in square
    brackets indicate the ID/OOD classes. 
    Asterisks mark methods proposed in this paper. Mahal, nnPU and MCD
($^\dagger$) use oracle information about the OOD data.
Repeated runs of $\method$ show a small variance $\sigma^2 < 0.01$ in
the detection metrics.
}}

\label{table:main_results}

\setlength{\tabcolsep}{1pt}
\hyphenpenalty10000
\begin{tabularx}{\textwidth} {@{}ll @{} @{\hskip 0.2cm} XXXXX  >{\centering\arraybackslash} XXXXX @{}} 
\toprule
& & \multicolumn{5}{c}{Other settings} & \multicolumn{5}{c}{SSND} \\ 
\cmidrule(l{2pt}r{2pt}){3-7}
\cmidrule(l{2pt}r{2pt}){8-12}
\makecell[l]{ ID data } & \makecell[l]{ OOD data } & \makecell[l]{ Vanilla\\Ensembles } & \makecell[l]{ Gram } & \makecell[l]{ DPN } & \makecell[l]{ OE } & \makecell[l]{ Mahal.$^\dagger$ } & \makecell[c]{ nnPU$^\dagger$} & \makecell[c]{ MCD$^\dagger$ } & \makecell[l]{ Mahal-U } & \makecell[l]{ Bin.\\Classif. * } & \makecell[l]{ $\method$ * }\\ 
\cmidrule(r){3-12}
& & \multicolumn{10}{c}{AUROC $\uparrow$ / TNR@95 $\uparrow$} \\
\midrule
\makecell[l]{ SVHN } & \makecell[l]{ CIFAR10 } & 0.97 / 0.88 & 0.97 / 0.86 & \bestnonreto{1.00} / \bestnonreto{1.00} & \bestnonreto{1.00} / \bestnonreto{1.00} & 0.99 / 0.98 & \bestnonreto{1.00} / \bestnonreto{1.00} & 0.97 / 0.85 & 0.99 / 0.95 & 1.00 / 1.00 & \bestreto{0.99} / \bestreto{0.98} \\
\makecell[l]{ CIFAR10 } & \makecell[l]{ SVHN } & 0.92 / 0.78 & \bestnonreto{1.00} / 0.98 & 0.95 / 0.85 & 0.97 / 0.89 & 0.99 / 0.96 & \bestnonreto{1.00} / \bestnonreto{1.00} & \bestnonreto{1.00} / 0.98 & 0.99 / 0.96 & 1.00 / 1.00 & \bestreto{1.00} / \bestreto{1.00} \\
\makecell[l]{ CIFAR100 } & \makecell[l]{ SVHN } & 0.84 / 0.48 & 0.99 / 0.97 & 0.77 / 0.44 & 0.82 / 0.50 & 0.98 / 0.90 & \bestnonreto{1.00} / \bestnonreto{1.00} & 0.97 / 0.73 & 0.98 / 0.92 & 1.00 / 1.00 & \bestreto{1.00} / \bestreto{1.00} \\

\midrule
\makecell[l]{ SVHN\\{[}0:4{]} } & \makecell[l]{ SVHN\\{[}5:9{]} } & 0.92 / 0.69 & 0.81 / 0.31 & 0.87 / 0.19 & 0.85 / 0.52 & 0.92 / 0.71 & \bestnonreto{0.96} / \bestnonreto{0.73} & 0.91 / 0.51 & 0.91 / 0.63 & 0.81 / 0.40 & \bestreto{0.95} / \bestreto{0.73} \\
\makecell[l]{ CIFAR10\\{[}0:4{]} } & \makecell[l]{ CIFAR10\\{[}5:9{]} } & 0.80 / 0.39 & 0.67 / 0.15 & \bestnonreto{0.82} / 0.32 & \bestnonreto{0.82} / \bestnonreto{0.41} & 0.79 / 0.27 & 0.61 / 0.11 & 0.69 / 0.25 & 0.64 / 0.13 & 0.85 / 0.43 & \bestreto{0.89} / \bestreto{0.57} \\
\makecell[l]{ CIFAR100\\{[}0:49{]} } & \makecell[l]{ CIFAR100\\{[}50:99{]} } & \bestnonreto{0.78} / \bestnonreto{0.35} & 0.71 / 0.16 & 0.70 / 0.26 & 0.74 / 0.31 & 0.72 / 0.20 & 0.53 / 0.06 & 0.70 / 0.26 & 0.72 / 0.19 & 0.66 / 0.13 & \bestreto{0.81} / \bestreto{0.41} \\

\bottomrule
\end{tabularx}
\vspace{0.7cm}

\end{table*}

\subsection{Baselines}

We compare our method against a wide range of
baselines that are applicable in the SSND setting.

\paragraph{Semi-supervised novelty detection.} We primarily compare $\method$ to
SSND approaches that are designed to incorporate a small set of unlabeled ID and
novel samples.

The \emph{MCD} method \citep{mcd_ood} trains an ensemble of two classifiers such
that 
one model gives high-entropy and the other yields low entropy predictive
distributions on the unlabeled samples. Furthermore, \emph{nnPU} \citep{Kiryo17}
considers a binary classification setting, in which the labeled data comes from
one class (i.e.\ ID samples, in our case), while the unlabeled set contains a
mixture of samples from both classes. Notably, both methods require oracle
knowledge that is usually unknown in the regular SSND setting: MCD uses test OOD
data for hyperparameter tuning while nnPU requires oracle knowledge of the ratio
of OOD samples in the unlabeled set.

In addition to these baselines, we also propose two natural extensions to the
SSND setting of two existing methods.  Firstly, we present a version of the
Mahalanobis approach (\emph{Mahal-U}) that is calibrated using the unlabeled
set, instead of using oracle OOD data. Secondly, since nnPU requires access to
the OOD ratio of the unlabeled set, we also consider a less burdensome
alternative: a \emph{binary classifier} trained to separate the training data
from the unlabeled set and regularized with early stopping like our method.

\paragraph{Unsupervised novelty detection (UND).} Naturally, one may ignore the
unlabeled data and use UND approaches. 
The current SOTA UND method on the usual benchmarks is the \emph{Gram method}
\citep{gram_ood}. Other UND approaches include \emph{vanilla ensembles}
\citep{balaji}, deep generative models (which tend to give undesirable results
for OOD detection \citep{Kirichenko20}), or various Bayesian approaches (which
are often poorly calibrated on OOD data \citep{ood_ovadia}).

Preliminary analyses revealed that generative models and methods trained with a
contrastive loss \citep{winkens2020} or with one-class classification
\citep{sohn2021} perform poorly on near OOD data sets (see
Appendix~\ref{sec:appendix_cifar10_cifar100} for a comparison; we use numbers
reported by the authors for works where we could not replicate their results).


\paragraph{Other methods.} We also compare with \emph{Outlier Exposure}
\citep{outlier_exposure} and \emph{Deep Prior Networks (DPN)} \citep{dpn} which use
TinyImages as known outliers during training, irrespective of the OOD set used
for evaluation. On the other hand, the \emph{Mahalanobis} baseline \citep{mahalanobis}
is tuned on samples from the same OOD distribution used for evaluation.
Finally, we also consider large transformer models pretrained on ImageNet21k and
fine-tuned on the ID training set \citep{fort2021}.

\subsection{Implementation details}
\label{sec:erd_eval}

\paragraph{Baseline hyperparameters.}
For all the baselines, we use the default hyperparameters suggested by their
authors on the respective ID data set (see
Appendix~\ref{sec:appendix_experiments} for more details). For the binary
classifier, nnPU, ViT, and vanilla ensembles, we choose the hyperparameters that
optimize the loss on an ID validation set.

\paragraph{$\method$ details.}\footnote{Our code is publicly available at
\href{https://bit.ly/3a7aQyN}{https://github.com/ericpts/reto}.}
We follow the procedure in Algorithm~\ref{algo:reto_training} to fine-tune each
model in the $\method$ ensemble starting from weights that are pretrained on the
labeled ID set $\sourceset$.\footnote{In the appendix we also train the models
  from random initializations, i.e.\ $\method$++, and obtain better novelty
detection at the cost of more training iterations.}
Unless otherwise specified, we train $K=3$ ResNet20 models \citep{He2015} using
3 randomly chosen class labels for $\labeledtarget$ and note that even ensembles
of two models produce good results (see
Appendix~\ref{sec:appendix_ensemble_size}). We stress that whenever applicable,
our choices disadvantage $\method$ for the comparison with the baselines, e.g.\
vanilla ensembles use $K=5$, and for most of the other approaches we use the
larger WideResNet-28-10. 
We select the early stopping time and other standard hyperparameters so as to
maximize validation accuracy.

\paragraph{Evaluation.} As in standard hypothesis testing, choosing different
thresholds for rejecting the null hypothesis leads to different false positive
and true positive rates (FPR and TPR, respectively). The ROC curve follows the
FPR and TPR for all possible threshold values and the area under the curve
(AUROC; larger values are better) captures the performance of a statistical test
without having to select a specific threshold. In addition, we also report the
TNR at a TPR of 95\% (TNR@95; larger values are better).\footnote{In practice,
  choosing a good rejection threshold is important.  A recent work \citep{Liu18}
  proposes a criterion for selecting the threshold that is tailored specifically
  to the SSND setting.  Alternatively, one can choose the threshold so as to
achieve a desired FPR, which we can estimate using a validation set of ID
samples.}

\paragraph{Computation cost.}
We only need to fine-tune two-model ensembles to get good performance with
$\method$ (see Appendix~\ref{sec:appendix_ensemble_size}). For instance, in
applications like the one in Figure~\ref{fig:practical_sketch}, $\method$
fine-tuning introduces little overhead and works well even with scarce resources
(e.g.\ it takes around 5 minutes on 2 GPUs for the settings in
Table~\ref{table:main_results}).
In contrast, other ensemble diversification methods require training different
models for each hyperparameter choice and have training losses that cannot be
easily parallelized (e.g.\ \citet{mcd_ood}). Moreover, the only other approach
that achieves comparable performance to our method on \emph{some} near OOD data 
uses large transformer models pretrained on a large and conveniently chosen
data set \citep{fort2021}.

\vspace{-0.2cm}
\subsection{Main results}
\vspace{-0.1cm}

\at{NEW: added comments on ViT results} We summarize the main empirical results
in Table~\ref{table:main_results}.  While most methods achieve near-perfect
detection for far OOD, $\method$ has a clear edge over the baselines for
novel-class detection within the same dataset -- even compared to methods
($\dagger$) that use oracle OOD information. For completeness, we present in
Appendix~\ref{sec:appendix_cifar10_cifar100} a comparison with more related
works.  These methods either show unsatisfactory performance on near OOD tasks,
or seem to work well only on certain specific data sets.
We elaborate on the potential causes of failure for these works in
Section~\ref{sec:setting}.



\begin{figure*}[t]
  \begin{subfigure}[t]{0.49\textwidth}
    \centering
    \includegraphics[width=0.8\textwidth]{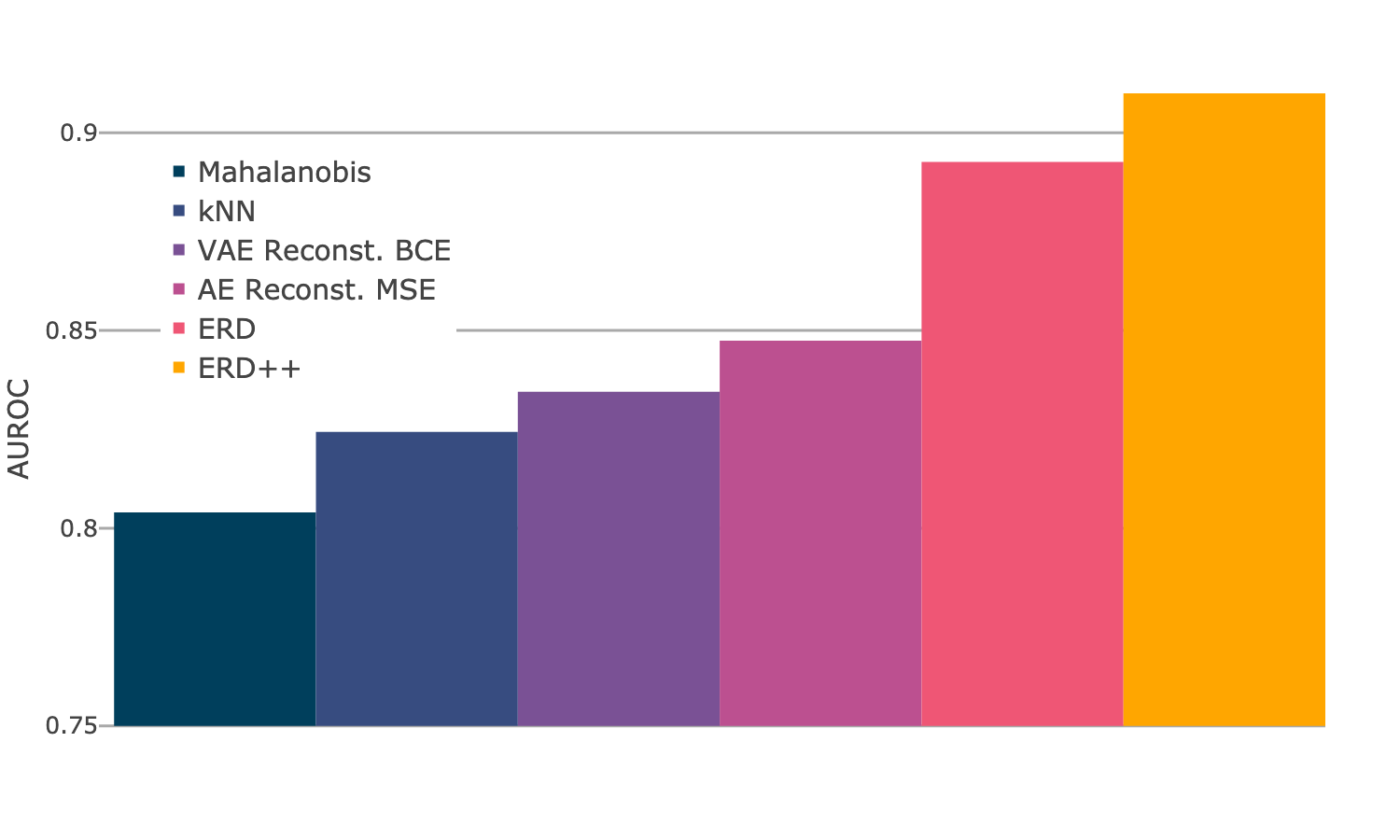}
    \caption{Novelty detection performance on medical data}
    \label{fig:avg_medical_ood_main_text}
  \end{subfigure}
  \hfill
  \begin{subfigure}[t]{0.49\textwidth}
    \centering
    \includegraphics[width=\textwidth]{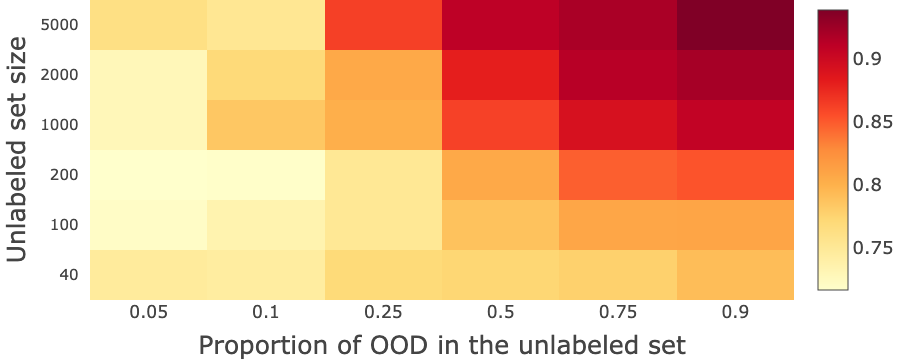}
    \caption{Effect of OOD proportion on detection}
    \label{fig:vary_target_main}
  \end{subfigure}

\caption{\small{\textbf{Left:} AUROC averaged over all scenarios
    in the medical novelty detection benchmark. The values for the baselines are
    computed using the code from \citet{Cao2020}. \textbf{Right:} The AUROC of a
    3-model $\method$ ensemble as the number and proportion of ID (CIFAR10[0:4])
and OOD (CIFAR10[5:9]) samples in the unlabeled set are varied (see also
Appendix~\ref{sec:appendix_vary_ood_ratio}).}}

\end{figure*}

For the medical novelty detection benchmark we show in
Figure~\ref{fig:avg_medical_ood_main_text} the average AUROC achieved by some
representative baselines taken from \citet{Cao2020}. Our method improves the
average AUROC from $0.85$ to $0.91$, compared to the best baseline.
We refer the reader to \citet{Cao2020} for precise details on the methods.
Appendix~\ref{sec:appendix_medical} contains more results, as well as additional baselines.


\vspace{-0.2cm}
\subsection{Ablation studies and limitations}
\label{sec:ablations}
\vspace{-0.1cm}

We also perform extensive experiments to understand the importance of specific
design choices and hyperparameters, and refer the reader to the appendix for
details.

\at{NEW: moved here all discussion of different OOD and cov shift}
\paragraph{Relaxing assumptions on OOD samples.} In
Table~\ref{table:main_results} we evaluate our approach on a holdout test set
that is drawn from the same distribution as the unlabeled set $\targetset$ used
for fine-tuning. However, we provide experiments in
Appendix~\ref{sec:appendix_different_ood} that show that novelty detection with
$\method$ continues to perform well even when the test set and $\targetset$ come
from different distributions (e.g.\ novel-class data in the test set also
suffers from corruptions).
Further, even though our main focus is novel-class detection, our experiments
(Appendix~\ref{sec:appendix_cov_shift}) indicate that $\method$ can also
successfully identify near OOD samples that suffer from only mild covariate
shift compared to the ID data (e.g. CIFAR10 vs corrupted CIFAR10 \citep{cifar_c}
or CIFAR10v2 \citep{recht}).  Finally,
Appendix~\ref{sec:appendix_transductive_results} shows that $\method$ ensembles
also perform well in a transductive setting \citep{scott08}, where the test set
coincides with $\targetset$.

\paragraph{Relaxing the assumptions of Proposition~\ref{proposition_informal}.}
Our theoretical results require that the ID classes in the training set and
the novel classes in $\targetset$ have similar cardinality. In fact, this
condition is unnecessarily strong as we show in our empirical analysis: In all
experimental settings we have significantly fewer OOD than ID training points.
%
We further investigate the impact of the size of the unlabeled set and of the
ratio of novel samples in it ($\frac{|\targetoodset|}{|\targetidset| +
|\targetoodset|}$) and find that $\method$ in fact maintains good performance
for a broad range of ratios in Figure~\ref{fig:vary_target_main}.

\paragraph{Sensitivity to hyperparameter choices.} We point out that $\method$
ensembles are particularly robust to changes in the hyperparameters like batch
size or learning rate (Appendix~\ref{sec:appendix_lr_bs}), or the choice of the
arbitrary label assigned to the unlabeled set
(Appendix~\ref{sec:appendix_ensemble_size}).  Further, we note that $\method$
ensembles with as few as two models already show remarkable novelty detection
performance and refer to Appendix~\ref{sec:appendix_ensemble_size} for
experiments with larger ensemble sizes.
Moreover, $\method$ performance improves with larger neural networks
(Appendix~\ref{sec:appendix_different_arch}), meaning that $\method$ will
benefit from any future advances in architecture design. 

\paragraph{Choice of disagreement score.} We show in
Table~\ref{table:score_comparison} in Appendix~\ref{sec:appendix_statistic},
that the training procedure alone (Algorithm~\ref{algo:reto_training}) does not
suffice for good novelty detection. For optimal results, $\method$ ensembles
need to be combined with a disagreement-based score like the one introduced in
Section~\ref{sec:disagreement}.
Finally, we show how the distribution of the disagreement score changes during
training for $\method$ (Appendix~\ref{sec:appendix_score_curves}) and explain
why regularizing disagreement is more challenging for near OOD data, compared to
easier, far OOD settings (Appendix~\ref{sec:appendix_learning_curves}).

\at{NEW: 1) SSND for online; 2) anomaly detection; 3) near OOD is more
  challenging for ERD too}
\paragraph{Limitations.} Despite the advantages of $\method$, like all prior SSND methods,
our approach
is not a good fit for online (real-time) novelty detection tasks.  Moreover,
$\method$ ensembles are not tailored to anomaly detection, where outliers are
particularly rare, since the unlabeled set should contain at least a small
number of samples from the novel classes (see
Figure~\ref{fig:vary_target_main} and Appendix~\ref{sec:appendix_vary_ood_ratio}). However, $\method$ ensembles are an ideal
candidate for applications that require highly accurate, offline novelty
detection, like the one illustrated in Figure~\ref{fig:practical_sketch}.

\section{Related work}
\label{sec:setting}

\begin{table*}[t]
  \tiny
  \centering

  \caption{Taxonomy of novelty detection methods, categorized according to data
    availability (\textbf{horizontal axis}) and probabilistic perspective
    (\textbf{vertical axis}). We
  \ensemble{highlight} the ensemble-based methods.}
    \label{table:taxonomy}

\newcommand{\taxonomyrow}[7]{#1 & #7 & #6 & #4 & #5 & #3 & #2 \\}

\begin{tabularx}{\textwidth}{@{} smhmMsm @{}}
\toprule
\taxonomyrow{ }{SND}{P-UND}{\makecell[l]{Different OOD\\ A-UND}}{\makecell[l]{Synthetic OOD\\ A-UND}}{SSND}{UND}
\midrule
\taxonomyrow{Learn $\iddist$}{[GKRB13, DKT19, RVGBM20]}{[RH21]}{ }{OC classif.\
  [SLYJP21, TMJS20]}{\impl{nnPU [KNPS17]}}{Generative e.g.\ [AAB18], OC classif.\
e.g.\ [SPSSW01]}
\midrule
\taxonomyrow{\makecell[l]{Learn $\iddist$\\using $y$}}{\impl{Mahala.\ [LLLS18]},
  \impl{\ensemble{MCD [YA19]}}}{ViT [FRL20]}{ }{Data augmentation for contrastive loss
[TMJS20, LA20]}{\impl{\ensemble{ERD (Ours)}}, SSND for shallow models [MBGBM10, BLS10], U-LAC
[DYZ14, ZZMZ20]}{\impl{Gram
[SO19]}, OpenHybrid [ZLGG20]}
\midrule
\taxonomyrow{\makecell[l]{Uncertainty\\of $P_{Y|X}$}}{ODIN [LLS18]}{ }{\impl{DPN
  [MG18]}, \impl{OE [HMD19]}}{GAN outputs [LLLS18], noise
[HTLI19] or uniform samples (\ensemble{[JLMG20]})}{---}{Bayesian methods e.g.\
[GG16], \impl{\ensemble{Vanilla Ensemble [LPB17]}}}
\bottomrule
\end{tabularx}

\vspace{0.9cm}

\end{table*}

In this section, we present an overview of different types of
related methods that are in principle applicable for solving semi-supervised
novelty detection.  In particular, we indicate caveats of these methods based on
their categorization with respect to 1) data availability and 2) the surrogate objective
they try to optimize. This taxonomy may also be of independent interest to
navigate the zoo of ND methods. We list a few representative approaches in
Table~\ref{table:taxonomy} and refer the reader to surveys such as
\citet{Bulusu2020} for a thorough literature overview.

\vspace{-0.2cm}
\subsection{Taxonomy according to data availability}



In this section we present related novelty detection methods that use varying
degrees of labeled OOD data for training.
We call \emph{test OOD} the novel-class data that we want to detect at test
time.

%

In a scenario like the one in Figure~\ref{fig:practical_sketch},
one can apply \boldemph{unsupervised novelty detection (UND)} methods
that ignore the unlabeled batch and only uses ID data during training
\citep{balaji, gram_ood, nalisnick}. However, these approaches lead to
poor novelty detection performance, especially on near OOD data.

There are methods that suggest to improve UND performance by using additional
data.
For example, during training one may use synthetically generated outliers (e.g.\
\citet{Tack2020, sohn2021}) or a different OOD data set that may be available (e.g.\ OE
and DPN use TinyImages) with samples
\emph{known to be outliers}. 
However, in order for these \boldemph{augmented unsupervised ND (A-UND)} methods
to work, they require that the OOD data used for training is similar to test OOD
samples. When this condition is not satisfied, A-UND performance deteriorates
drastically (see Table~\ref{table:main_results}). However, by definition, novel
data is unknown and the only information about the OOD data that is
realistically available is in the unlabeled set like in SSND. Therefore, it is
unknown what an appropriate choice of the training OOD data is for A-UND
methods.

\at{NEW: position vs ViT} Another line of work uses pretrained models to
incorporate additional data that is close to test OOD samples, i.e.\
\boldemph{pretrained UND (P-UND)}.
\citet{fort2021} use large transformer models pretrained on ImageNet21k and
achieve good near OOD detection performance when ID and OOD data are similar to
ImageNet samples (e.g.\ CIFAR10/CIFAR100). However, our experiments in
Appendix~\ref{sec:appendix_vit} reveal that this method performs poorly on all
other near OOD data sets, including unseen FashionMNIST or SVHN classes and
X-rays of unknown diseases.  This unsatisfactory performance is apparent when ID
and OOD data
do not share visual features with the pretraining data (i.e.\ ImageNet21k).
Since collecting such large troves of ``similar'' data for pre-training is often
not possible in practical applications (as medical imaging), the use case of
their method is rather limited.


Furthermore, a few popular methods use test OOD data for calibration or
hyperparameter tuning \citep{mcd_ood, mahalanobis, odin, Ruff2020}, which is not
applicable in practice. Clearly, knowing the test OOD distribution a priori
turns the problem into \boldemph{supervised ND (SND)}, and hence, violates the
fundamental assumption that OOD data is unforeseeable. 

As we have already seen, current \boldemph{SSND} approaches (e.g.\ MCD, nnPU)
perform poorly for complex models such as neural networks. We note that SSND is
similar to using unlabeled data for learning with augmented classes (U-LAC)
\citep{Da2014, guo20, yujie2020} and is related to transductive novelty
detection \citep{scott08, guo20}, where the test set coincides with the
unlabeled set used for training.

\subsection{Taxonomy according to probabilistic perspective}

Apart from data availability, the methods that we can use in a practical SSND
scenario implicitly or explicitly use a different principle based on a
probabilistic model.  For example, novel-class samples are a subset of the
points that are out-of-distribution in the literal sense, i.e.\ $\iddist(x)
< \alpha$. One can hence \textbf{learn $\iddist$} from unlabeled ID data, which
is however notoriously difficult in high dimensions.

Similarly, from a Bayesian viewpoint, the predictive variance
is larger for OOD samples with $\iddist(x)<\alpha$. Hence, one could instead
compute the posterior $\iddist (y|x)$ and flag points with large variance (i.e.\
high \textbf{predictive uncertainty}). This circumvents the problem with
estimating $\iddist$. However, Bayesian estimates of uncertainty that accompany
NN predictions tend to not be accurate on OOD data \citep{ood_ovadia}, resulting in poor novelty detection performance.

When the labels are available for the training set, we can instead partially
\textbf{learn $\iddist$ using $y$}. For instance, one could use generative
modeling to estimate the set of $x$ for which $\iddist(x)>\alpha$ via
$\iddist(x|y)$ \cite{mahalanobis, gram_ood}.
Alternatively, given a loss and function space, we may use the labels
indirectly, like in ERD, and use properties of the approximated population error
that imply small or large $\iddist$. 

\section{Conclusion}
\label{sec:conclusion}

In summary, we propose an SSND procedure that exploits unlabeled data
effectively to generate an ensemble with \emph{regularized} disagreement, which
achieves remarkable novelty detection performance. Our SSND method
does not need labeled OOD data during training unlike many other
related works summarized in Table~\ref{table:taxonomy}. 

We leave as future work a thorough investigation of the impact of the
labeling scheme of the unlabeled set on the sample complexity of the method, as
well as an analysis of the trade-off governed by the complexity of the model
class.



\newpage

\subsubsection*{Acknowledgments} 

We are grateful to Călin Cruceru, Gideon Dresdner, Alexander Immer, Sidak Pal
Singh and Armeen Taeb for feedback on the manuscript and to Ayush Garg for
preliminary experiments. We also thank the anonymous reviewers for their helpful
remarks.

\bibliography{arxiv_paper}
\bibliographystyle{abbrv}

\newpage

\appendix
\section{Theoretical statements}
\label{sec:appendix_theory}

\begin{definition}[$(\eps, \rho)$-clusterable data set]

  We say that a data set $\DD=\{ (x_i, y_i) \}_{i=1}^n$ is \emph{($\eps,
  \rho$)-clusterable} for fixed $\eps > 0$ and $\rho \in [0, 1]$ if there exists
  a partitioning of it into subsets $\{ C_1, ..., C_K \},$ which we call
  \emph{clusters}, each with their associated unit-norm cluster center $c_i$,
  that satisfy the following conditions:


  \begin{itemize}[leftmargin=*]

    \item $\bigcup_{i=1}^K C_i = \DD$ and $C_i \cap C_j = \emptyset, \forall i,
      j \in [K]$; 

    \item all the points in a cluster lie in the $\eps$-neighborhood of their
      corresponding cluster center, i.e.\ $||x - c_i||_2 \le \eps$ for all $x
      \in C_i$ and all $i \in [K]$;

    \item a fraction of at least $1-\rho$ of the points in each cluster $C_i$
      have the same label, which we call the \emph{cluster label} and denote
      $y^*(c_i)$. The remaining points suffer from label noise;

    \item if two cluster $C_i$ and $C_j$ have different labels, then their
      centers are $2\eps$ far from each other, i.e.\ $||c_i - c_j||_2 \ge
      2\eps$;

    \item the clusters are balanced i.e.\ for all $i \in [K], \alpha_1
      \frac{n}{K} \le |C_i| \le \alpha_2 \frac{n}{K}$, where $\alpha_1$ and
      $\alpha_2$ are two positive constants.

  \end{itemize}

\end{definition}

In our case, for a fixed label $\targetlabel\in\YY$, we assume that the set
$\sourceset \cup \labeledtarget$ is $(\eps, \rho)$-clusterable into $K$
clusters. We further assume that each cluster $C_i$ only includes a few noisy
samples from $\wronglylabeledtargetid$, i.e.\ $\frac{|C_i
\cap\wronglylabeledtargetid|}{|C_i|} \le \rho$ and that for clusters $C_i$ whose
cluster label is not $\targetlabel$, i.e.\ $y^*(c_i) \neq \targetlabel$, it
holds that $C_i \cap \labeledtargetood = \emptyset$. 

We define the matrices $C:=[c_1, ..., c_K]^T \in \RR^{K \times d}$ and
$\Sigma:=(CC^T)\bigodot \EE_g[\phi'(Cg)\phi'(Cg)^T]$, with $g \sim \gauss(0,
I_d)$ and where $\bigodot$ denotes the elementwise product. We use $\| \cdot \|$
and $\lambda_{min}(\cdot)$ to denote the spectral norm and the smallest
eigenvalue of a matrix, respectively.

For prediction, we consider a 2-layer neural network model with $p$ hidden
units, where $p \gtrsim \frac{K^2\|C\|^4}{\lambda_{min}(\Sigma)^4}$. We can
write this model as follows:

\begin{align}
  x \mapsto f(x; W) = v^T\phi(Wx),
\end{align}

The first layer weights $W$ are initialized with random values drawn from
$\gauss(0, 1)$, while the last layer weights $v$ have fixed values: half of them
are set to $1/p$ and the other half is $-1/p$. We consider activation functions
$\phi$ with bounded first and second order derivatives, i.e.\ $|\phi'(x)|\le
\Gamma$ and $\phi''(x)\le \Gamma$.
We use the squared loss for training, i.e.\ $\LL(W)=\frac{1}{2}\sum_{i=0}^n (y_i
- f(x_i; W))^2$ and take gradient descent steps to find the optimum of the loss
function, i.e.\ $W_{\tau+1} = W_\tau - \eta \nabla \LL(W_\tau)$, where the step
size is set to $\eta \simeq \frac{K}{n\|C\|^2}$.

We can now state the following proposition:

\begin{proposition}
  \label{proposition_appendix}

  Assume that $\rho \le \delta / 8$ and $\eps \le \alpha \delta
  \lambda_{min}(\Sigma)^2 / K^2$, where $\delta$ is a constant such that $\delta
  \le \frac{2}{|\YY - 1|}$ and $\alpha$ is a constant that depends on $\Gamma$.
  Then it holds with high probability $1 - 3 / K^{100} - Ke^{-100d}$ over the
  initialization of the weights that the neural network trained on $\sourceset
  \cup \labeledtarget$ perfectly fits $\sourceset$, $\correctlylabeledtargetid$
  and $\labeledtargetood$, but not $\wronglylabeledtargetid$, after
  $T=c_4\frac{\|C\|^2}{\lambda_{min}(\Sigma)}$ iterations.

\end{proposition}

This result shows that there exists an optimal stopping time at which the neural
network predicts the correct label on all ID points and the label $\targetlabel$
on all the OOD points. As we will see later in the proof, the proposition is
derived from a more general result which shows that the early stopped model
predicts these labels not only on the points in $\targetset$ but also in an
$\eps$-neighborhood around cluster centers. Hence, an $\method$ ensemble can be
used to detect holdout OOD samples similar to the ones in $\targetset$, after
being tuned on $\targetset$. This follows the intuition that classifiers
regularized with early stopping are smooth and generalize well.

The clusterable data model is generic enough to include data sets with
non-linear decision boundaries. Moreover, notice that the condition in
Proposition~\ref{proposition_appendix} is satisfied when $\sourceset \cup
\labeledtargetid$ is $(\eps, \rho)$-clusterable and $\labeledtargetood$ is
$\eps$-clusterable and if the cluster centers of $\labeledtargetood$ are at
distance at least $2\eps$ from the cluster centers of $\sourceset \cup
\labeledtargetid$. A situation in which these requirements are met is, for
instance, when the OOD data comes from novel classes, when all classes
(including the unseen ones that are not in the training set) are well separated,
with cluster centers at least $2\eps$ away in Euclidean distance. In addition,
in order to limit the amount of label noise in each cluster, it is necessary
that the number of incorrectly labeled samples in $\wronglylabeledtargetid$ is
small, relative to the size of $\sourceset$.

In practice, we only need that the decision boundary separating
$\labeledtargetood$ from $\sourceset$ is easier to learn than the classifier
required to interpolate the incorrectly labeled $\wronglylabeledtargetid$, which
is often the case, provided that $\labeledtargetood$ is large enough and the OOD
samples come from novel classes.

We now provide the proof for Proposition~\ref{proposition_appendix}:

\begin{proof}

  We begin by restating a result from \citet{mahdi}:

\begin{theorem}[\citep{mahdi}]
  \label{thm:mahdi}

  Let $\DD:=\{(x_i, y_i)\}\in \RR^d \times \YY$ be an $(\eps, \rho)$-clusterable
  training set, with $\eps \le c_1 \delta \lambda_{min}(\Sigma)^2 / K^2$ and
  $\rho \le \delta / 8$, where $\delta$ is a constant that satisfies $\delta \le
  \frac{2}{|\YY|-1}$. Consider a two-layer neural network as described above,
  and train it with gradient descent starting from initial weights sampled
  i.i.d.\ from $\gauss(0, 1)$. Assume further that the step size is $\eta =
  c_2\frac{K}{n\|C\|^2}$ and that the number of hidden units $p$ is at least
  $c_3 \frac{K^2\|C\|^4}{\lambda_{min}(\Sigma)^4}$. Under these conditions, it
  holds with probability at least $1 - 3 / K^{100} - Ke^{-100d}$ over the random
  draws of the initial weights, that after
  $T=c_4\frac{\|C\|^2}{\lambda_{min}(\Sigma)}$ gradient descent steps, the
  neural network $x\mapsto f(x; W_T)$ predicts the correct cluster label for all
  points in the $\eps$-neighborhood of the cluster center, namely:

  \begin{align}
    \arg\max_{y \in \YY} | f(x; W_T) - \onehot(y) |= y^*(c_i), \text{ for all } x \text{ with }
    \| x - c_i \|_2 \le \eps \text{ and all clusters } i \in [K],
  \end{align}

  where $\onehot: \YY \to \{0, 1\}^{|\YY|}$ yields one-hot embeddings of the
  labels. The constants $c_1, c_2, c_3, c_4$ depend only on $\Gamma$.

\end{theorem}

Notice that, under the assumptions introduced above, the set $\sourceset \cup
\labeledtarget$ is $(\eps, \rho)$-clusterable, since the incorrectly labeled ID
points in $\wronglylabeledtargetid$ constitute at most a fraction $\rho$ of the
clusters they belong to. As a consequence,
Proposition~\ref{proposition_appendix} follows directly from
Theorem~\ref{thm:mahdi}.

\end{proof}



\vspace{-0.5cm}
\section{Disagreement score for novelty detection}
\label{sec:appendix_statistic}

\begin{figure}[t]
  \centering
  \includegraphics[width=0.8\textwidth]{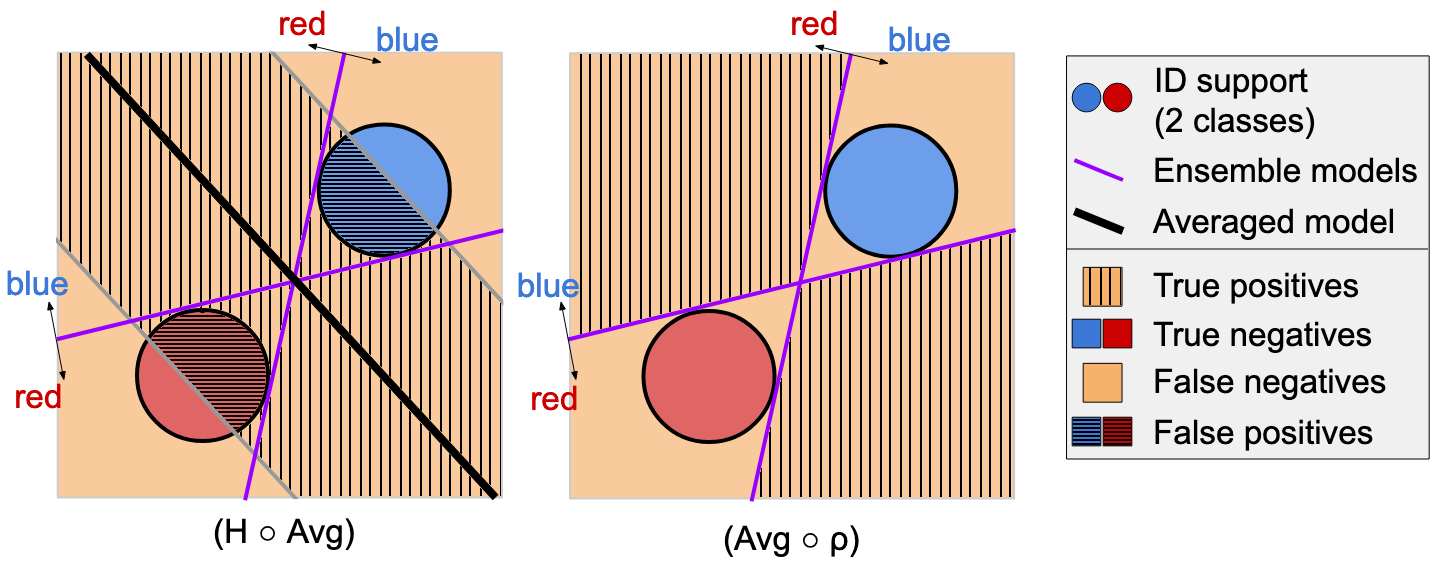}

  \caption{\small{Cartoon illustration showing a diverse ensemble of linear
      binary classifiers. We compare novelty detection performance for two
      aggregation scores: $\entavg$ (\textbf{Left}) and $\Tdis$ with
      $\rho(f_1(x), f_2(x))=\mathbb{1}_{\sgn(f_1(x))\neq\sgn(f_2(x))}$
      (\textbf{Right}). The two metrics achieve similar TPRs, but using
      $\entavg$ instead of our score, $\Tdis$, leads to more false positives,
      since the former simply flags as OOD a band around the averaged model
      (solid black line) and does not take advantage of the ensemble's
  diversity.}}

  \label{fig:ensemble_disagreement}
\end{figure}

As we argue in Section~\ref{sec:earlystopping},
Algorithm~\ref{algo:reto_training} produces an ensemble that disagrees on OOD
data, and hence, we want to devise a scalar score that reflects this model
diversity.
Previous works \citep{balaji, ood_ovadia} first average the softmax predictions
of the models in the ensemble and then use the entropy as a metric, i.e.\
$\entavg(f_1(x), ..., f_K(x)):=-\sum_{i=1}^{|\YY|} (f(x))_i \log (f(x))_i$ where
$f(x) := \frac{1}{K} \sum_{i=1}^K f_i(x)$ and $(f(x))_i$ is the $i^{\text{th}}$
element of $f(x) \in [0,1]^{|\YY|}$\footnote{We abuse notation slightly and
  denote our disagreement metric as $\Tdis$ to contrast it with the ensemble
  entropy metric $\entavg$, which first takes the average of the softmax outputs
and only afterwards computes the score.}. We argue later that averaging discards
information about the diversity of the models.

Recall that our average pairwise \emph{disagreement} between the outputs of $K$
models in an ensemble reads:\footnote{We abuse notation slightly and denote our
  disagreement metric as $\Tdis$ to contrast it with the ensemble entropy metric
  $\entavg$, which first takes the average of the softmax outputs and only
afterwards computes the score.}

\begin{align}
  \label{eq:statistic}
  \Tdis(f_1(x), ..., f_K(x)) := \frac{2}{K(K-1)} \sum_{i\neq j} \dis \left(f_i(x),
  f_j(x)\right),
\end{align}

\noindent where $\dis$ is a measure of disagreement between the softmax outputs
of two predictors, for example the total variation distance
$\dis_{\text{TV}}(f_i(x), f_j(x))=\frac{1}{2} \|f_i(x) - f_j(x) \|_1$ used in
our experiments.

We briefly highlight the reason why averaging softmax outputs \emph{first} like
in previous works relinquishes all the benefits of having a more diverse
ensemble, as opposed to the proposed pairwise score in
Equation~\ref{eq:statistic}. Recall that varying thresholds yield different true
negative and true positive rates (TNR and TPR, respectively) for a given
statistic.
In the sketch in Figure~\ref{fig:ensemble_disagreement} we show that the score
we propose, $\Tdis$, achieves a higher TNR compared to $\entavg$, for a fixed
TPR, which is a common way of evaluating statistical tests. Notice that the
detection region for $\entavg$ is always limited to a band around the average
model for any threshold value $t_0$. In order for the $\entavg$ to have large
TPR, this band needs to be wide, leading to many false positives. Instead, our
disagreement score exploits the diversity of the models to more accurately
detect OOD data. 

\begin{figure}[t]
  \begin{center}
    \includegraphics[width=0.8\textwidth]{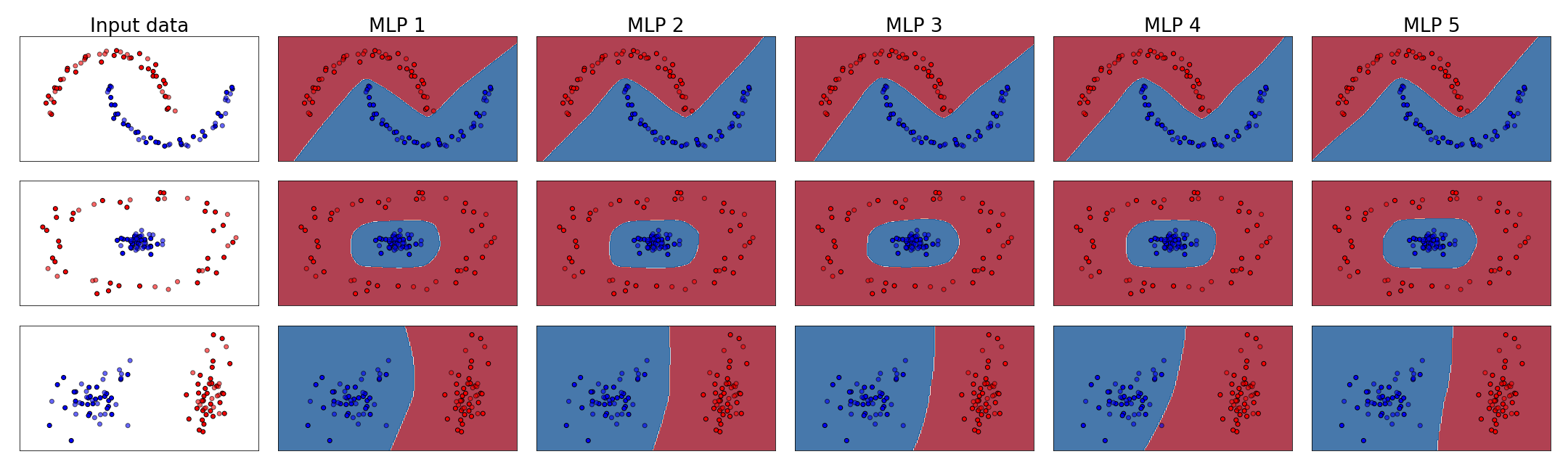}
  \end{center}

  \caption{Relying only on the randomness of SGD and of the weight
    initialization to diversify models is not enough, as it often yields similar
    classifiers. Each column shows a different predictor trained from random
    initializations with Adam. All models have the same 1-hidden layer MLP
  architecture.}

  \label{fig:vanilla_diversity}
\end{figure}

We now provide further quantitative evidence to support the intuition presented
in Figure~\ref{fig:ensemble_disagreement}.  The aggregation metric is tailored
to exploit ensemble diversity, which makes it particularly beneficial for
$\method$.  On the other hand, Vanilla Ensembles only rely on the stochasticity
of the training process and the random initializations of the weights to produce
diverse models, which often leads to classifiers that are strikingly similar as
we show in Figure~\ref{fig:vanilla_diversity} for a few 2D data sets. As a
consequence, using our disagreement score $\Tdis$ for Vanilla Ensembles can
sometimes hurt novelty detection performance. To see this, consider the extreme
situation in which the models in the ensemble are identical, i.e.\ $f_1 = f_2$.
Then it follows that $\Tdis(f_1(x), f_2(x)) = 0$, for all test points $x$ and
for any function $\rho$ that satisfies the distance axioms.  

We note that the disagreement score that we propose takes a form that is similar
to previous diversity scores, e.g.\ \citet{Zhang2010, mcd_ood}. In the context
of regression, one can measure uncertainty using the variance of the outputs
metric previously employed in works such as \citet{gal2016}. However, we point
out that using the output variance requires that the ensemble is the result of
sampling from a random process (e.g.\ sampling different training data for the
models, or sampling different parameters from a posterior). In our framework, we
obtain the ensemble by solving a different optimization problem for each of the
models by assigning a different label to the unlabeled data.  Therefore, despite
their similarities, our disagreement score and the output variance are, on a
conceptual level, fundamentally different metrics.

Table~\ref{table:score_comparison} shows that $\Tdis$ leads to worse novelty 
detection performance for Vanilla Ensembles, compared to using the entropy of
the average softmax score, $\entavg$, which was proposed in prior work.
However, if the ensembles are indeed diverse, as we argue is the case for our
method $\method$ (see Section~\ref{sec:earlystopping}), then there is a clear
advantage to using a score that, unlike $\entavg$, takes diversity into account,
as shown in Table~\ref{table:score_comparison} for 5-model $\method$ ensembles.

\begin{table}[h]
\tiny

\caption{\small{The disagreement score that we propose $\Tdis$ exploits ensemble
    diversity and benefits in particular $\method$ ensembles. Novelty detection
    performance is significantly improved when using $\Tdis$ compared to the
    previously proposed $\entavg$ metric. Since Vanilla Ensemble are not diverse
    enough, a score that relies on model diversity can hurt novelty detection
    performance. We highlight the AUROC and the TNR@95 obtained with the score
    function that is $\bestnonreto{best for Vanilla Ensemble}$ and the
$\bestreto{best for \method}$.}}

\label{table:score_comparison}
\begin{center}

\hyphenpenalty10000
\begin{tabularx}{0.8\textwidth}{ll| XXXX}
\toprule
\makecell{ID data} & \makecell{OOD data} & \makecell{Vanilla\\Ensembles\\$\entavg$} &  \makecell{Vanilla\\Ensembles\\$\Tdis$} & \makecell{ERD\\$\entavg$} &  \makecell{ERD\\$\Tdis$} \\
& & \multicolumn{4}{c}{AUROC $\uparrow$ / TNR@95 $\uparrow$ } \\
\midrule
$\text{ SVHN }$ & $\text{ CIFAR10 }$ & \bestnonreto{0.97} / \bestnonreto{0.88} & 0.96 / \bestnonreto{0.89} & 0.86 / 0.85 & \bestreto{0.99} / \bestreto{0.97} \\
$\text{ CIFAR10 }$ & $\text{ SVHN }$ & \bestnonreto{0.92} / \bestnonreto{0.78} & 0.91 / \bestnonreto{0.78} & 0.92 / 0.92 & \bestreto{1.00} / \bestreto{1.00} \\
$\text{ CIFAR100 }$ & $\text{ SVHN }$ & \bestnonreto{0.84} / \bestnonreto{0.48} & 0.79 / 0.46 & 0.36 / 0.35 & \bestreto{1.00} / \bestreto{1.00} \\
$\text{ SVHN[0:4] }$ & $\text{ SVHN[5:9] }$ & \bestnonreto{0.92} / \bestnonreto{0.69} & 0.91 / \bestnonreto{0.69} & \bestreto{0.94} / \bestreto{0.66} & \bestreto{0.94} / \bestreto{0.66} \\
$\text{ CIFAR10[0:4] }$ & $\text{ CIFAR10[5:9] }$ & \bestnonreto{0.80} / \bestnonreto{0.39} & \bestnonreto{0.80} / \bestnonreto{0.39} & \bestreto{0.91} / 0.65 & \bestreto{0.91} / \bestreto{0.66} \\
$\text{ CIFAR100[0:49] }$ & $\text{ CIFAR100[50:99] }$ & \bestnonreto{0.78} / \bestnonreto{0.35} & 0.76 / 0.34 & 0.63 / 0.38 & \bestreto{0.81} / \bestreto{0.40} \\

\midrule
\multicolumn{2}{c|}{Average} & \bestnonreto{0.87} / \bestnonreto{0.60} & 0.86 / 0.59 & 0.77 / 0.64 & \bestreto{0.94} / \bestreto{0.78} \\

\bottomrule
\end{tabularx}

\end{center}
\end{table}

\begin{figure*}[h]
  \centering
  \begin{subfigure}[t]{0.3\textwidth}
    \centering
    \includegraphics[width=\textwidth]{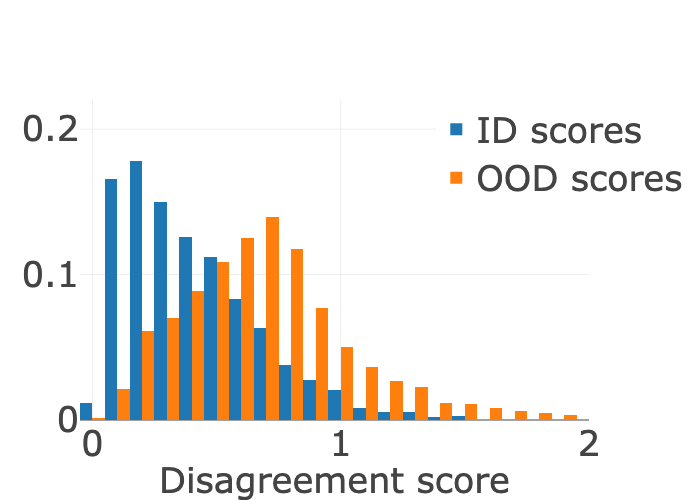}
    \caption{Not enough diversity (MCD)}
    \label{fig:scores_mcd_es}
  \end{subfigure}
  \begin{subfigure}[t]{0.3\textwidth}
    \centering
    \includegraphics[width=\textwidth]{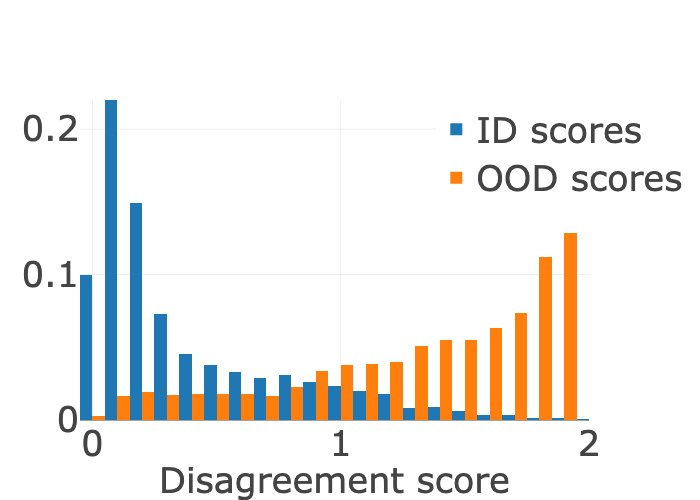}
    \caption{Regularized diversity (ERD)}
    \label{fig:scores_erd}
  \end{subfigure}

\caption{Distribution of disagreement scores on ID and OOD data
  for an ensemble that is not diverse enough (\textbf{Left}), and an ensemble
  with regularized disagreement (\textbf{Right}). Note that MCD is early-stopped
  using oracle OOD data. ID=CIFAR10[0:4], OOD=CIFAR10[5:9].}

\end{figure*}

We highlight once again that other methods that attempt to obtain diverse
ensembles, such as MCD, fail to train models with sufficient disagreement, even
when they use oracle OOD for hyperparameter tuning
(Figure~\ref{fig:scores_mcd_es}).

\section{Taxonomy of OOD detection methods according to overall objective}
\label{sec:appendix_related_work}

We now provide more details regarding the categorization of OOD detection
approaches based on the different surrogate objectives that they
use in order to detect OOD samples.

\paragraph{Learning the ID marginal $\iddist$.} We loosely define OOD samples as
all $x$ for which $\iddist(x) <\alpha$, for a small constant $\alpha > 0$.
Therefore, if we had access to the marginal training distribution $\iddist$,
we would have perfect OOD detection. Realistically, however, $\iddist$ is
unknown, and we need to resort to estimating it. Explicit density estimation
with generative models \citep{ganomaly2018, nalisnick} is inherently
difficult in high dimensions. 
Alternatively, one-class classification \citep{mari2010, Ruff2020, sohn2021} and
PU learning approaches \citep{duPlessis14, Kiryo17} try to directly learn a
discriminator between ID and OOD data in the presence of known (e.g.\ A-UND) or
unknown (e.g.\ SSND) OOD data. However, these methods tend to produce
indistinguishable representations for inliers and outliers when the ID
distribution consists of many diverse classes.

\paragraph{Learning $\iddist$ using label information (ours).} Since in a
prediction problem, the ID training set has class labels, one can take advantage
of that additional information to distinguish points in the support of $\iddist$
from OOD data. For instance, \citet{mahalanobis, gram_ood} propose to use the
intermediate representations of neural networks trained for prediction to detect
OOD data. Often, the task is to also simultaneously predict well on ID data, a
problem known as open-set recognition \citep{Geng2021} and
tackled by approaches like OpenHybrid \citep{openhybrid2020}.

\paragraph{Learning uncertainty estimates for $P_{Y|X}$.} In the prediction
setting, calibrated uncertainty estimates error could naturally be used to
detect OOD samples. Many uncertainty quantification methods are based on a
Bayesian framework \citep{gal2016, dpn} or calibration
improvement \citep{odin, Hafner2019}. However, neither of them perform as well
as other OOD methods mentioned above \citep{ood_ovadia}.


\section{Experiment details}
\label{sec:appendix_experiments}

\subsection{Baselines}

In this section we describe in detail the baselines with which we compare our
method and describe how we choose their hyperparameters. For all baselines we
use the hyperparameters suggested by the authors for the respective data sets
(e.g.\ different hyperparameters for CIFAR10 or ImageNet). For all methods, we
use pretrained models provided by the authors. However, we note that for the
novel-class settings, pretraining on the entire training set means that the
model is exposed to the OOD classes as well, which is undesirable. Therefore,
for these settings we pretrain only on the split of the training set that
contains the ID classes. Since the classification problem is similar to the
original one of training on the entire training set, we use the same
hyperparameters that the authors report in the original papers.

Moreover, we point out that even though different methods use different model
architectures, that is not inherently unreasonable when the goal is novelty
detection, since it is not clear if a complex model is more desirable than a
smaller model. For this reason, we use the model architecture recommended by the
authors of the baselines and which was used to produce the good results reported
in their published works. For Vanilla Ensembles and for $\method$ we show
results for different architectures in
Appendix~\ref{sec:appendix_different_arch}.

\begin{itemize}


  \item \textbf{Vanilla Ensembles} \citep{balaji}: We train an ensemble on the
    training set according to the true labels. For a test sample, we average the
    outputs of the softmax probabilities predicted by the models, and use the
    entropy of the resulting distribution as the score for the hypothesis test
    described in Section~\ref{sec:disagreement}. We use ensembles of 5 models,
    with the same architecture and hyperparameters as the ones used for
    $\method$. Hyperparameters are tuned to achieve good validation accuracy.

  \item \textbf{Gram method} \citep{gram_ood}: The Gram baseline is similar to
    the Mahalanobis method in that both use the intermediate feature
    representations obtained with a deep neural network to determine whether a
    test point is an outlier. However, what sets the Gram method apart is the
    fact that it does not need any OOD data for training or calibration. We use
    the pretrained models provided by the authors, or train our own, using the
    same methodology as described for the Mahalanobis baseline. For OOD
    detection, we use the code published by the authors. We note that for MLP
    models, the Gram method is difficult to tune and we could not find a
    configuration that works well, despite our best efforts and following the
    suggestions proposed during our communication with the authors.

  \item \textbf{Deep Prior Networks (DPN)} \citep{dpn}: DPN is a Bayesian Method
    that trains a neural network (Prior Network) to parametrize a Dirichlet
    distribution over the class probabilities.  We train a WideResNet WRN-28-10
    for $100$ epochs using SGD with momentum $0.9$, with an initial learning
    rate of $0.01$, which is decayed by $0.2$ at epochs $50$, $70$, and $90$.
    For MNIST, we use EMINST/Letters as OOD for tuning. For all other settings,
    we use TinyImages as OOD for tuning.

  \item \textbf{Outlier Exposure} \citep{outlier_exposure}: This approach makes
    a model's softmax predictions close to the uniform distribution on the known
    outliers, while maintaining a good classification performance on the
    training distribution. We use the WideResNet architecture (WRN). For
    fine-tuning, we use the settings recommended by the authors, namely we train
    for $10$ epochs with learning rate $0.001$. For training from scratch, we
    train for $100$ epochs with an initial learning rate of $0.1$.  When the
    training data set is either CIFAR10/CIFAR100 or ImageNet, we use the default
    WRN parameters of the author's code, namely $40$ layers, $2$ widen-factor,
    droprate $0.3$.  When the training dataset is SVHN, we use the author's
    recommended parameters of $16$ layers, $4$ widen-factor and droprate $0.4$.
    All settings use the cosine annealing learning rate scheduler provided with
    the author's code, without any modifications. For all settings, we use
    TinyImages as known OOD data during training. In
    Section~\ref{sec:appendix_more_oe} we show results for known OOD data that
    is similar to the OOD data used for testing.

  \item \textbf{Mahalanobis} \citep{mahalanobis}: The method pretrains models on
    the labeled training data. For a test data point, it uses the intermediate
    representations of each layer as ``extracted features''. It then performs
    binary classification using logistic regression using these extracted
    features. In the original setting, the classification is done on
    ``training'' ID vs ``training'' OOD samples (which are from the same
    distribution as the test OOD samples).  Furthermore, hyperparameter tuning
    for the optimal amount of noise is performed on validation ID and OOD data.
    We use the WRN-28-10 architecture, pretrained for $200$ epochs.  The initial
    learning rate is $0.1$, which is decayed at epochs $60$, $120$, and $160$ by
    $0.2$. We use SGD with momentum $0.9$, and the standard weight decay of $5
    \cdot 10^{-4}$. The code published for the Mahalanobis method performs a
    hyperparameter search automatically for each of the data sets. 

\end{itemize}

The following baselines attempt to leverage the unlabeled data that is available
in applications such as the one depicted in Figure~\ref{fig:practical_sketch},
similar to $\method$. 

\begin{itemize}

  \item \textbf{Non-negative PU learning (nnPU)} \citep{Kiryo17}: The method
    trains a binary predictor to distinguish between a set of known positives
    (in our case the ID data) and a set that contains a mixture of positives and
    negatives (in our case the unlabeled set). To prevent the interpolation of
    all the unlabeled samples, \citet{Kiryo17} proposes a regularized objective.
    It is important to note that most training objectives in the PU learning
    literature require that the ratio between the positives and negatives in the
    unlabeled set is known or easy to estimate. For our experiments we always
    use the exact OOD ratio to train the nnPU baseline. Therefore, we obtain an
    upper bound on the AUROC/TNR@95. If the ratio is estimated from finite
    samples, then estimation errors may lead to slightly worse OOD detection
    performance. We perform a grid search over the learning rate and the
    threshold that appears in the nnPU regularizer and pick the option with the
    best validation accuracy measured on a holdout set with only positive
    samples (in our case, ID data). 

  \item \textbf{Maximum Classifier Discrepancy (MCD)} \citep{mcd_ood}: The MCD
    method trains two classifiers at the same time, and makes them disagree on
    the unlabeled data, while maintaining good classification performance.  We
    use the WRN-28-10 architecture as suggested in the paper.  We did not change
    the default parameters which came with the author's code, so weight decay is
    $10^{-4}$, and the optimizer is SGD with momentum $0.9$.  When available
    (for CIFAR10 and CIFAR100), we use the pretrained models provided by the
    authors. For the other training datasets, we use their methodology to
    generate pretrained models: We train a WRN-28-10 for 200 epochs.  The
    learning rate starts at 0.1 and drops by a factor of 10 at $50\%$ and $75\%$
    of the training progress.

  \item \textbf{Mahalanobis-U}: This is a slightly different version of the
    Mahalanobis baseline, for which we use early-stopped logistic regression to
    distinguish between the training set and an unlabeled set with ID and OOD
    samples (instead of discriminating a known OOD set from the inliers). The
    early stopping iteration is chosen to minimize the classification errors on
    a validation set that contains only ID data (recall that we do not assume to
    know which are the OOD samples).

\end{itemize}

In addition to these approaches that have been introduced in prior work, we also
propose a strong novel baseline that
that bares some similarity to PU learning and to $\method$. 

\begin{itemize}

 \item \textbf{Binary classifier} The approach consists in discriminating
   between the labeled ID training set and the mixed unlabeled set, that
   contains both ID and OOD data. We use regularization to prevent the trivial
   solution for which the entire unlabeled set is predicted as OOD. Unlike PU
   learning, the binary classifier does not require that the OOD ratio in the
   test distribution is known. The approach is similar to a method described in
   \citep{scott08} which also requires that the OOD ratio of the unlabeled set is
   known.  We tune the learning rate and the weight of the unlabeled samples in
   the training loss by performing a grid search and selecting the configuration
   with the best validation accuracy, computed on a holdout set containing only
   ID samples.  We note that the binary classifier that appears in
   Section~\ref{sec:appendix_medical} in the medical benchmark, is not the same
   as this baseline. For more details on the binary classifier that appears in
   the medical data experiments we refer the reader to \citet{Cao2020}.

\end{itemize}

\vspace{-0.2cm}
\subsection{Training configuration for $\method$}

For $\method$ we always use hyperparameters that give the best validation
accuracy when training a model on the ID training set. In other words, we pick
hyperparameter values that lead to good ID generalization and do not perform
further hyperparameter tuning for the different OOD data sets on which we
evaluate our approach. We point out that, if the ID labeled set is known to
suffer from class imbalance, subpopulation imbalance or label noise, any
training method that addresses these issues can be used instead of standard
empirical risk minimization to train our ensemble (e.g.\ see 
\citet{mahdi}).

For MNIST and FashionMNIST, we train ensembles of 3-layer MLP models with ReLU
activations. Each intermediate layer has 100 neurons. The models are optimized
using Adam, with a learning rate of $0.001$, for $10$ epochs.

For SVHN, CIFAR10/CIFAR100 and ImageNet, we train ensembles of ResNet20
\citep{He2015}. The models are initialized with weights pretrained for $100$
epochs on the labeled training set. We fine-tune each model for 10 epochs using
SGD with momentum $0.9$, and a learning rate of $0.001$.  The weights are
trained with an $\ell_2$ regularization coefficient of $5e-4$.  We use a batch
size of 128 for all scenarios, unless explicitly stated otherwise. We used the
same hyperparameters for all settings. 

For pretraining, we perform SGD for 100 epochs and use the same architecture and
hyperparameters as described above, with the exception of the learning rate that
starts at $0.1$, and is multiplied by $0.2$ at epochs $50$, $70$ and $90$.

Apart from $\method$, which fine-tunes the ensemble models starting from
pretrained weights, we also present in the Appendix results for $\method$++.
This variant of our method trains the models from random initializations, and
hence needs more iterations to converge, making it more computationally
expensive than $\method$. We train all models in the $\method$++ ensembles for
$100$ epochs with a learning rate that starts at $0.1$, and is multiplied by
$0.2$ at epochs $50$, $70$ and $90$. All other hyperparameters are the same as
for $\method$ ensembles.

For the medical data sets, we train a Densenet-121 as the authors do in the
original paper \citep{Cao2020}. For $\method$++, we do not use random weight
initializations, but instead we start with the ImageNet weights provided with
Tensorflow. The training configuration is exactly the same as for ResNet20,
except that we use a batch size of 32 due to GPU memory restrictions, and for
fine tuning we use a constant learning rate of $10^{-5}$.

\vspace{-0.2cm}
\subsection{Computational considerations for $\method$}

We note that $\method$ models reach the optimal stopping time within the first
10 epochs on all the data sets that we consider, which amounts to around $6$
minutes of training time 
if the models in the ensemble are fine-tuned in parallel on NVIDIA 1080 Ti GPUs.
This is substantially better than the cost of fine-tuning a large ViT
transformer model (which takes about 1 hour for 2500 iterations on the same
hardware). Moreover, since the loss we use to train the ensemble decouples over
the models, it allows for easy parallelization, unlike objectives like MCD where
the ensemble models are intertwined.


\vspace{-0.1cm}
\section{ID and OOD data sets}
\label{sec:appendix_datasets}
\vspace{-0.2cm}

\subsection{Data sets}

For evaluation, we use the following image data sets: MNIST \citep{mnist},
Fashion MNIST \citep{fashion}, SVHN \citep{svhn}, CIFAR10 and CIFAR100
\citep{cifar}.

For the experiments using MNIST and FashionMNIST the training set size is 50K,
the validation size is 10K, and the test ID and test OOD sizes are both 10K.
For SVHN, CIFAR10 and CIFAR100, the training set size is 40K, the validation
size is 10K, and the unlabeled set contains 10K samples: 5K are ID and 5K are
OOD. For evaluation, we use a holdout set of 10K examples (half ID, half OOD).
For the settings that use half of the classes as ID and the other half as OOD,
all the sizes are divided by 2.

\vspace{-0.2cm}
\subsection{Samples for the settings with novel classes}

\vspace{-0.2cm}
\begin{figure}[H]
  \centering
  \begin{subfigure}[b]{0.4\textwidth}
     \centering

    \includegraphics[width=\textwidth]{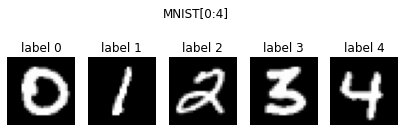}
    \includegraphics[width=\textwidth]{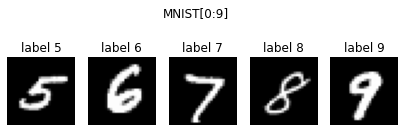}
    \includegraphics[width=\textwidth]{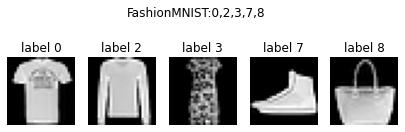}
    \includegraphics[width=\textwidth]{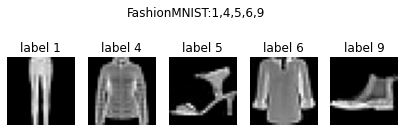}

     \caption{}
  \end{subfigure}
  \hfill
  \begin{subfigure}[b]{0.4\textwidth}
     \centering

    \includegraphics[width=\textwidth]{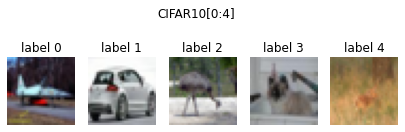}
    \includegraphics[width=\textwidth]{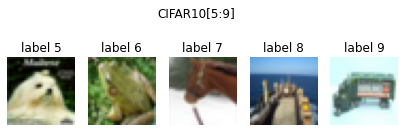}
    \includegraphics[width=\textwidth]{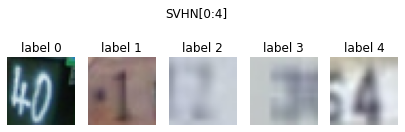}
    \includegraphics[width=\textwidth]{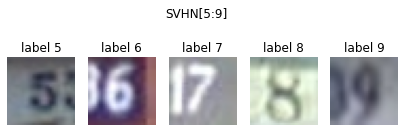}

     \caption{}
  \end{subfigure}

  \caption{(a) Data samples for the MNIST/FashionMNIST splits. (b) Data samples for the CIFAR10/SVHN splits.}
  \label{fig:data_samples_mnist_splits}

\end{figure}


\section{More experiments}
\label{sec:appendix_more_experiments}

We now present more experimental results that provide additional insights about
the proposed approach. We note that, unless otherwise specified, we use 5-model
ERD ensembles in this section.

\vspace{-0.2cm}
\subsection{Evaluation on the unlabeled set}
\label{sec:appendix_transductive_results}

In the main text we describe how one can leverage the unlabeled set $\targetset$
to obtain an novelty detection algorithm that accurately identifies outliers at test
time that similar to the ones in $\targetset$. It is, however, possible to also
use our method $\method$ to flag the OOD samples contained in the same set
$\targetset$ used for fine-tuning the ensemble. In
Table~\ref{table:ss_vs_transductive} we show that the novelty detection performance
of $\method$ is similar regardless of whether we use $\targetset$ for
evaluation, or a holdout test set $\testset$ drawn from the same distribution as
$\targetset$.

\begin{table}[H]
\tiny

\caption{Comparison between the novelty detection performance of $\method$ when
using a holdout test set $\testset$ for evaluation, or the same unlabeled set
$\targetset$ that was used for fine-tuning the models.}

\begin{center}

\hyphenpenalty10000
\begin{tabularx}{0.48\textwidth}{ll| cc}
\toprule
\makecell{ID data} & \makecell{OOD data} & \makecell{$\method$\\(eval on $\testset$)} & \makecell{$\method$\\(eval on $\targetset$)} \\
& & \multicolumn{2}{c}{AUROC $\uparrow$ / TNR@95 $\uparrow$ } \\
\midrule
$\text{ SVHN }$ & $\text{ CIFAR10 }$ & 1.00 / 0.99 & 1.00 / 0.99 \\
$\text{ CIFAR10 }$ & $\text{ SVHN }$ & 1.00 / 1.00 & 1.00 / 1.00 \\
$\text{ CIFAR100 }$ & $\text{ SVHN }$ & 1.00 / 1.00 & 1.00 / 1.00 \\

\midrule
$\text{ FMNIST[0,2,3,7,8] }$ & $\text{ FMNIST[1,4,5,6,9] }$ & 0.94 / 0.67 & 0.94 / 0.67 \\

$\text{ SVHN[0:4] }$ & $\text{ SVHN[5:9] }$ & 0.95 / 0.74 & 0.96 / 0.79 \\
$\text{ CIFAR10[0:4] }$ & $\text{ CIFAR10[5:9] }$ & 0.93 / 0.70 & 0.93 / 0.69 \\
$\text{ CIFAR100[0:49] }$ & $\text{ CIFAR100[50:99] }$ & 0.82 / 0.44 & 0.80 / 0.36 \\

\midrule
\multicolumn{2}{c|}{Average} & 0.95 / 0.79 & 0.95 / 0.79 \\

\bottomrule
\end{tabularx}

\label{table:ss_vs_transductive}
\end{center}
\end{table}

\vspace{-0.3cm}
\subsection{Comparison with other related works}
\label{sec:appendix_cifar10_cifar100}

We compare 5-model ERD ensembles to more OOD detection approaches. For various
reasons we did not run these methods ourselves on the data sets for which we
evaluate our method in Section~\ref{sec:experiments} (e.g.\ code not available,
unable to replicate published results, poor performance reported by the authors
etc). We collected the AUROC numbers presented in
Table~\ref{table_cifar10_cifar100} from the papers that introduce each method.
We note that our approach shows an excellent overall performance, being
consistently better than or on par with the related works that we consider.
While the method of \citet{fort2021} performs significantly better than all
other baselines on CIFAR10/CIFAR100 tasks, we argue in
Appendix~\ref{sec:appendix_vit} that this is primarily due to the convenient
choice of the data set used for pretraining the transformer models (i.e.\
Imagenet21k) which is strikingly similar to the ID and OOD data.

OpenHybrid \citep{openhybrid2020} is an open set recognition approach which
reports great near OOD detection performance. We note that, despite our best
efforts, we did not manage to match in our own experiments the results reported
in the paper, even after communicating with the authors and using the code that
they have provided. Moreover, we point out that the performance of OpenHybrid
seems to deteriorate significantly when the ID data consists of numerous
classes, as is the case for CIFAR100.

Furthermore, we note that generative models \citep{nalisnick, 
ganomaly2018} and one-class classification approaches \citep{Ruff2020, Tack2020,
sohn2021} showed generally bad performance, in particular on near OOD data. When
the ID training set is made up of several diverse classes, it is difficult to
represent accurately all the ID data, and only the ID data.

\begin{table*}[h]
\tiny
\centering

\caption{AUROC numbers collected from the literature for a number of relevant
OOD detection methods. We note that the method of \citet{fort2021} ($^\dagger$) uses a large scale
visual transformer models pretrained on a superset of the OOD data, i.e.\
ImageNet21k, while the method of \citet{sehwag2021} ($^*$) uses oracle OOD
samples for training from the same data set as test OOD. For the settings with random classes, the numbers are averages over 5
draws and the standard deviation is always strictly smaller than $0.01$ for our
method.}
\label{table_cifar10_cifar100}

\begin{tabularx}{\textwidth} {@{}ll @{} @{\hskip 0.1cm} XXXXXXXXX @{}} 
  \toprule
ID data  & OOD data & \citet{fort2021}$^\dagger$ & \citet{openhybrid2020} &
\citet{winkens2020} & \citet{Tack2020} & \citet{sehwag2021}$^*$ &\citet{liu2020hybrid} &
\citet{yujie2020} & ERD (ours) & ERD++ (ours)                   \\
\midrule

CIFAR10  & CIFAR100 & 98.52                             & 0.95          & 0.92
         & 0.92           &     0.93               & 0.91  & - & 0.92       & 0.95 \\
CIFAR100 & CIFAR10  & 96.23                             & 0.85
         & 0.78                                 & - & 0.78
         & - & -                                     & 0.91       & 0.94 \\
\midrule
\makecell[l]{SVHN: 6\\ random\\ classes} & \makecell[l]{SVHN: 4\\ random\\
classes}  & -                             & 0.94
          & -               & -                  & -
          & -                                & 0.91      & 0.94       & 0.94 \\
\makecell[l]{CIFAR10: 6\\ random\\ classes} & \makecell[l]{CIFAR10: 4\\ random\\
classes}  & -                             & 0.94
          & -                  & -               & -
          & -                                & 0.85      & 0.94       & 0.97 \\

  \bottomrule
\end{tabularx}
\vspace{0.5cm}

\end{table*}

\subsection{Shortcomings of pretrained ViT models for novelty detection}
\label{sec:appendix_vit}

In this section we provide further experimental results pointing to the fact
that large pretrained transformer models \citep{fort2021} can only detect near
OOD samples from certain specific data sets, and do not generalize well more
broadly.

\paragraph{Implementation details.} We fine-tune visual transformer (ViT) models
pretrained on Imagenet21k according the methodology described in
\citet{fort2021}. We report results using the ViT-S-16 architecture (~22 million
trainable parameters) which we fine-tune for 2500 iterations on labeled ID data.
We use the hyperparameters suggested by the authors and always ensure that the
prediction accuracy of the fine-tuned model on ID data is in the expected range.
The code published by the authors uses three different test statistics to detect
OOD data: the maximum softmax probability \citep{Hendrycks2017}, the vanilla
Mahalanobis distance \citep{mahalanobis} and a recently proposed variant of the
Mahalanobis approach \citep{Ren2021}. In Table~\ref{table:vit} we present only
the metrics obtained with the best-performing test statistic for ViT. We stress
that this favors the ViT method significantly, as different test statistics seem
to perform better on different data sets. Since test OOD data is unknown, it is
not possible to select which test statistic to use a priori, and hence, we use
oracle knowledge to give ViT models an unfair advantage.

\paragraph{Experimental results.} In Table~\ref{table:vit} we compare pretrained
visual transformers with 5-model $\method$ and $\method$++ ensembles.  Notably,
the data sets can be partitioned in two clusters, based on ViT novelty detection
performance. On the one hand, if the ID or OOD data comes from CIFAR10 or
CIFAR100, ViT models can detect novel-class samples well. Perhaps surprisingly,
ViT fails short of detecting OOD data perfectly (i.e. AUROC and TNR@95 of 1) on
easy tasks such as CIFAR10 vs SVHN or CIFAR100 vs SVHN, unlike $\method$ and a
number of other baseline approaches.

On the other hand, ViT shows remarkably poor performance on all other data sets,
when neither the ID nor the OOD data come from CIFAR10/CIFAR100. This includes
some of the novel disease use cases from the medical OOD detection benchmark
(see Appendix~\ref{sec:appendix_medical} for more details about the data sets).
This unsatisfactory performance persists even for larger ViT models (we
have tried ViT-S-16 and ViT-B-16 architectures), when fine-tuning for more
iterations (we have tried both 2500 and 10000 iterations), or when varying
hyperparameters such as the learning rate.

\paragraph{Intuition for why ViT fails.} We conjecture that the novelty
detection performance with pretrained ViT models relies heavily on the choice of
the pretraining data set. In particular, we hypothesize that, since
CIFAR10/CIFAR100 classes are included in the Imagenet21k data set used for
pretraining, the models learn features that are useful for distinguishing ID and
OOD classes when the ID and/or OOD data comes from CIFAR10/CIFAR100. Hence, this
would explain the good performance of pretrained models on the data sets at the
top of Table~\ref{table:vit}. On the other hand, when ID and OOD data is
strikingly different from the pretraining data, both ID and OOD samples are
projected to the same concentrated region of the representation space, which
makes it difficult to detect novel-class points.  Moreover, the process of
fine-tuning as it is described in \citet{fort2021} seems to not help to
alleviate this problem. This leads to the poor performance observed on the near
OOD data sets at the bottom of Table~\ref{table:vit}.

In conclusion, having a large pretraining data set seems to be beneficial when
the OOD data shares many visual and semantic features in common with the
pretraining data. However, in real-world applications it is often difficult to
collect such large data sets, which makes the applicability of pretrained ViT
models limited to only certain specific scenarios.

\begin{table}[H]
\tiny

\caption{Pretrained ViT models tend to perform well when the ID and OOD data is
  semantically similar to (or even included in) the pretraining data, e.g.\
  CIFAR10, CIFAR100 (top part), and their detection performance deteriorates
  drastically otherwise (bottom part). We compare ViT-S-16 models pretrained on
Imagenet21k with 5-model $\method$ and $\method$++ ensembles and
\bestnonreto{highlight} the best method. See Appendix~\ref{sec:appendix_medical}
for more details about the medical data sets.}

\label{table:vit}

\centering

\hyphenpenalty10000
\begin{tabularx}{0.59\linewidth}{ll| ccc}
\toprule
\makecell{ID data} & \makecell{OOD data} & \makecell{ViT} & \makecell{$\method$} & \makecell{$\method$++} \\
& & \multicolumn{3}{c}{AUROC $\uparrow$ / TNR@95 $\uparrow$ } \\
\midrule
$\text{ SVHN }$ & $\text{ CIFAR10 }$ & 0.98 / 0.90 & \bestnonreto{1.00 / 0.99} & \bestnonreto{1.00 / 0.99} \\
$\text{ CIFAR10 }$ & $\text{ SVHN }$ & 0.99 / 0.98 & \bestnonreto{1.00 / 1.00} & \bestnonreto{1.00 / 1.00} \\
$\text{ CIFAR100 }$ & $\text{ SVHN }$ & 0.96 / 0.84 & \bestnonreto{1.00 / 1.00} & \bestnonreto{1.00 / 1.00} \\
$\text{ CIFAR10[0:4] }$ & $\text{ CIFAR10[5:9] }$ & \bestnonreto{0.97 / 0.83} & 0.93 / 0.70 & 0.96 / 0.79 \\
$\text{ CIFAR100[0:49] }$ & $\text{ CIFAR100[50:99] }$ & \bestnonreto{0.90 / 0.56} & 0.82 / 0.44 & 0.85 / 0.45 \\

\midrule
$\text{ FMNIST[0,2,3,7,8] }$ & $\text{ FMNIST[1,4,5,6,9] }$ & 0.88 / 0.49 & 0.94 / 0.67 & \bestnonreto{0.95 / 0.71} \\

$\text{ SVHN[0:4] }$ & $\text{ SVHN[5:9] }$ & 0.89 / 0.68 & 0.95 / 0.74 & \bestnonreto{0.96 / 0.77} \\

$\text{ NIH ID (Xray) }$ & $\text{ PC (Xray) }$ & 0.87 / 0.38 & 0.94 / 0.63 & \bestnonreto{0.99 / 0.97} \\
$\text{ NIH ID (Xray) }$ & $\text{ NIH OOD (Xray) }$ & \bestnonreto{0.54 / 0.04} & 0.46 / 0.04 & 0.50 / 0.04 \\
$\text{ PC ID (Xray) }$ & $\text{ PC UC2 (Xray) }$ & 0.89 / 0.61 & \bestnonreto{0.99} / 0.99 & \bestnonreto{0.99 / 1.00} \\
$\text{ PC ID (Xray) }$ & $\text{ PC UC3 (Xray) }$ & 0.54 / 0.07 & \bestnonreto{0.77 / 0.17} & 0.72 / 0.08 \\
$\text{ DRD (fundus) }$ & $\text{ RIGA (fundus) }$ & 0.61 / 0.20 & 0.91 / 0.73 & \bestnonreto{1.00 / 0.98} \\

\bottomrule
\end{tabularx}

\end{table}

\vspace{-0.5cm}
\subsection{OOD detection for data with covariate shift}
\label{sec:appendix_cov_shift}

\vspace{-0.2cm}
In this section we evaluate the baselines and the method that we propose on
settings in which the OOD data suffers from covariate shift.
The goal is to identify all samples that come from the shifted distribution,
regardless of how strong the shift is. Notice that mild shifts may be easier to
tackle by domain adaptation, but when the goal is OOD detection they
pose a more difficult challenge.

We want to stress that in practice one may not be interested in identifying
\emph{all} samples with distribution shift as OOD, since a classifier may still
produce correct predictions on some of them. 
In contrast, when data suffers from covariate shift we can try to learn
predictors that perform well on both the training and the test distribution, and
we may use a measure of predictive uncertainty to identify only those test
samples on which the classifier cannot make confident predictions. Nevertheless,
we use these covariate shift settings as a challenging OOD detection benchmark
and show in Table~\ref{table:all_resnet_results} that our method $\method$ does
indeed outperform prior baselines on these difficult settings.

\begin{figure*}[h]
\vspace{-0.2cm}
  \centering
  \hspace*{\fill}%
  \begin{subfigure}[r]{0.49\textwidth}
  \centering
  \includegraphics[width=\textwidth]{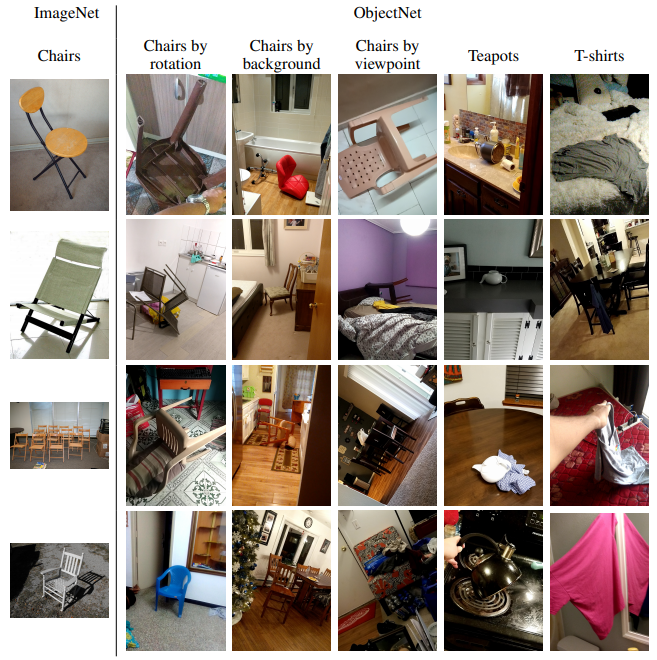}

\end{subfigure}
\hfill
\begin{subfigure}[c]{0.5\textwidth}
  \centering
  \includegraphics[width=\textwidth]{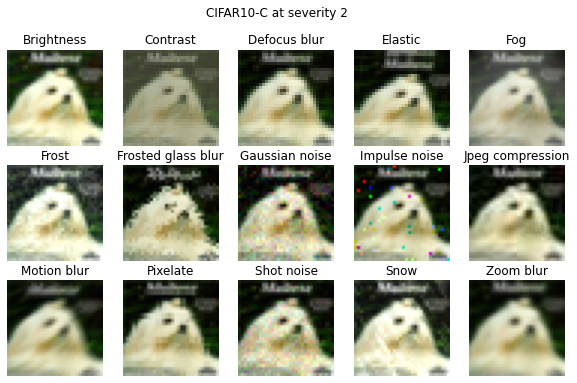}
  \includegraphics[width=\textwidth]{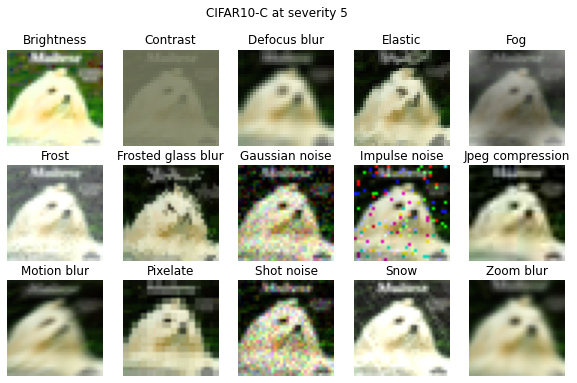}
\end{subfigure}

  \caption{Left: Samples from ImageNet and ObjectNet taken from the original
  paper by \citep{Barbu2019}. Right: Data samples for the corrupted CIFAR10-C
data set.}

  \label{fig:data_samples_cifar_and_objectnet}
\end{figure*}

We use as outliers corrupted variants of CIFAR10 and CIFAR100 \citep{cifar_c}, as
well as a scenario where ImageNet \citep{Deng2009} is used as ID data and
ObjectNet \citep{Barbu2019} as OOD, both resized to 32x32.
Figure~\ref{fig:data_samples_cifar_and_objectnet} shows samples from these data
sets. The Gram and nnPU baselines do not give satisfactory results on the
difficult CIFAR10/CIFAR100 settings in Table~\ref{table:main_results} and thus
we do not consider them for the covariate shift cases.  For the SSND methods
(e.g.\ MCD, Mahal-U and $\method$/$\method$++) we evaluate on the same unlabeled
set that is used for training (see the discussion in
Section~\ref{sec:appendix_transductive_results}).

Furthermore, we present results on distinguishing between CIFAR10 \citep{cifar}
and CIFAR10v2 \citep{recht}, a data set meant to be drawn from the same
distribution as CIFAR10 (generated from the Tiny Images collection).  In \citet{recht}, the authors argue
that CIFAR10 and CIFAR10v2 come from very similar distributions. They provide
supporting evidence by training a binary classifier to distinguish between them,
and observing that the accuracy that is obtained of 52.9\% is very close to
random.

Our experiments show that the two data sets are actually distinguishable,
contrary to what previous work has argued. First, our own binary classifier
trained on CIFAR10 vs CIFAR10v2 obtains a test accuracy of 67\%, without any
hyperparameter tuning. The model we use is a ResNet20 trained for 200 epochs
using SGD with momentum 0.9. The learning rate is decayed by 0.2 at epochs 90,
140, 160 and 180. We use 1600 examples from each data set for training, and we
validate using 400 examples from each data set.

\begin{table}[H]
\tiny
\caption{OOD detection performance on CIFAR10 vs CIFAR10v2}
\label{table:cifar10v2}
\begin{center}
\setlength{\tabcolsep}{1pt}
\hyphenpenalty10000
\begin{tabularx}{\textwidth}{@{}ll @{}| @{\hskip 0.2cm} X X X X X X X X  @{}}
\toprule
\makecell[l]{ ID data } & \makecell[l]{ OOD data } & \makecell[l]{ Vanilla\\Ensembles } & \makecell[l]{ DPN } & \makecell[l]{ OE } & \makecell[l]{ Mahal. } & \makecell[l]{ MCD } & \makecell[l]{ Mahal-U } & \makecell[l]{ $\method$ } & \makecell[l]{ $\method$++ } \\
& & \multicolumn{8}{c}{AUROC $\uparrow$ / TNR@95 $\uparrow$} \\
\midrule
\makecell[l]{ CIFAR10 } & \makecell[l]{ CIFAR10v2 } & \bestnonreto{0.64} / \bestnonreto{0.13} & 0.63 / 0.09 & \bestnonreto{0.64} / 0.12 & 0.55 / 0.08 & 0.58 / 0.10 & 0.56 / 0.07 & 0.76 / 0.26 & \bestreto{0.91} / \bestreto{0.80} \\

\bottomrule
\end{tabularx}
\end{center}
\end{table}

Our OOD detection experiments (presented in Table~\ref{table:cifar10v2}) show
that most baselines are able to distinguish between the two data sets, with
$\method$ achieving the highest performance. The methods which require OOD data
for tuning (Outlier Exposure and DPN) use CIFAR100.

\begin{table}[H]
\tiny

\caption{OOD detection performance on data with covariate shift. For $\method$
and vanilla ensembles, we train 5 ResNet20 models for each setting. The
evaluation metrics are computed on the unlabeled set.}

\begin{center}

\setlength{\tabcolsep}{1pt}
\hyphenpenalty10000
\begin{tabularx}{\textwidth}{@{}ll @{}| @{\hskip 0.2cm} X X X X X X X X  @{}}
\toprule
\makecell[l]{ ID data } & \makecell[l]{ OOD data } & \makecell[l]{ Vanilla\\Ensembles } & \makecell[l]{ DPN } & \makecell[l]{ OE } & \makecell[l]{ Mahal. } & \makecell[l]{ MCD } & \makecell[l]{ Mahal-U } & \makecell[l]{ $\method$ } & \makecell[l]{ $\method$++ } \\
& & \multicolumn{8}{c}{AUROC $\uparrow$ / TNR@95 $\uparrow$} \\
\midrule
$\text{ CIFAR10 }$ & $\text{ CIFAR10-C sev 2 (A) }$ & 0.68 / 0.20 & 0.73 / 0.31 & 0.70 / 0.20 & \bestnonreto{0.84} / \bestnonreto{0.53} & 0.82 / 0.50 & 0.75 / 0.38 & 0.96 / 0.86 & \bestreto{0.99} / \bestreto{0.95} \\
$\text{ CIFAR10 }$ & $\text{ CIFAR10-C sev 2 (W) }$ & 0.51 / 0.05 & 0.47 / 0.03 & 0.52 / 0.06 & \bestnonreto{0.58} / \bestnonreto{0.08} & 0.52 / 0.06 & 0.55 / 0.07 & 0.68 / 0.19 & \bestreto{0.86} / \bestreto{0.41} \\
$\text{ CIFAR10 }$ & $\text{ CIFAR10-C sev 5 (A) }$ & 0.84 / 0.49 & 0.89 / 0.60 & 0.86 / 0.54 & 0.94 / 0.80 & \bestnonreto{0.95} / \bestnonreto{0.84} & 0.88 / 0.63 & \bestreto{1.00} / 0.99 & \bestreto{1.00} / \bestreto{1.00} \\
$\text{ CIFAR10 }$ & $\text{ CIFAR10-C sev 5 (W) }$ & 0.60 / 0.10 & 0.72 / 0.10 & 0.63 / 0.11 & \bestnonreto{0.78} / \bestnonreto{0.27} & 0.60 / 0.08 & 0.68 / 0.12 & 0.98 / 0.86 & \bestreto{1.00} / \bestreto{1.00} \\
\midrule
$\text{ CIFAR100 }$ & $\text{ CIFAR100-C sev 2 (A) }$ & 0.68 / 0.20 & 0.62 / 0.18 & 0.65 / 0.19 & \bestnonreto{0.82} / \bestnonreto{0.48} & 0.72 / 0.29 & 0.67 / 0.22 & 0.94 / 0.76 & \bestreto{0.97} / \bestreto{0.86} \\
$\text{ CIFAR100 }$ & $\text{ CIFAR100-C sev 2 (W) }$ & 0.52 / 0.06 & 0.32 / 0.03 & 0.52 / 0.06 & \bestnonreto{0.55} / \bestnonreto{0.07} & 0.52 / 0.06 & 0.55 / 0.06 & 0.71 / 0.19 & \bestreto{0.86} / \bestreto{0.44} \\
$\text{ CIFAR100 }$ & $\text{ CIFAR100-C sev 5 (A) }$ & 0.78 / 0.37 & 0.74 / 0.36 & 0.76 / 0.37 & \bestnonreto{0.92} / \bestnonreto{0.72} & 0.91 / 0.65 & 0.84 / 0.55 & 0.99 / 0.97 & \bestreto{1.00} / \bestreto{0.99} \\
$\text{ CIFAR100 }$ & $\text{ CIFAR100-C sev 5 (W) }$ & 0.64 / 0.14 & 0.49 / 0.12 & 0.62 / 0.13 & \bestnonreto{0.71} / \bestnonreto{0.19} & 0.60 / 0.10 & 0.63 / 0.13 & 0.96 / 0.71 & \bestreto{0.98} / \bestreto{0.89} \\

\midrule
$\text{ Tiny ImageNet }$ & $\text{ Tiny ObjectNet }$ & 0.82 / 0.49 & 0.70 / 0.32 & 0.79 / 0.37 & 0.75 / 0.26 & \bestnonreto{0.99} / \bestnonreto{0.98} & 0.72 / 0.25 & 0.98 / 0.88 & \bestreto{0.99} / \bestreto{0.98} \\

\midrule
\multicolumn{2}{c | @{\hskip 0.2cm}}{Average} & 0.67 / 0.23 & 0.63 / 0.23 & 0.67 / 0.23 & \bestnonreto{0.76} / 0.38 & 0.74 / \bestnonreto{0.39} & 0.70 / 0.27 & 0.91 / 0.71 & \bestreto{0.96} / \bestreto{0.83} \\

\bottomrule
\end{tabularx}

\label{table:all_resnet_results}
\end{center}
\end{table}

\vspace{-0.5cm}
\subsection{Results with a smaller unlabeled set}
\label{sec:appendix_small_test_set}

We now show that our method performs well even when the unlabeled set is
significantly smaller. In particular, we show in the table below that $\method$
maintains a high AUROC and TNR@95 even when only 1,000 unlabeled samples are
used for fine-tuning (500 ID and 500 OOD). 

\begin{table}[H]
\tiny

\caption{Experiments with a test set of size 1,000, with an equal number of ID
and OOD test samples. For $\method$ and vanilla ensembles, we train 5 ResNet20 models for
each setting. The evaluation metrics are computed on the unlabeled set.}
\vspace{-0.5cm}

\begin{center}

\setlength{\tabcolsep}{1pt}
\hyphenpenalty10000
\begin{tabularx}{\textwidth}{@{}ll @{}| @{\hskip 0.2cm} X X X X X X X  @{}}
\toprule
\makecell[l]{ ID data } & \makecell[l]{ OOD data } & \makecell[l]{ Vanilla\\Ensembles } & \makecell[l]{ DPN } & \makecell[l]{ OE } & \makecell[l]{ Mahal. } & \makecell[l]{ MCD } & \makecell[l]{ Mahal-U } & \makecell[l]{ $\method$ } \\
& & \multicolumn{7}{c}{AUROC $\uparrow$ / TNR@95 $\uparrow$} \\
\midrule
$\text{ SVHN }$ & $\text{ CIFAR10 }$ & 0.97 / 0.88 & \bestnonreto{1.00} / \bestnonreto{1.00} & \bestnonreto{1.00} / \bestnonreto{1.00} & 0.99 / 0.98 & 0.97 / 0.85 & 0.99 / 0.95 & \bestreto{1.00} / \bestreto{0.99} \\
$\text{ CIFAR10 }$ & $\text{ SVHN }$ & 0.92 / 0.78 & 0.95 / 0.85 & 0.97 / 0.89 & \bestnonreto{0.99} / \bestnonreto{0.96} & 1.00 / 0.98 & 0.99 / 0.96 & \bestreto{1.00} / \bestreto{1.00} \\
$\text{ CIFAR100 }$ & $\text{ SVHN }$ & 0.84 / 0.48 & 0.77 / 0.44 & 0.82 / 0.50 & \bestnonreto{0.98} / \bestnonreto{0.90} & 0.97 / 0.73 & 0.98 / 0.92 & \bestreto{0.99} / \bestreto{1.00} \\
$\text{ SVHN[0:4] }$ & $\text{ SVHN[5:9] }$ & \bestnonreto{0.92} / 0.69 & 0.87 / 0.19 & 0.85 / 0.52 & \bestnonreto{0.92} / \bestnonreto{0.71} & 0.91 / 0.51 & 0.91 / 0.63 & \bestreto{0.97} / \bestreto{0.86} \\
$\text{ CIFAR10[0:4] }$ & $\text{ CIFAR10[5:9] }$ & 0.80 / 0.39 & \bestnonreto{0.82} / 0.32 & \bestnonreto{0.82} / \bestnonreto{0.41} & 0.79 / 0.27 & 0.69 / 0.25 & 0.64 / 0.13 & \bestreto{0.87} / \bestreto{0.50} \\
$\text{ CIFAR100[0:49] }$ & $\text{ CIFAR100[50:99] }$ & \bestnonreto{0.78} / \bestnonreto{0.35} & 0.70 / 0.26 & 0.74 / 0.31 & 0.72 / 0.20 & 0.70 / 0.26 & 0.72 / 0.19 & \bestreto{0.79} / \bestreto{0.38} \\

\midrule
$\text{ CIFAR10 }$ & $\text{ CIFAR10-C sev 2 (A) }$ & 0.68 / 0.20 & 0.73 / 0.31 & 0.70 / 0.20 & \bestnonreto{0.84} / \bestnonreto{0.53} & 0.82 / 0.50 & 0.75 / 0.38 & \bestreto{0.91} / \bestreto{0.71} \\
$\text{ CIFAR10 }$ & $\text{ CIFAR10-C sev 2 (W) }$ & 0.51 / 0.05 & 0.47 / 0.03 & 0.52 / 0.06 & \bestnonreto{0.58} / \bestnonreto{0.08} & 0.52 / 0.06 & 0.55 / 0.07 & \bestreto{0.57} / \bestreto{0.09} \\
$\text{ CIFAR10 }$ & $\text{ CIFAR10-C sev 5 (A) }$ & 0.84 / 0.49 & 0.89 / 0.60 & 0.86 / 0.54 & \bestnonreto{0.94} / \bestnonreto{0.80} & 0.95 / 0.84 & 0.88 / 0.63 & \bestreto{0.99} / \bestreto{0.95} \\
$\text{ CIFAR10 }$ & $\text{ CIFAR10-C sev 5 (W) }$ & 0.60 / 0.10 & 0.72 / 0.10 & 0.63 / 0.11 & \bestnonreto{0.78} / \bestnonreto{0.27} & 0.60 / 0.08 & 0.68 / 0.12 & \bestreto{0.92} / \bestreto{0.67} \\
\midrule
$\text{ CIFAR100 }$ & $\text{ CIFAR100-C sev 2 (A) }$ & 0.68 / 0.20 & 0.62 / 0.18 & 0.65 / 0.19 & \bestnonreto{0.82} / \bestnonreto{0.48} & 0.72 / 0.29 & 0.67 / 0.22 & \bestreto{0.84} / \bestreto{0.48} \\
$\text{ CIFAR100 }$ & $\text{ CIFAR100-C sev 2 (W) }$ & 0.52 / 0.06 & 0.32 / 0.03 & 0.52 / 0.06 & \bestnonreto{0.55} / \bestnonreto{0.07} & 0.52 / 0.06 & \bestreto{0.55} / 0.06 & \bestreto{0.55} / \bestreto{0.07} \\
$\text{ CIFAR100 }$ & $\text{ CIFAR100-C sev 5 (A) }$ & 0.78 / 0.37 & 0.74 / 0.36 & 0.76 / 0.37 & \bestnonreto{0.92} / \bestnonreto{0.72} & 0.91 / 0.65 & 0.84 / 0.55 & \bestreto{0.96} / \bestreto{0.80} \\
$\text{ CIFAR100 }$ & $\text{ CIFAR100-C sev 5 (W) }$ & 0.64 / 0.14 & 0.49 / 0.12 & 0.62 / 0.13 & \bestnonreto{0.71} / \bestnonreto{0.19} & 0.60 / 0.10 & 0.63 / 0.13 & \bestreto{0.81} / \bestreto{0.25} \\

\midrule
\multicolumn{2}{c | @{\hskip 0.2cm}}{Average} & 0.75 / 0.37 & 0.72 / 0.34 & 0.75 / 0.38 & \bestnonreto{0.82} / \bestnonreto{0.51} & 0.78 / 0.44 & 0.77 / 0.42 & \bestreto{0.87} / \bestreto{0.62} \\

\bottomrule
\end{tabularx}

\end{center}
\end{table}

\vspace{-0.5cm}
\subsection{More results for Outlier Exposure}
\label{sec:appendix_more_oe}

The Outlier Exposure method needs access to a set of OOD samples during
training. The numbers we report in the rest of paper for Outlier Exposure are
obtained by using the TinyImages data set as the OOD samples that are seen
during training.  In this section we explore the use of an OOD$_{\text{train}}$
data set that is more similar to the OOD data observed at test time. This is a
much easier setting for the Outlier Exposure method: the closer
OOD$_{\text{train}}$ is to OOD$_{\text{test}}$, the easier it will be for the
model tuned on OOD$_{\text{train}}$ to detect the test OOD samples.

In the table below we focus only on the settings with corruptions. For each
corruption type, we use the lower severity corruption as OOD$_{\text{train}}$
and evaluate on the higher severity data and vice versa. We report for each metric
the average taken over all corruptions (A), and the value for the worst-case
setting (W).

\begin{table}[H]
  \small

\begin{center}

\hyphenpenalty10000
\begin{tabularx}{0.8\textwidth}{ll| cc}
\toprule
\makecell{ID data} & \makecell{OOD data} & \makecell{OE (trained on sev5)} & \makecell{OE (trained on sev2)} \\
& & \multicolumn{2}{c}{AUROC $\uparrow$} \\
\midrule
$\text{ CIFAR10 }$ & $\text{ CIFAR10-C sev 2 (A) }$ & 0.89 & N/A \\
$\text{ CIFAR10 }$ & $\text{ CIFAR10-C sev 2 (W) }$ & 0.65 & N/A \\
$\text{ CIFAR10 }$ & $\text{ CIFAR10-C sev 5 (A) }$ & N/A & 0.98 \\
$\text{ CIFAR10 }$ & $\text{ CIFAR10-C sev 5 (W) }$ & N/A & 0.78 \\
\midrule
$\text{ CIFAR100 }$ & $\text{ CIFAR100-C sev 2 (A) }$ & 0.85 & N/A \\
$\text{ CIFAR100 }$ & $\text{ CIFAR100-C sev 2 (W) }$ & 0.59 & N/A \\
$\text{ CIFAR100 }$ & $\text{ CIFAR100-C sev 5 (A) }$ & N/A & 0.97 \\
$\text{ CIFAR100 }$ & $\text{ CIFAR100-C sev 5 (W) }$ & N/A & 0.67 \\

\midrule
\multicolumn{2}{c|}{Average} & 0.87 & 0.98 \\

\bottomrule
\end{tabularx}

\end{center}

\label{table:oe}
    \caption{Results for Outlier Exposure, when using the same corruption type,
    but with a higher/lower severity, as OOD data seen during training.}

\end{table}

\vspace{-0.5cm}
\subsection{Results on MNIST and FashionMNIST}

\begin{table}[H]
\tiny

\caption{Results on MNIST/FashionMNIST settings. For $\method$ and vanilla
ensembles, we train 5 3-hidden layer MLP models for each setting. The evaluation metrics are computed on the unlabeled set.}

\begin{center}

\setlength{\tabcolsep}{1pt}
\hyphenpenalty10000
\begin{tabularx}{\textwidth}{@{}ll @{}| @{\hskip 0.2cm} X X X X X X X X X X  @{}}
\toprule
\makecell[l]{ ID data } & \makecell[l]{ OOD data } & \makecell[l]{ Vanilla\\Ensembles } & \makecell[l]{ DPN } & \makecell[l]{ OE } & \makecell[l]{ Mahal. } & \makecell[l]{ nnPU } & \makecell[l]{ MCD } & \makecell[l]{ Mahal-U } & \makecell[l]{ Bin.\\Classif. } & \makecell[l]{ $\method$ } & \makecell[l]{ $\method$++ } \\
& & \multicolumn{10}{c}{AUROC $\uparrow$ / TNR@95 $\uparrow$} \\
\midrule
$\text{ MNIST }$ & $\text{ FMNIST }$ & 0.81 / 0.01 & \bestnonreto{1.00} / \bestnonreto{1.00} & \bestnonreto{1.00} / \bestnonreto{1.00} & \bestnonreto{1.00} / \bestnonreto{1.00} & \bestnonreto{1.00} / \bestnonreto{1.00} & \bestnonreto{1.00} / 0.98 & \bestnonreto{1.00} / \bestnonreto{1.00} & 1.00 / 1.00 & \bestreto{1.00} / \bestreto{1.00} & \bestreto{1.00} / \bestreto{1.00} \\
$\text{ FMNIST }$ & $\text{ MNIST }$ & 0.87 / 0.42 & \bestnonreto{1.00} / \bestnonreto{1.00} & 0.68 / 0.16 & 0.99 / 0.97 & \bestnonreto{1.00} / \bestnonreto{1.00} & \bestnonreto{1.00} / \bestnonreto{1.00} & 0.99 / 0.96 & 1.00 / 1.00 & \bestreto{1.00} / \bestreto{1.00} & \bestreto{1.00} / \bestreto{1.00} \\
$\text{ MNIST[0:4] }$ & $\text{ MNIST[5:9] }$ & 0.94 / 0.72 & \bestnonreto{0.99} / 0.97 & 0.95 / 0.78 & \bestnonreto{0.99} / \bestnonreto{0.98} & \bestnonreto{0.99} / 0.97 & 0.96 / 0.76 & \bestnonreto{0.99} / \bestnonreto{0.98} & 0.99 / 0.94 & \bestreto{0.99} / 0.96 & \bestreto{0.99} / \bestreto{0.97} \\
$\text{ FMNIST[0,2,3,7,8] }$ & $\text{ FMNIST[1,4,5,6,9] }$ & 0.64 / 0.07 & 0.77 / 0.15 & 0.66 / 0.12 & 0.77 / 0.20 & \bestnonreto{0.95} / \bestnonreto{0.71} & 0.78 / 0.30 & 0.82 / 0.39 & 0.95 / 0.66 & \bestreto{0.94} / 0.67 & \bestreto{0.94} / \bestreto{0.68} \\

\midrule
\multicolumn{2}{c | @{\hskip 0.2cm}}{Average} & 0.82 / 0.30 & 0.94 / 0.78 & 0.82 / 0.51 & 0.94 / 0.79 & \bestnonreto{0.98} / \bestnonreto{0.92} & 0.94 / 0.76 & 0.95 / 0.83 & 0.98 / 0.90 & \bestreto{0.98} / \bestreto{0.91} & \bestreto{0.98} / \bestreto{0.91} \\

\bottomrule
\end{tabularx}

\end{center}
\end{table}

\vspace{-0.4cm}
For FashionMNIST we chose this particular split (i.e. classes 0,2,3,7,8 vs
classes 1,4,5,6,9) because the two partitions are more similar to each other.
This makes novelty detection more difficult than the 0-4 vs 5-9 split.

\vspace{-0.2cm}
\subsection{Vanilla and $\method$ Ensembles with different architectures}
\label{sec:appendix_different_arch}

In this section we present OOD detection results for 5-model Vanilla and
$\method$ ensembles with different architecture choices, and note that the
better performance of our method is maintained across model classes. Moreover,
we observe that $\method$ benefits from employing more complex models, like the
WideResNet.

\begin{table}[H]
\tiny

\caption{Results with three different architectures for Vanilla and $\method$
  ensembles. All ensembles comprise 5 models. For the corruption data sets, we
  report for each metric the average taken over all corruptions (A), and the
  value for the worst-case setting (W). The evaluation metrics are computed on
the unlabeled set.}

\begin{center}

\hyphenpenalty10000
\begin{tabularx}{\textwidth}{@{}ll @{} @{\hskip 0.2cm} XX >{\centering\arraybackslash} XX >{\centering\arraybackslash} XX@{}}
\toprule
& & \multicolumn{2}{c}{VGG16} & \multicolumn{2}{c}{ResNet20} & \multicolumn{2}{c}{WideResNet-28-10} \\
\cmidrule(l{2pt}r{2pt}){3-4}
\cmidrule(l{2pt}r{2pt}){5-6}
\cmidrule(l{2pt}r{2pt}){7-8}
\makecell{ID data} & \makecell{OOD data} & \makecell{Vanilla\\Ensembles} &  \makecell{ERD} & \makecell{Vanilla\\Ensembles} &  \makecell{ERD} & \makecell{Vanilla\\Ensembles} &  \makecell{ERD} \\
& & \multicolumn{6}{c}{AUROC $\uparrow$ / TNR@95 $\uparrow$ } \\
\midrule
$\text{ SVHN }$ & $\text{ CIFAR10 }$ & 0.97 / 0.88 & 0.99 / 0.94 & 0.97 / 0.88 & 0.99 / 0.97 & 0.96 / 0.86 & 1.00 / 0.99 \\
$\text{ CIFAR10 }$ & $\text{ SVHN }$ & 0.88 / 0.69 & 1.00 / 1.00 & 0.92 / 0.78 & 1.00 / 1.00 & 0.94 / 0.81 & 1.00 / 1.00 \\
$\text{ SVHN[0:4] }$ & $\text{ SVHN[5:9] }$ & 0.89 / 0.60 & 0.93 / 0.63 & 0.92 / 0.69 & 0.94 / 0.66 & 0.91 / 0.62 & 0.96 / 0.78 \\
$\text{ CIFAR10[0:4] }$ & $\text{ CIFAR10[5:9] }$ & 0.74 / 0.29 & 0.91 / 0.63 & 0.80 / 0.39 & 0.91 / 0.66 & 0.80 / 0.35 & 0.94 / 0.71 \\

\midrule
$\text{ CIFAR10 }$ & $\text{ CIFAR10-C sev 2 (A) }$ & 0.66 / 0.17 & 0.94 / 0.79 & 0.68 / 0.20 & 0.96 / 0.86 & 0.69 / 0.18 & 0.98 / 0.90 \\
$\text{ CIFAR10 }$ & $\text{ CIFAR10-C sev 2 (W) }$ & 0.51 / 0.05 & 0.68 / 0.19 & 0.51 / 0.05 & 0.68 / 0.19 & 0.51 / 0.05 & 0.84 / 0.35 \\
$\text{ CIFAR10 }$ & $\text{ CIFAR10-C sev 5 (A) }$ & 0.80 / 0.41 & 0.99 / 0.96 & 0.84 / 0.49 & 1.00 / 0.99 & 0.84 / 0.47 & 1.00 / 1.00 \\
$\text{ CIFAR10 }$ & $\text{ CIFAR10-C sev 5 (W) }$ & 0.58 / 0.10 & 0.95 / 0.72 & 0.60 / 0.10 & 0.98 / 0.86 & 0.59 / 0.09 & 0.99 / 0.97 \\

\midrule
\multicolumn{2}{c}{Average} & 0.75 / 0.40 & 0.92 / 0.73 & 0.78 / 0.45 & 0.93 / 0.77 & 0.78 / 0.43 & 0.96 / 0.84 \\

\bottomrule
\end{tabularx}

\end{center}
\end{table}

\vspace{-0.5cm}
\subsection{Impact of the ensemble size and of the choice of arbitrary label}
\label{sec:appendix_ensemble_size}

In this section we show novelty detection results with our method using a smaller
number of models for the ensembles. We notice that the performance is not
affected substantially, indicating that the computation cost of our approach
could be further reduced by fine-tuning smaller ensembles.

\begin{table}[H]
\tiny

\caption{Results obtained with smaller ensembles for $\method$. The numbers for
  $K < 5$ are averages over 3 runs, where we use a different set of arbitrary
  labels for each run to illustrate our method's stability with respect the
  choice of labels to be assigned to the unlabeled set. We note that the
standard deviations are small ($\sigma \le 0.01$ for the AUROC values and
$\sigma \le 0.08$ for the TNR@95 values).}

\begin{center}

\hyphenpenalty10000
\begin{tabularx}{\textwidth}{@{}l@{\hskip -0.01cm}l @{} @{\hskip 0.1cm} XX >{\centering\arraybackslash} XX >{\centering\arraybackslash} XX >{\centering\arraybackslash} XX@{}}
\toprule
& & \multicolumn{2}{c}{K=2} & \multicolumn{2}{c}{K=3} & \multicolumn{2}{c}{K=4} & \multicolumn{2}{c}{K=5} \\
\cmidrule(l{2pt}r{2pt}){3-4}
\cmidrule(l{2pt}r{2pt}){5-6}
\cmidrule(l{2pt}r{2pt}){7-8}
\cmidrule(l{2pt}r{2pt}){9-10}
\makecell{ID data} & \makecell{OOD data} & \makecell{ERD} & \makecell{ERD++} & \makecell{ERD} & \makecell{ERD++} & \makecell{ERD} & \makecell{ERD++} & \makecell{ERD} & \makecell{ERD++} \\
& & \multicolumn{6}{c}{AUROC $\uparrow$ / TNR@95 $\uparrow$ } \\
\midrule
$\text{ SVHN }$ & $\text{ CIFAR10 }$ & 0.99 / 0.98 & 0.99 / 0.99 & 0.99 / 0.98 & 1.00 / 0.99 & 0.99 / 0.98 & 1.00 / 0.99 & 1.00 / 0.99 & 1.00 / 0.99 \\
$\text{ CIFAR10 }$ & $\text{ SVHN }$ & 1.00 / 1.00 & 1.00 / 1.00 & 1.00 / 1.00 & 1.00 / 1.00 & 1.00 / 1.00 & 1.00 / 1.00 & 1.00 / 1.00 & 1.00 / 1.00 \\
$\text{ CIFAR100 }$ & $\text{ SVHN }$ & 1.00 / 1.00 & 1.00 / 1.00 & 1.00 / 1.00 & 1.00 / 1.00 & 1.00 / 1.00 & 1.00 / 1.00 & 1.00 / 1.00 & 1.00 / 1.00 \\
$\text{ SVHN[0:4] }$ & $\text{ SVHN[5:9] }$ & 0.95 / 0.69 & 0.94 / 0.68 & 0.95 / 0.73 & 0.95 / 0.75 & 0.96 / 0.76 & 0.96 / 0.77 & 0.95 / 0.74 & 0.96 / 0.77 \\
$\text{ CIFAR10[0:4] }$ & $\text{ CIFAR10[5:9] }$ & 0.89 / 0.55 & 0.92 / 0.58 & 0.89 / 0.57 & 0.94 / 0.70 & 0.90 / 0.57 & 0.95 / 0.73 & 0.93 / 0.70 & 0.96 / 0.79 \\
$\text{ CIFAR100[0:49] }$ & $\text{ CIFAR100[50:99] }$ & 0.81 / 0.40 & 0.82 / 0.43 & 0.81 / 0.41 & 0.84 / 0.44 & 0.81 / 0.41 & 0.84 / 0.44 & 0.82 / 0.44 & 0.85 / 0.45 \\

\midrule
\multicolumn{2}{c}{Average} & 0.94 / 0.77 & 0.95 / 0.78 & 0.94 / 0.78 & 0.95 / 0.81 & 0.94 / 0.79 & 0.96 / 0.82 & 0.95 / 0.81 & 0.96 / 0.83 \\

\bottomrule
\end{tabularx}

\end{center}
\end{table}

\vspace{-0.5cm}
\paragraph{Impact of the choice of arbitrary labels.} Furthermore, we note that
in the table we report averages over 3 runs of our method, where for each
run we use a different subset of $\YY$ to assign arbitrary labels to the
unlabeled data. We do this in order to assess the stability of $\method$
ensembles to the choice of the arbitrary labels and notice that the novelty 
detection performance metrics do not vary significantly.  Concretely, the
standard deviations are consistently below $0.01$ for all data sets for the
AUROC metric, and below $0.07$ for the TNR@95 metric.

\vspace{-0.2cm}
\subsection{Detection performance on different OOD data}
\label{sec:appendix_different_ood}

In this section we investigate whether the proposed method maintains its good
novelty detection performance when the test-time OOD data comes from a different
data set compared to the OOD data that is present in the unlabeled set used for
fine-tuning. In particular, we are interested if our approach can still identify
outliers in situations when they suffer from various corruptions. This scenario
can sometimes occur in practice, when machine failure or uncurated data can lead
to mild distribution shift.

Concretely, we focus on the difficult near OOD scenarios and take as ID half of
the CIFAR10 or CIFAR100 classes, while the other half is OOD. For this
experiment, we fine-tune the ERD ensembles using clean OOD data from the other
half of CIFAR10 and CIFAR100, respectively. For evaluation, we use clean ID
data and corrupted OOD samples from CIFAR10-C and CIFAR100-C, respectively.
We give more details on these corrupted data sets in
Appendix~\ref{sec:appendix_cov_shift}. We consider corruptions of severity 2 and
5 from all corruptions types.

In Table~\ref{table:different_ood} we show the average AUROC and the worst AUROC
over all corruption types for vanilla and ERD ensembles. Note that our approach
maintains a similar performance compared to the numbers presented in
Table~\ref{table:main_results} for same test-time OOD data. It is also
noteworthy that all the average AUROC values are consistently larger than the
baselines in Table~\ref{table:main_results}.

\begin{table}[H]
  \small
  \begin{center}

    \caption{Results obtained when evaluating on an OOD data set different from
    the one used for fine-tuning. All ERD ensembles are tuned on clean ID and
  OOD data and are evaluated on OOD data with corruptions.}

\hyphenpenalty10000
\begin{tabularx}{0.88\textwidth}{lll| cc}
\toprule
\makecell{ID data} & \makecell{OOD data in\\unlabeled set} & \makecell{Test-time\\OOD data} & \makecell{Vanilla\\Ensemble} & \makecell{ERD} \\
                   & & & \multicolumn{2}{c}{AUROC $\uparrow$} \\
\midrule
$\text{ CIFAR10[0:4] }$ & $\text{ CIFAR10[5:9] }$ & $\text{ CIFAR10[5:9]-C sev 2 (A) }$ & 0.82 & 0.93 \\
$\text{ CIFAR10[0:4] }$ & $\text{ CIFAR10[5:9] }$ & $\text{ CIFAR10[5:9]-C sev 2 (W) }$ & 0.77 & 0.88 \\
$\text{ CIFAR10[0:4] }$ & $\text{ CIFAR10[5:9] }$ & $\text{ CIFAR10[5:9]-C sev 5 (A) }$ & 0.85 & 0.91 \\
$\text{ CIFAR10[0:4] }$ & $\text{ CIFAR10[5:9] }$ & $\text{ CIFAR10[5:9]-C sev 5 (W) }$ & 0.79 & 0.86 \\
\midrule
$\text{ CIFAR100[0:49] }$ & $\text{ CIFAR100[50:99] }$ & $\text{ CIFAR100[50:99]-C sev 2 (A) }$ & 0.78 & 0.84 \\
$\text{ CIFAR100[0:49] }$ & $\text{ CIFAR100[50:99] }$ & $\text{ CIFAR100[50:99]-C sev 2 (W) }$ & 0.75 & 0.78 \\
$\text{ CIFAR100[0:49] }$ & $\text{ CIFAR100[50:99] }$ & $\text{ CIFAR100[50:99]-C sev 5 (A) }$ & 0.77 & 0.83 \\
$\text{ CIFAR100[0:49] }$ & $\text{ CIFAR100[50:99] }$ & $\text{ CIFAR100[50:99]-C sev 5 (W) }$ & 0.63 & 0.78 \\

\bottomrule
\end{tabularx}

\label{table:different_ood}
\end{center}
\end{table}



\vspace{-0.6cm}
\section{Medical OOD detection benchmark}
\label{sec:appendix_medical}

\vspace{-0.3cm}
The medical OOD detection benchmark is organized as follows. There are four
training (ID) data sets, from three different domains: two data sets with chest
X-rays, one with fundus imaging and one with histology images. For each ID data
set, the authors consider three different OOD scenarios:

\vspace{-0.4cm}
\begin{enumerate}[leftmargin=*]
  \compresslist

  \item Use case 1: The OOD data set contains images from a completely different
    domain, similar to our category of easy OOD detection settings.

  \item Use case 2: The OOD data set contains images with various corruptions,
    similar to the hard covariate shift settings that we consider in
    Section~\ref{sec:appendix_cov_shift}.

  \item Use case 3: The OOD data set contains images that come from novel
    classes, not seen during training.

\end{enumerate}

\vspace{-0.7cm}
\begin{figure}[H]
  \begin{center}
    \includegraphics[width=0.7\textwidth]{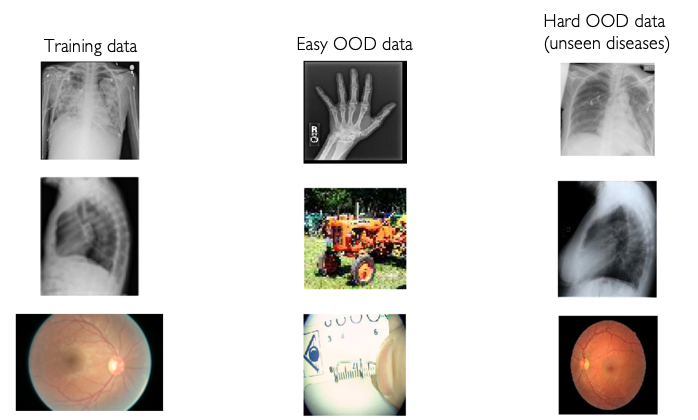}
  \end{center}

\vspace{-0.5cm}
  \caption{Samples from the medical image benchmark. There are 3 ID data sets
  containing frontal and lateral chest X-rays and retinal images. Hard OOD
samples contain images of diseases that are not present in the training set.}

  \label{fig:medical_samples}
\end{figure}

The authors evaluate a number of methods on all these scenarios. The methods can
be roughly categorized as follows:

\begin{enumerate}[leftmargin=*]
  \compresslist

  \item Data-only methods: Fully non-parametric approaches like kNN.

  \item Classifier-only methods: Methods that use a classifier trained on the
    training set, e.g.\ ODIN \citep{odin}, Mahalanobis \citep{mahalanobis}. $\method$
    falls into this category as well.

  \item Methods with Auxiliary Models: Methods that use an autoencoder or a
    generative model, like a Variational Autoencoder or a Generative Adversarial
    Network. Some of these approaches can be expensive to train and difficult to
    optimize and tune.

\end{enumerate}

We stress the fact that for most of these methods the authors use (known) OOD
data during training. Oftentimes the OOD samples observed during training come
from a data set that is very similar to the OOD data used for evaluation.  For
exact details regarding the data sets and the methods used for the benchmark, we
refer the reader to \citet{Cao2020}. 

\begin{figure}[H]
  \begin{center}
    \includegraphics[width=\textwidth]{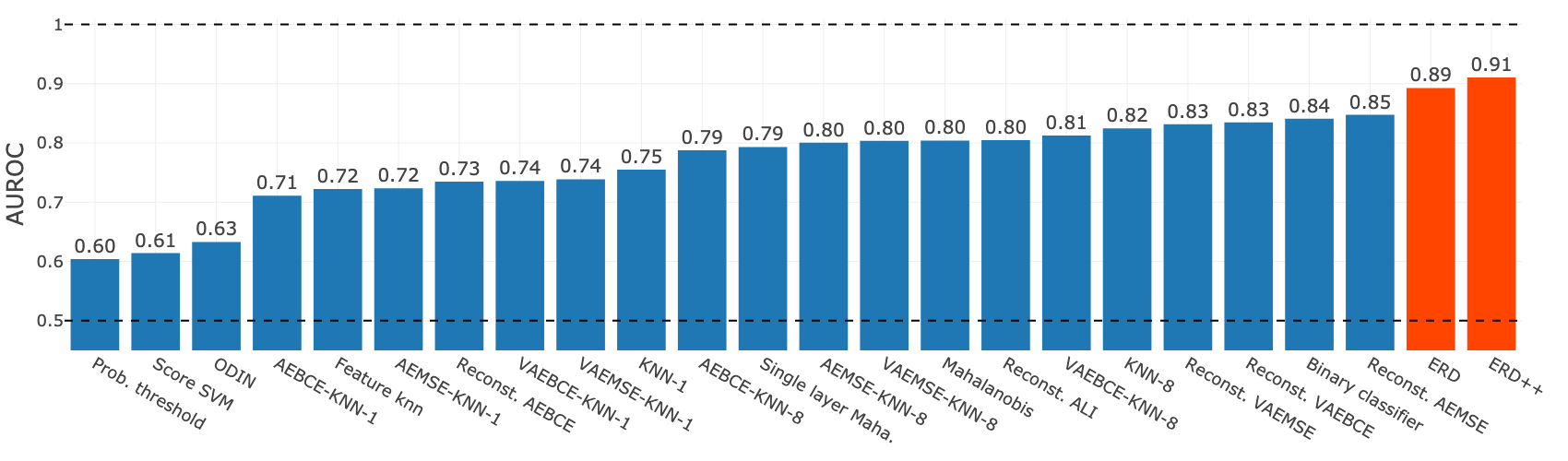}
  \end{center}

  \caption{AUROC averaged over all scenarios in the medical OOD detection
    benchmark \citep{Cao2020}. The values for all the baselines are computed
    using code made available by the authors of \citet{Cao2020}. Notably,
    most of the baselines assume oracle knowledge of OOD data at training time.}

  \label{fig:avg_medical_ood}
\end{figure}

\begin{figure}[H]
  \begin{center}
    \includegraphics[width=\textwidth]{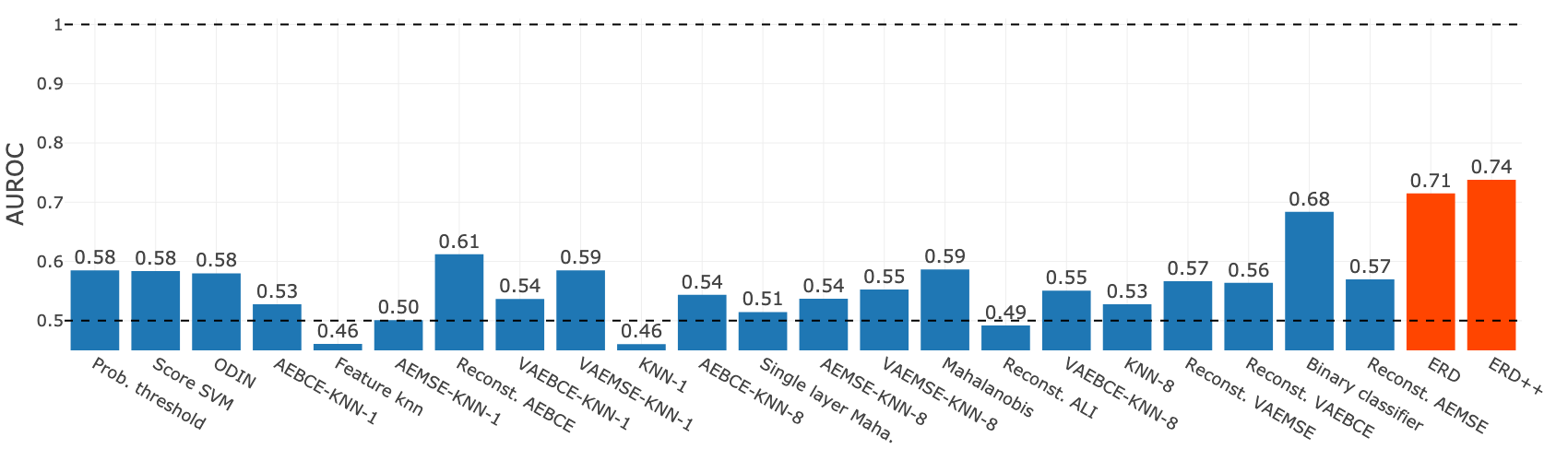}
  \end{center}

  \caption{AUROC averaged over the novel-class scenarios in the medical OOD
  detection benchmark \citep{Cao2020}, i.e.\ only use case 3.}

  \label{fig:avg_medical_novel_class}
\end{figure}

In addition, in Figure~\ref{fig:avg_medical_novel_class} we present the average
taken over only the novel-class settings in the medical benchmark. We observe
that the performance of all methods is drastically affected, all of them
performing much worse than the average presented in
Figure~\ref{fig:avg_medical_ood}. This stark decrease in AUROC and TNR@95
indicates that novelty detection is indeed a challenging task for OOD detection
methods even in realistic settings. Nevertheless, 2-model ERD ensembles maintain
a better performance than the baselines.

\begin{figure}[H]
  \begin{center}
    \includegraphics[width=0.7\textwidth]{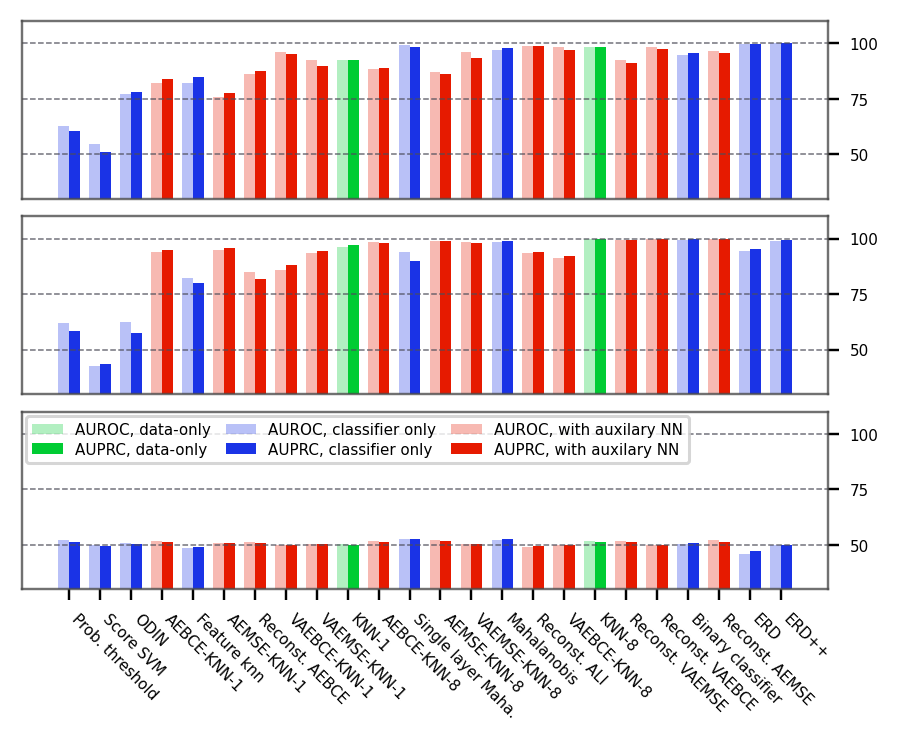}
  \end{center}

  \vspace{-0.5cm}
  \caption{Comparison between $\method$ and the various baselines on the NIH chest
  X-ray data set, for use case 1 (top), use case 2 (middle) and use case 3
(bottom). Baselines ordered as in Figure~\ref{fig:avg_medical_ood}.}
  \vspace{-0.2cm}

  \label{fig:medical_nih}
\end{figure}

\begin{figure}[H]
  \begin{center}
    \includegraphics[width=0.7\textwidth]{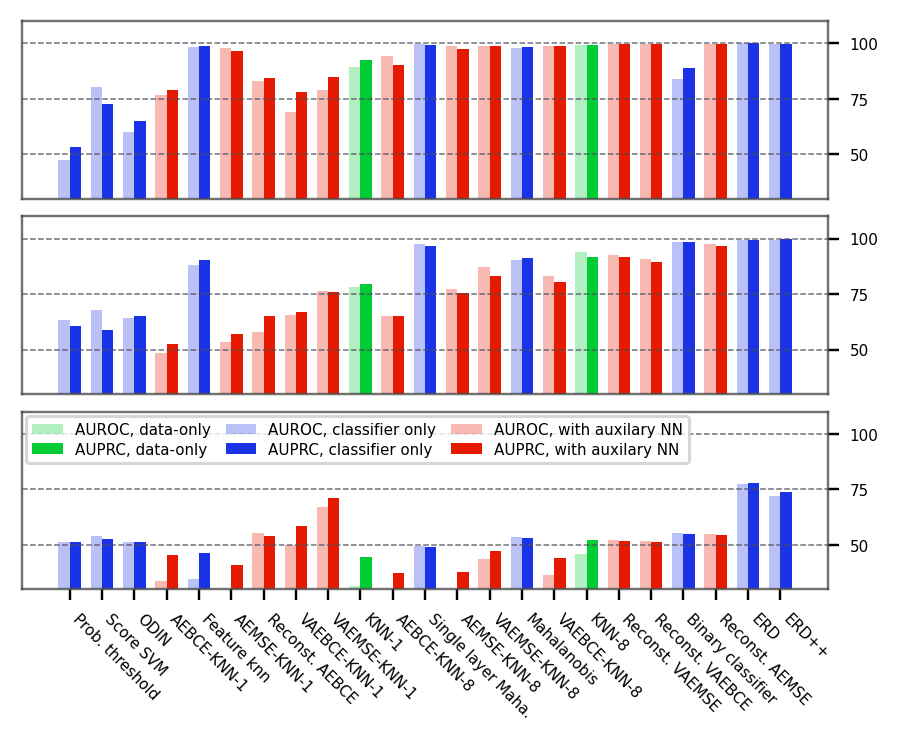}
  \end{center}

  \vspace{-0.5cm}
  \caption{Comparison between $\method$ and the various baselines on the PC chest
  X-ray data set, for use case 1 (top), use case 2 (middle) and use case 3
(bottom). Baselines ordered as in Figure~\ref{fig:avg_medical_ood}.}
  \vspace{-0.2cm}

  \label{fig:medical_pad}
\end{figure}

\begin{figure}[H]
  \begin{center}
    \includegraphics[width=0.7\textwidth]{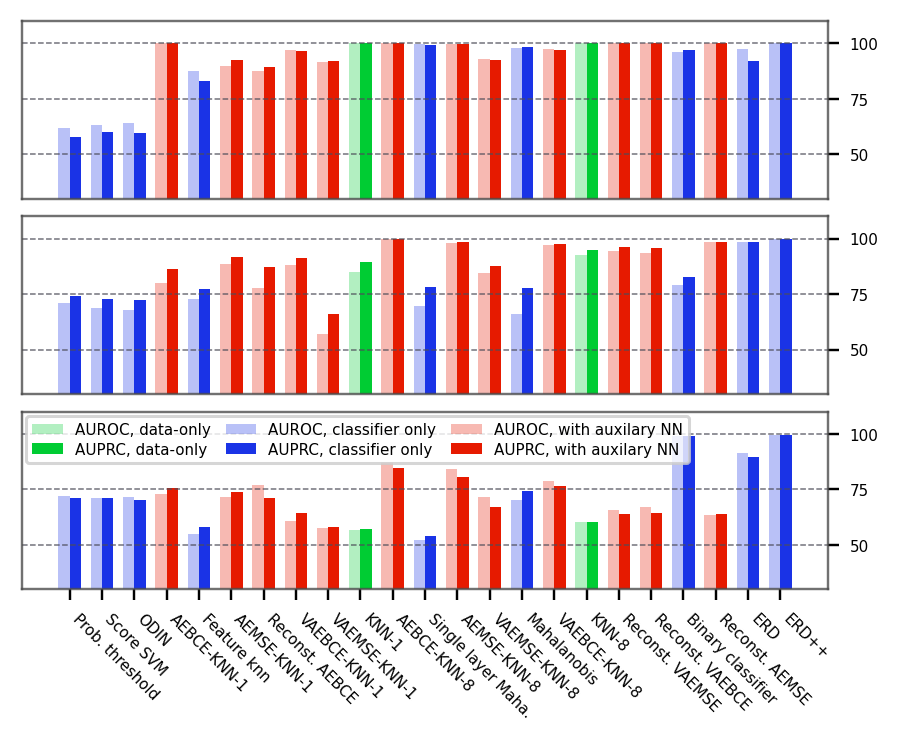}
  \end{center}

  \vspace{-0.5cm}
  \caption{Comparison between $\method$ and the various baselines on the DRD fundus
    imaging data set, for use case 1 (top), use case 2 (middle) and use case 3
(bottom). Baselines ordered as in Figure~\ref{fig:avg_medical_ood}.}

  \label{fig:medical_drd}
\end{figure}

In Figures~\ref{fig:medical_nih}, \ref{fig:medical_pad}, \ref{fig:medical_drd}
we present AUROC and AUPR (Area under the Precision Recall curve) for $\method$
for each of the training data sets, and each of the use cases.
Figure~\ref{fig:avg_medical_ood} presents averages over all settings that we
considered, for all the baseline methods in the benchmark.  Notably, $\method$
performs well consistently across data sets. The baselines are ordered by their
average performance on all the settings (see Figure~\ref{fig:avg_medical_ood}).

For all medical benchmarks, the unlabeled set is balanced, with an equal
number of ID and OOD samples (subsampling the bigger data set, if necessary). We
use the unlabeled set for evaluation.

  \vspace{-0.5cm}

\vspace{-0.3cm}


\vspace{-1cm}
\section{Effect of learning rate and batch size}
\label{sec:appendix_lr_bs}

\begin{figure}[H]
    \centering
    \includegraphics[width=0.5\textwidth]{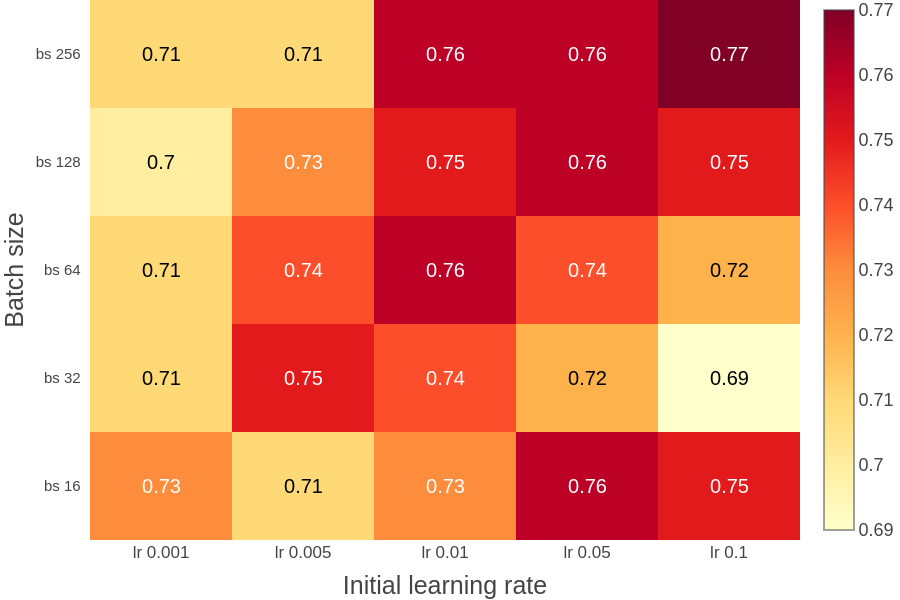}

    \caption{
        AUROCs obtained with an ensemble of WRN-28-10 models, as the initial
        learning rate and the batch size are varied.
        We used the hardest setting, CIFAR100:0-50 as ID, and
        CIFAR100:50-100 as OOD.
    }
    \label{fig:lr_and_bs_hyperparam}
\end{figure}

We show now that $\method$ ensembles are not too sensitive to the choice of
hyperparameters. We illustrate this by varying the learning rate and the batch
size, the hyperparameters that we identify as most impactful. As
Figure~\ref{fig:lr_and_bs_hyperparam} shows, many different configurations lead
to similar novelty detection performance.

\vspace{0.5cm}

\vspace{-0.2cm}
\section{Additional figure showing the dependence on the unlabeled set configuration}
\label{sec:appendix_vary_ood_ratio}

The configuration of the unlabeled set (i.e.\ the size of the unlabeled set, the
ratio of OOD samples in the unlabeled set) influences the performance of our
method, as illustrated in Figure~\ref{fig:vary_target_main}. Below, we show that
the same trend persists for different data sets too, e.g.\ when we consider
CIFAR10 as ID data and SVHN as OOD data.

\vspace{0.5cm}
\begin{figure}[h]
  \centering
  \includegraphics[width=0.7\textwidth]{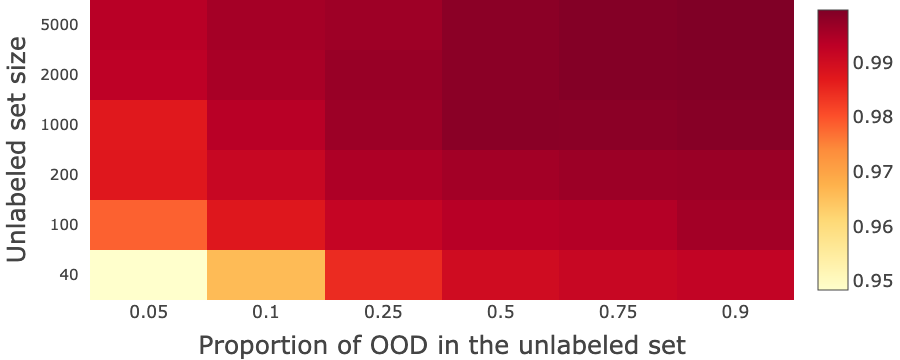}

  \label{fig:vary_target}

  \caption{ The AUROC of a 3-model ERD ensemble as the number and proportion of
  ID (CIFAR10) and OOD (SVHN) samples in the unlabeled set are varied.  }

\end{figure}


\section{Learning curves for other data sets}
\label{sec:appendix_learning_curves}

In addition to Figure~\ref{fig:training_curves}, we present in this section learning curves for
other data sets as well. The trend that persists throughout all figures is that
the arbitrary label is learned first on the unlabeled OOD data. Choosing a
stopping time before the validation accuracy starts to deteriorate prevents the
model from fitting the arbitrary label on unlabeled ID data.

\begin{figure*}[h]
  \centering

  \begin{subfigure}[r]{0.49\textwidth}
    \centering
    \includegraphics[width=1\textwidth]{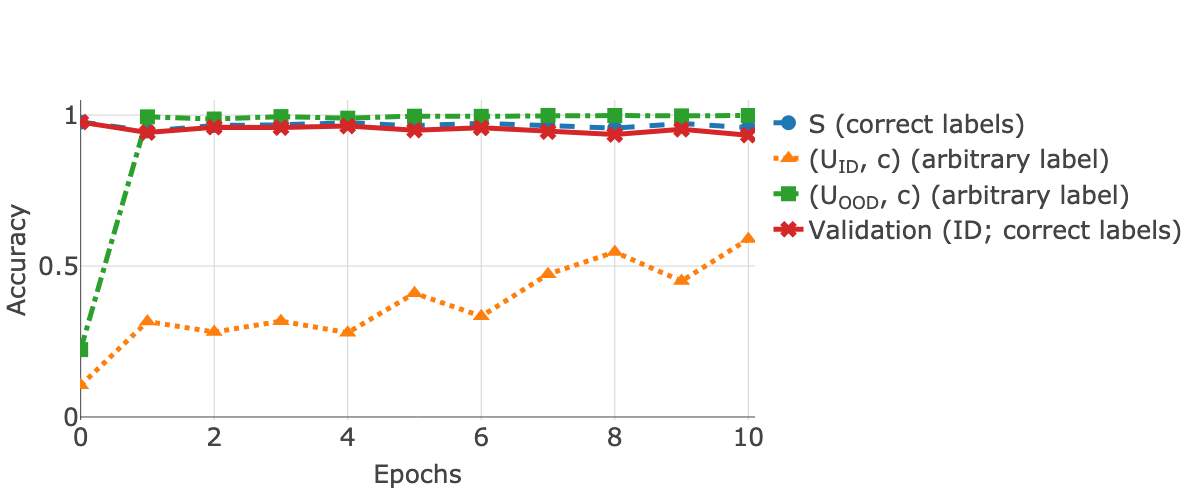}
    \caption{ID = SVHN; OOD = CIFAR10.}
  \end{subfigure}
  \hfill
  \begin{subfigure}[r]{0.49\textwidth}
    \centering
    \includegraphics[width=1\textwidth]{figures/training_curves_pretrained_svhn_cropped01234.png}
    \caption{ID = SVHN[0-4]; OOD = SVHN[5-9].}
  \end{subfigure}

  \begin{subfigure}[r]{0.49\textwidth}
    \centering
    \includegraphics[width=1\textwidth]{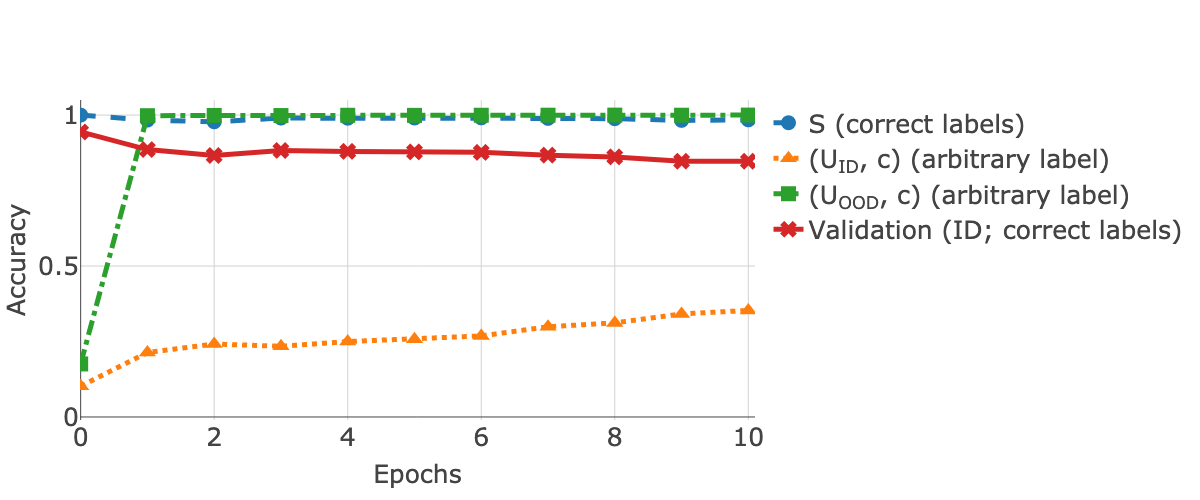}
    \caption{ID = CIFAR10; OOD = SVHN.}
  \end{subfigure}
  \hfill
  \begin{subfigure}[r]{0.49\textwidth}
    \centering
    \includegraphics[width=1\textwidth]{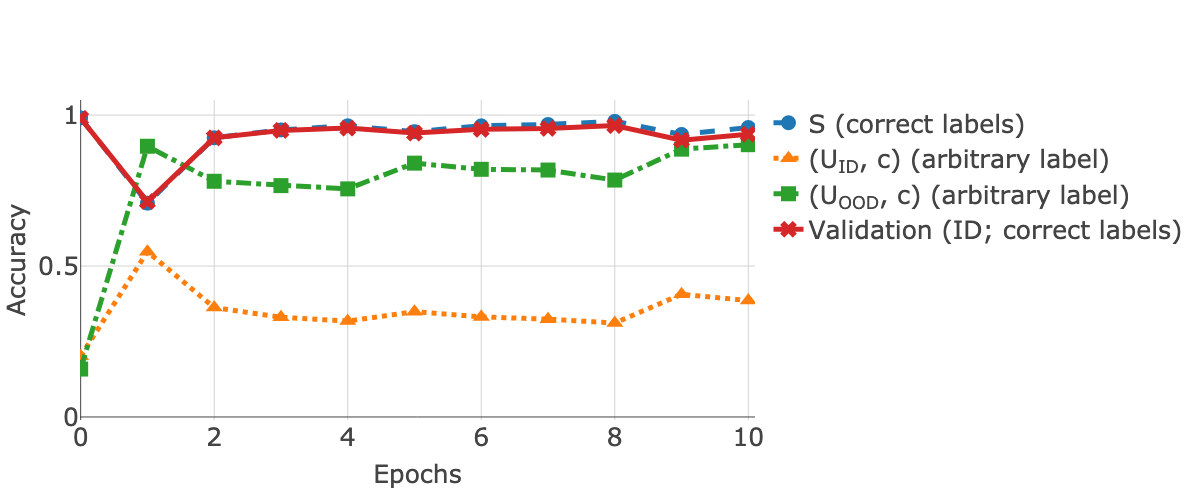}
    \caption{ID = CIFAR10[0-4]; OOD = CIFAR10[5-9].}
    \label{fig:learning_curves_cifar_split}
  \end{subfigure}

  \begin{subfigure}[r]{0.49\textwidth}
    \centering
    \includegraphics[width=1\textwidth]{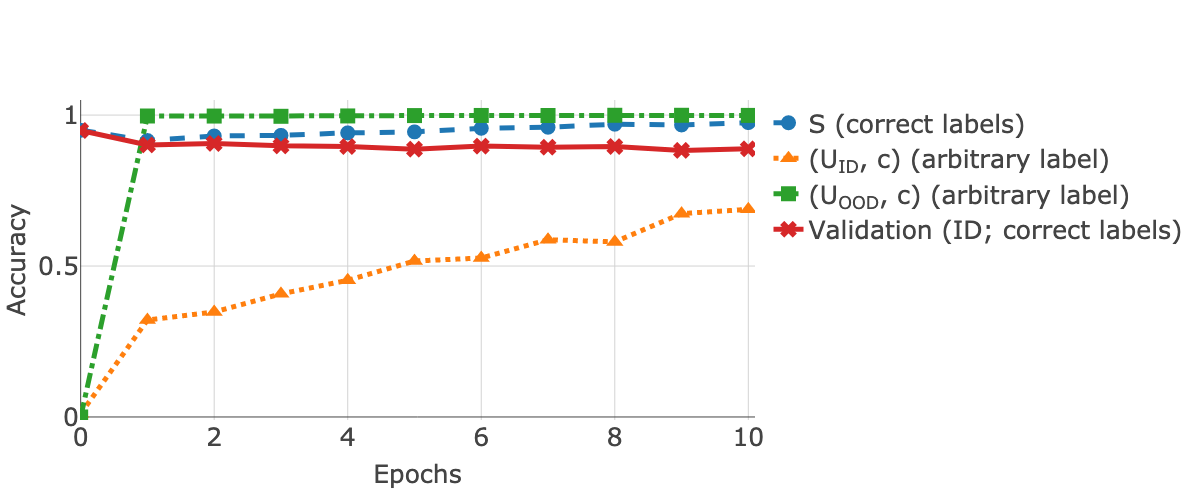}
    \caption{ID = CIFAR100; OOD = SVHN.}
  \end{subfigure}
  \hfill
  \begin{subfigure}[r]{0.49\textwidth}
    \centering
    \includegraphics[width=1\textwidth]{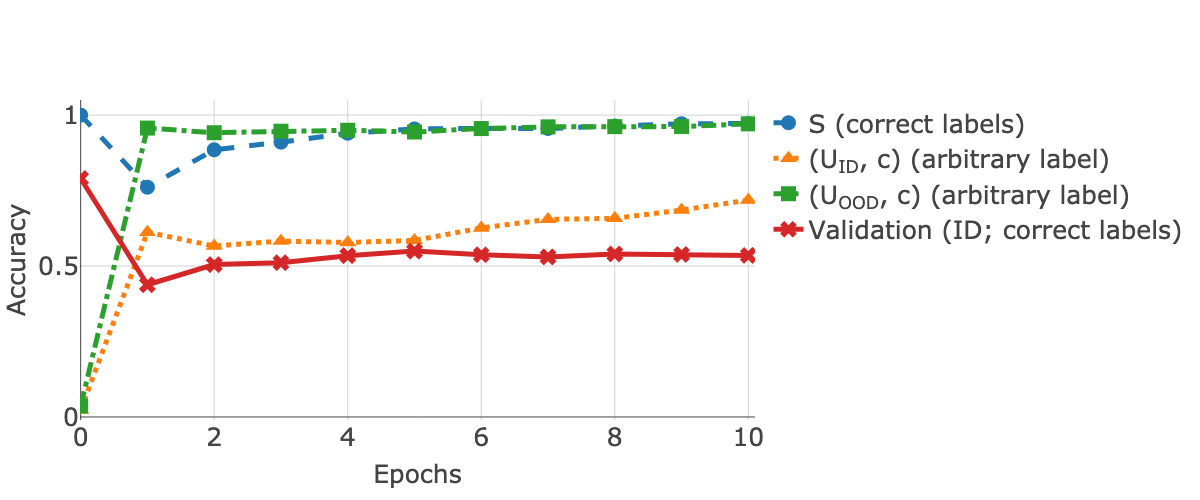}
    \caption{ID = CIFAR100[0-49]; OOD = CIFAR100[50-99].}
  \end{subfigure}

  \caption{ \small{Accuracy measured while fine-tuning a model
      pretrained on $\sourceset$ (epoch 0 indicates values obtained with the
      initial pretrained weights). The samples in $\labeledtargetood$ are fit first, while
      the model reaches high accuracy on $\labeledtargetid$ much later. We
      fine-tune for at least one epoch and then early stop when the validation
      accuracy starts decreasing.}}

  \label{fig:appendix_training_curves}
\end{figure*}

\paragraph{Impact of near OOD data on training $\method$ ensembles.} The
learning curves illustrated in Figure~\ref{fig:appendix_training_curves} provide
insight into what happens when the OOD data is similar to the ID training
samples and the impact that has on training the proposed method. In particular,
notice that for CIFAR10[0-4] vs CIFAR10[5-9] in
Figure~\ref{fig:learning_curves_cifar_split}, the models require more training
epochs before reaching an accuracy on unlabeled OOD samples of 100\%. The
learning of the arbitrary label on the OOD samples is delayed by the fact that
the ID and OOD data are similar, and hence, the bias of the correctly labeled
training set has a strong effect on the predictions of the models on the OOD
inputs. Since we early stop when the validation accuracy starts deteriorating
(e.g.\ at around epoch $8$ in Figure~\ref{fig:learning_curves_cifar_split}), we
end up using models that do not interpolate the arbitrary label on the OOD
samples. Therefore, the ensemble does not disagree on the entirety of the OOD
data in the unlabeled set, which leads to lower novelty detection performance.
Importantly, however, our empirical evaluation reveals that the drop in
performance for $\method$ ensembles is substantially smaller than what we
observe for other OOD detection methods, even on near OOD data sets.

\vspace{-0.3cm}
\section{Evolution of disagreement score during fine-tuning}
\label{sec:appendix_score_curves}

In this section we illustrate how the distribution of the disagreement score
changes during fine-tuning for ID and OOD data, for a 5-model ERD ensemble.
Thus, we can further understand why the performance of the $\method$ ensembles
is impacted by near OOD data.

\begin{figure*}[t]
  \centering

  \begin{subfigure}[r]{0.49\textwidth}
    \centering
    \includegraphics[width=1\textwidth]{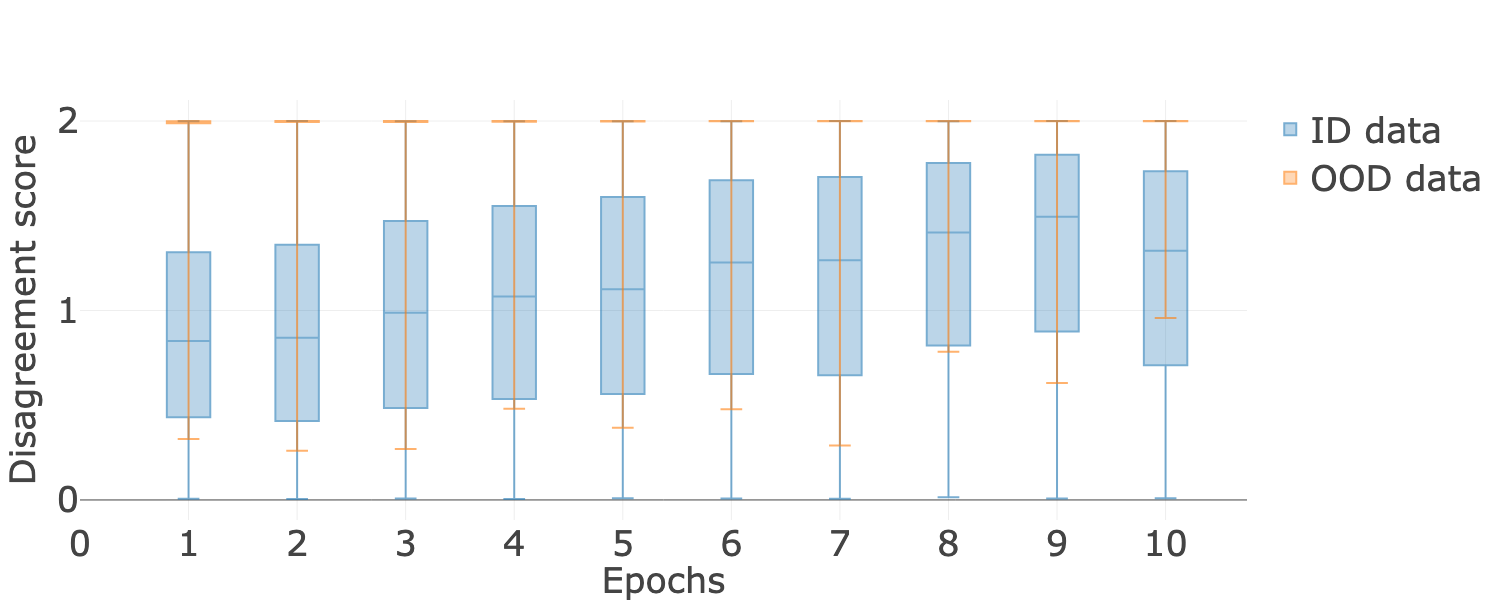}
    \caption{ID = SVHN; OOD = CIFAR10.}
  \end{subfigure}
  \hfill
  \begin{subfigure}[r]{0.49\textwidth}
    \centering
    \includegraphics[width=1\textwidth]{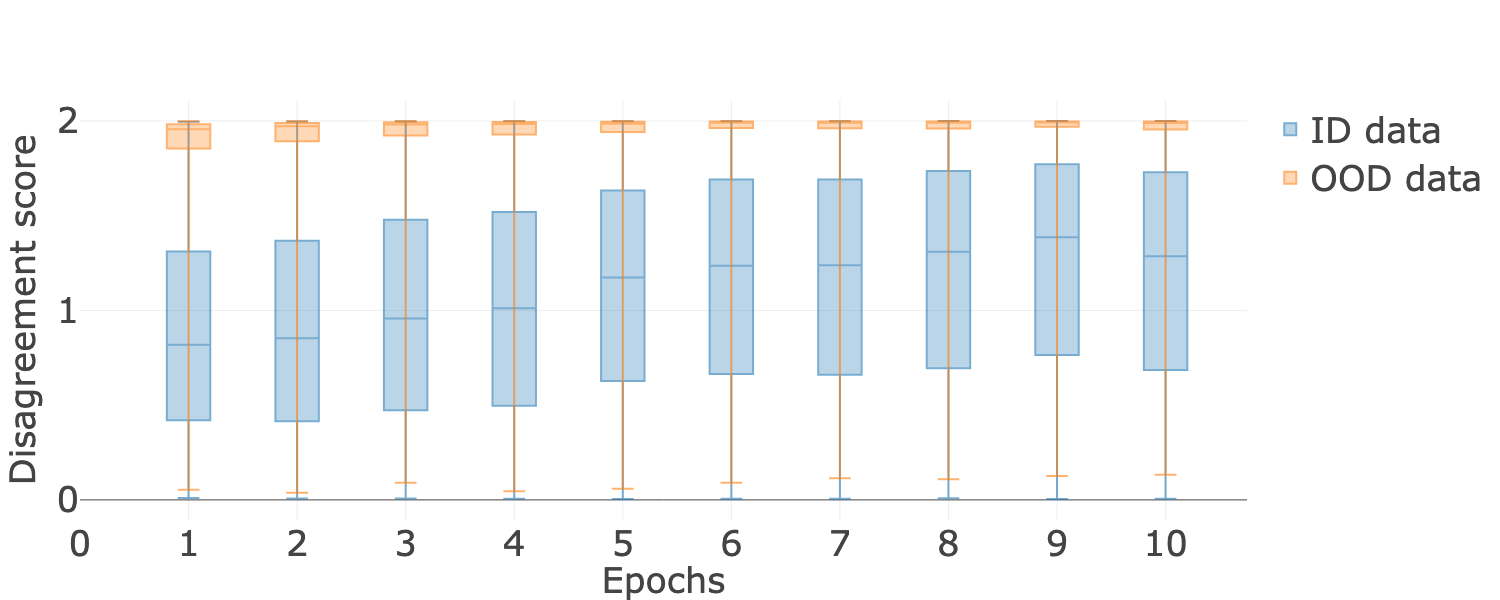}
    \caption{ID = SVHN[0-4]; OOD = SVHN[5-9].}
  \end{subfigure}

  \begin{subfigure}[r]{0.49\textwidth}
    \centering
    \includegraphics[width=1\textwidth]{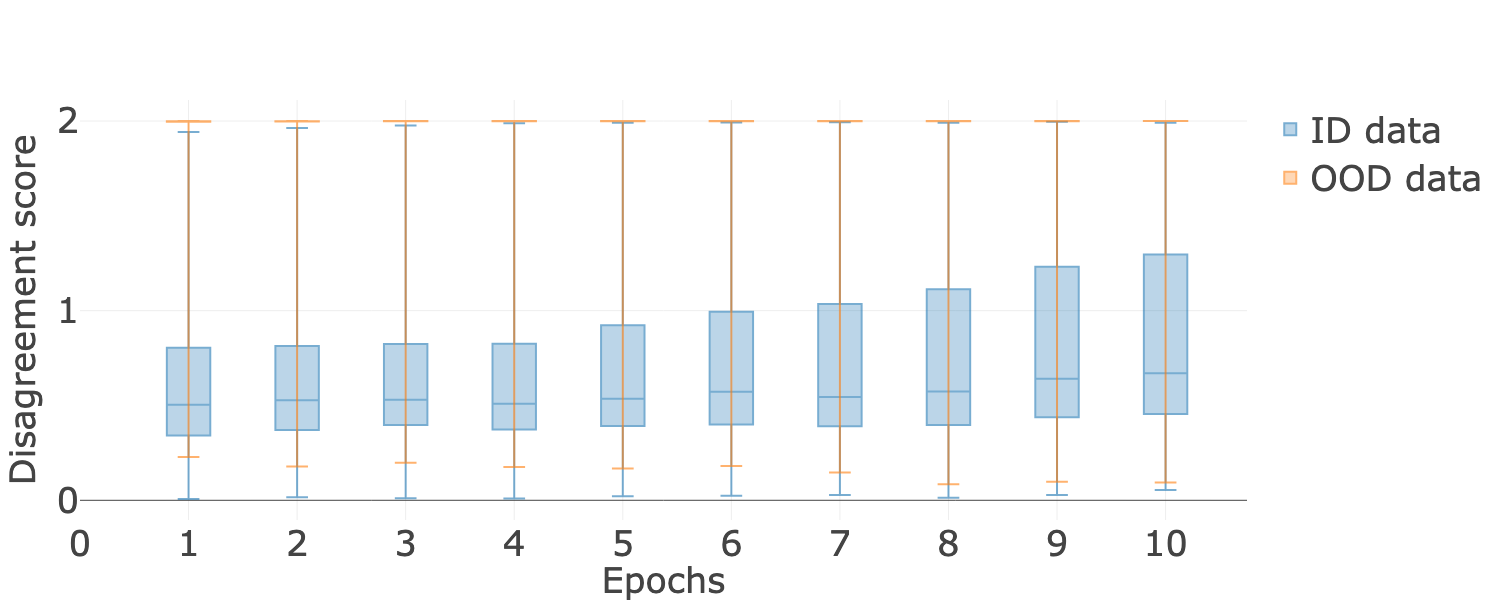}
    \caption{ID = CIFAR10; OOD = SVHN.}
  \end{subfigure}
  \hfill
  \begin{subfigure}[r]{0.49\textwidth}
    \centering
    \includegraphics[width=1\textwidth]{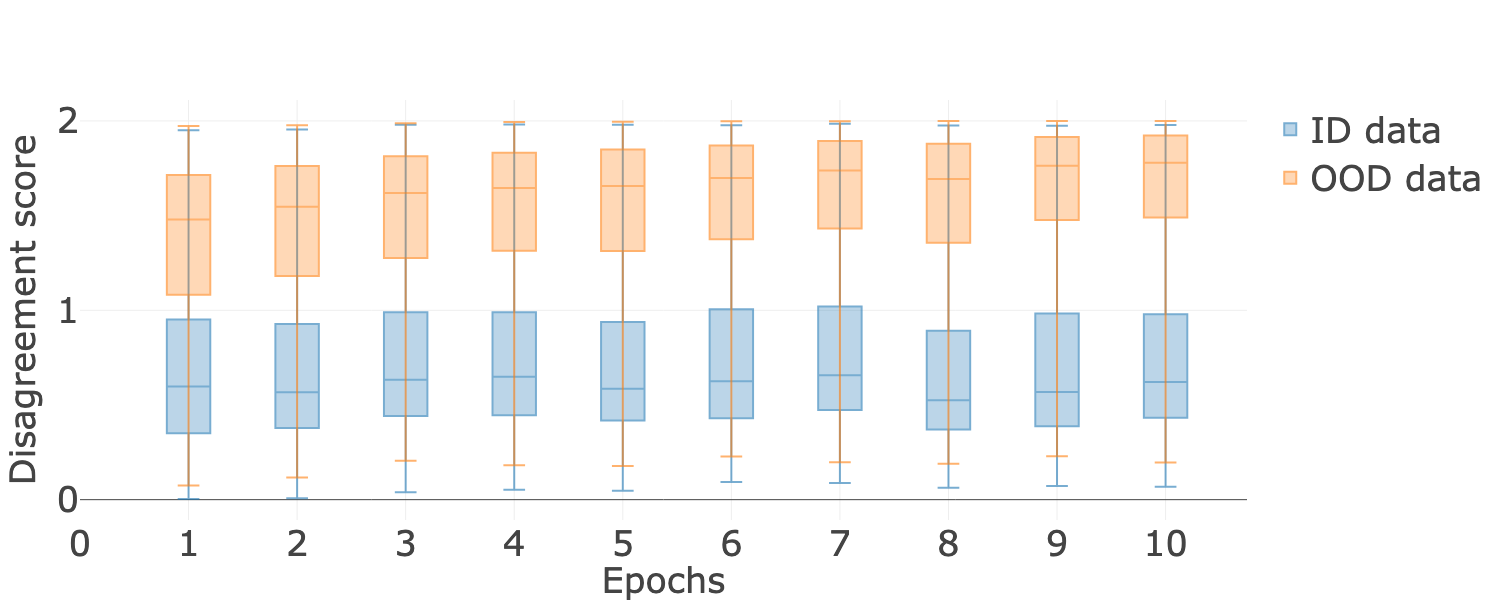}
    \caption{ID = CIFAR10[0-4]; OOD = CIFAR10[5-9].}
  \end{subfigure}

  \begin{subfigure}[r]{0.49\textwidth}
    \centering
    \includegraphics[width=1\textwidth]{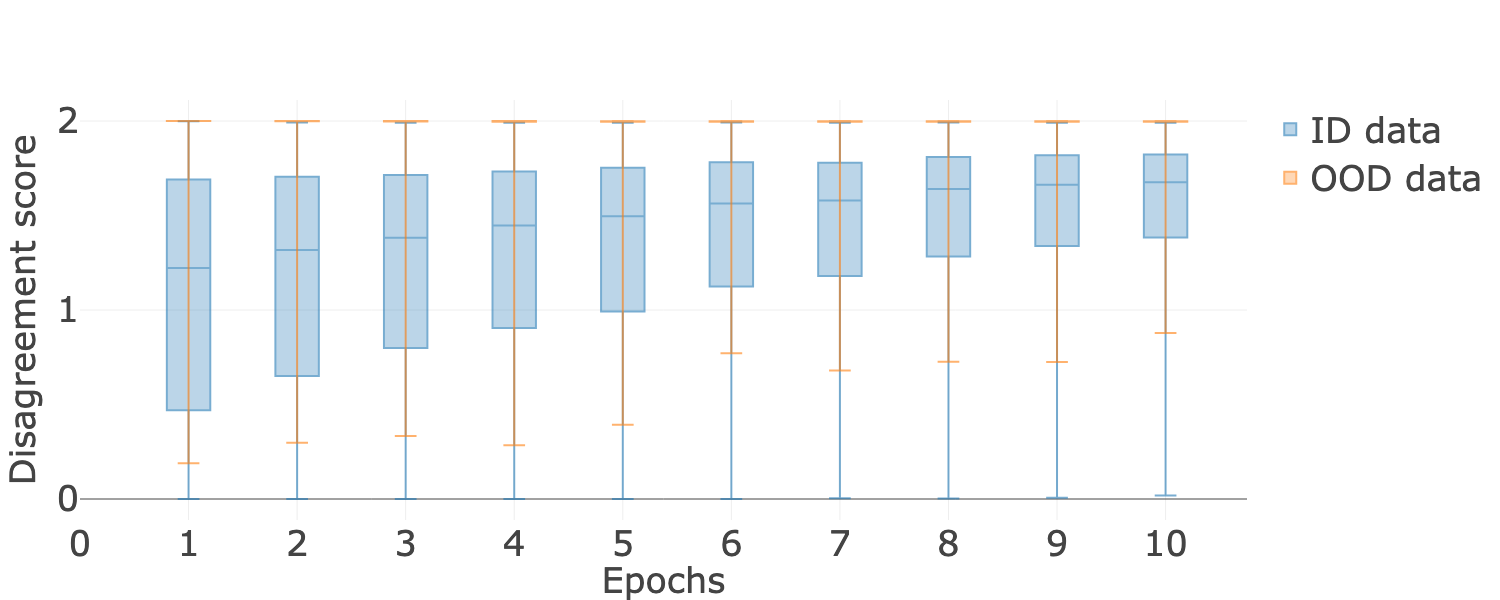}
    \caption{ID = CIFAR100; OOD = SVHN.}
  \end{subfigure}
  \hfill
  \begin{subfigure}[r]{0.49\textwidth}
    \centering
    \includegraphics[width=1\textwidth]{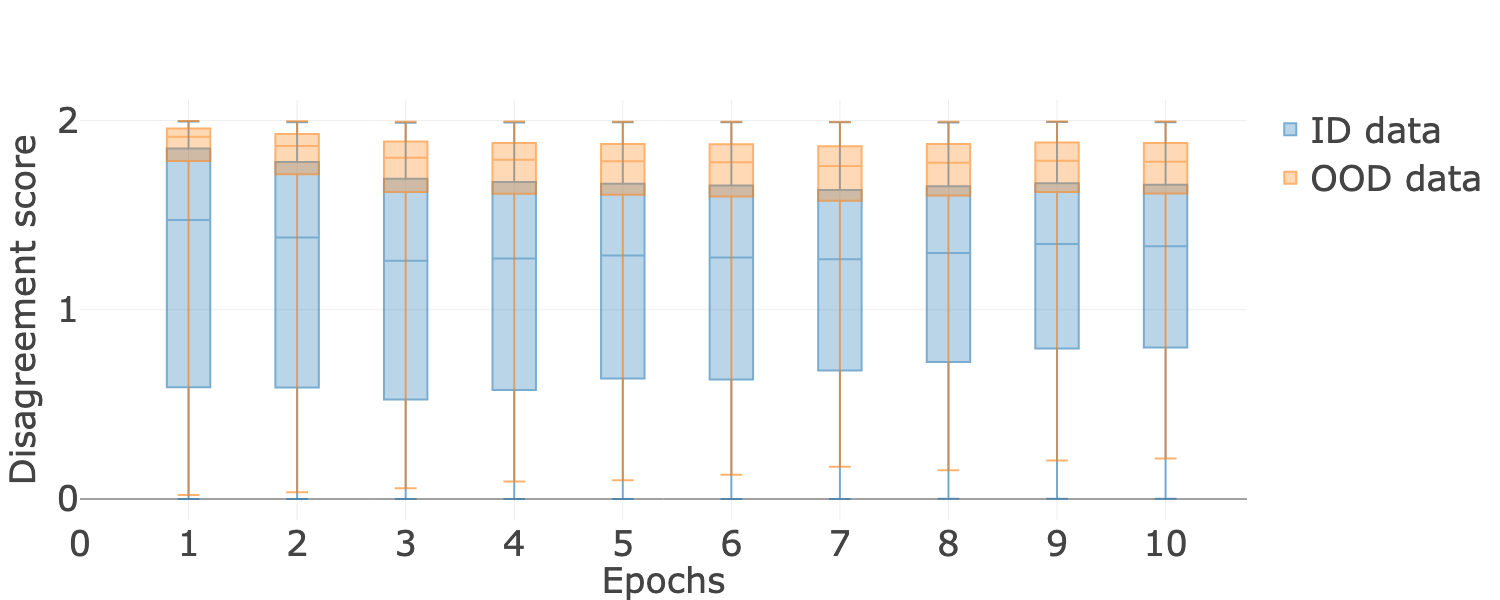}
    \caption{ID = CIFAR100[0-49]; OOD = CIFAR100[50-99].}
  \end{subfigure}

  \caption{ \small{The distribution of the disagreement score measured during
  fine-tuning on ID and OOD data (blue and orange boxes, respectively). The box
  indicates the lower and upper quartiles of the distribution, while the
  middle line represents the median and the whiskers show the extreme values.
  Notice that the distributions of the scores are easier to distinguish for far
  OOD data (left column), and tend to overlap more for near OOD settings (right
  column).}}

  \label{fig:appendix_score_curves}
\end{figure*}

Figure~\ref{fig:appendix_score_curves} reveals that for far OOD data (the left
column) the disagreement scores computed on OOD samples are well separated from
the disagreement scores on ID data (note that disagreement on OOD data is so
concentrated around the maximum value of $2$ that the boxes are essentially
reduced to a line segment). On the other hand, for near OOD data (the right
column) there is sometimes significant overlap between the disagreement scores
on ID and OOD data, which leads to the slightly lower AUROC values that we
report in Table~\ref{table:main_results}.

The figures also illustrate how the disagreement on the ID data tends to
increase as we fine-tune the ensemble for longer, as a consequence of the models
fitting the arbitrary labels on the unlabeled ID samples. Conversely, in most
instances one epoch suffices for fitting the arbitrary label on the OOD data.

We need to make one important remark: While in the figure we present
disagreement scores for the ensemble obtained after each epoch of fine-tuning,
we stress that the final $\method$ ensemble need not be selected among these. In
particular, since each model for $\method$ is early stopped separately,
potentially at a different iteration, it is likely that the $\method$ ensemble
contains models fine-tuned for a different number of iterations. Since we select
the $\method$ ensembles from a strictly larger set, the final ensemble selected
by the our proposed approach will be at least as good at distinguishing ID and
OOD data as the best ensemble depicted in
Figure~\ref{fig:appendix_score_curves}.

\end{document}